\documentclass[runningheads]{llncs}

\pdfoutput=1

\usepackage{eccv}
\usepackage{eccvabbrv}
\usepackage{comment}
\usepackage{graphicx}
\usepackage{booktabs}
\usepackage{epsfig}
\usepackage{enumitem}
\usepackage{multirow}
\usepackage{arydshln}	%
\usepackage{pdflscape}
\usepackage{rotating}
\usepackage{makecell}
\usepackage{wrapfig}
\usepackage{sidecap}
\usepackage{arydshln}
\usepackage{tabularray}
\usepackage{dsfont}
\usepackage{afterpage}
\usepackage{float}
\usepackage[usestackEOL]{stackengine}
\usepackage[hyperfootnotes=false]{hyperref}
\usepackage{placeins}

\usepackage{pifont}

\newcommand{\cmark}{\ding{51}}%
\newcommand{\xmark}{\ding{55}}%

\newcommand{\spacedbullet}{\enspace\textbullet\enspace}

\usepackage{colortbl}
\definecolor{colorFst}{HTML}{bde6cd}       %
\definecolor{colorSnd}{HTML}{e4eebc}       %
\definecolor{colorTrd}{rgb}{0.97, 0.91, 0.81}

\newcommand{\fs}{\cellcolor{colorFst}\bf}   %
\newcommand{\nd}{\cellcolor{colorSnd}\underline}      %
\newcommand{\rd}{\cellcolor{colorTrd}}      %

\newcommand{\PAR}[1]{\vskip4pt \noindent{\bf #1~}}

\begin{document}

\title{NeRFmentation: Improving Monocular Depth Estimation with NeRF-based Data Augmentation} 

\titlerunning{NeRFmentation}

\author{Casimir Feldmann\textsuperscript{\textasteriskcentered{}}\inst{1} \and
Niall Siegenheim\textsuperscript{\textasteriskcentered{}}\inst{1} \and
Nikolas Hars\inst{1} \and \\
Lovro Rabuzin\inst{1} \and
Mert Ertugrul\inst{1} \and
Luca Wolfart\inst{1} \and \\
Marc Pollefeys\inst{1,2} \and
Zuria Bauer\inst{1,3} \and
Martin R. Oswald\inst{1,4} \\
\vspace{6pt}
\email{\{cfeldmann, sniall, nihars, mertugrul, lrabuzin, \\ wolfartl, pomarc, zbauer, moswald\}@ethz.ch}
}

\authorrunning{C.~Feldmann et al.}

\institute{ETH Zürich \and
Microsoft \and
University of Alicante \and
University of Amsterdam
}

\maketitle

{\let\thefootnote\relax\footnote{{\textsuperscript{\textasteriskcentered{}}Authors contributed equally}}}

\begin{abstract}
\vspace{-20pt}
The capabilities of monocular depth estimation (MDE) models are limited by the availability of sufficient and diverse datasets. In the case of MDE models for autonomous driving, this issue is exacerbated by the linearity of the captured data trajectories. We propose a NeRF-based data augmentation pipeline to introduce synthetic data with more diverse viewing directions into training datasets and demonstrate the benefits of our approach to model performance and robustness. Our data augmentation pipeline, which we call \textit{NeRFmentation}, trains NeRFs on each scene in a dataset, filters out subpar NeRFs based on relevant metrics, and uses them to generate synthetic RGB-D images captured from new viewing directions. In this work, we apply our technique in conjunction with three state-of-the-art MDE architectures on the popular autonomous driving dataset, KITTI, augmenting its training set of the Eigen split. We evaluate the resulting performance gain on the original test set, a separate popular driving dataset, and our own synthetic test set. 

\begin{figure}
\centering
    \hspace{4pt}\includegraphics[width=0.95\textwidth]{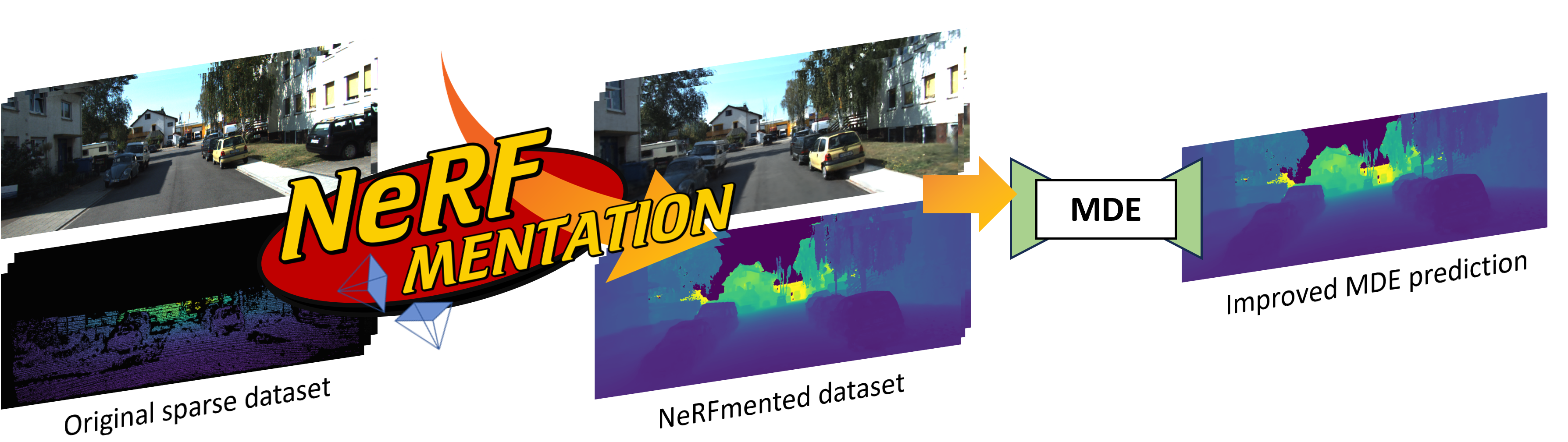}
    \caption{\textbf{NeRFmentation.} We propose a novel dataset augmentation pipeline that utilizes NeRF-generated data rendered from poses unseen in the original dataset to improve the robustness of Monocular Depth Estimation networks. The method is intended to improve generalization capability in cases where the initial dataset provides limited spatial variability. The RGB image is taken from the KITTI Dataset \cite{geiger_kitti_2013}.}
    \label{fig:teaserfigure}
\end{figure}

\end{abstract}    
\section{Introduction}

As the field of computer vision continues to advance at a rapid pace, the pursuit of safer and more reliable autonomous driving systems remains paramount. Within this context, monocular depth estimation (MDE) plays an important role, offering a key solution to the complex challenge of depth perception in dynamic road environments. However, it also comes with extra challenges compared to stereo and general multi-camera depth estimation. 
Accurately estimating the depth from a single image remains a challenging problem due to the inherent scale ambiguity in 2D projections of 3D scenes, which leads to an infinite number of geometrically plausible 3D reprojections for every 2D image. 
Therefore, monocular depth estimation is an underdetermined problem, which relies on extracting and interpreting visual cues in images correctly.
To address this challenge, numerous approaches leverage large-scale datasets to train deep neural networks, allowing them to learn to interpret visual cues and transform them into complex mappings between image features and corresponding depth information.

In order to make monocular depth estimators fit for real-world usage, their generalizability to unseen environments is of utmost importance. However, so far the overwhelming majority of research in this field has been dedicated to improving in-distribution test errors specifically on the KITTI dataset \cite{geiger_kitti_2013}. There is little knowledge available on how state-of-the-art models trained on these datasets perform on zero-shot out-of-distribution evaluation \cite{van_dijk_how_2019, ranftl_towards_2020, guizilini_towards_2023}.

Furthermore, we believe that improving the generalizability of MDE models does not require the costly and complex creation of new datasets, but can be achieved using dataset augmentation. In recent years, Neural Radiance Field (NeRF) \cite{mildenhall_nerf_2020} methods have emerged, which can encode complete 3D scenes. NeRFs have the ability to generate photorealistic and geometrically consistent, dense RGB-D images from novel viewpoints with remarkable quality. Therefore, they are powerful enough to use their output to train downstream models, such as monocular depth estimators.
Our main \textbf{contributions} are the following:
\begin{itemize}[itemsep=0pt,topsep=3pt,leftmargin=12pt,rightmargin=2pt,label=$\bullet$]
    \item We propose \textit{NeRFmentation}, a simple but effective dataset augmentation pipeline which uses NeRFs to generate high-quality RGB-D images from novel viewpoints to enlarge and diversify existing real-world depth estimation datasets to increase the robustness of trained models.
    \item We present extensive experiments on how well a variety of state-of-the-art models, trained on the KITTI dataset \cite{geiger_kitti_2013}, perform on zero-shot out-of-distribution evaluation, namely on the Waymo Open Dataset \cite{sun_scalability_2020}.
    \item We show that \textit{NeRFmentation} outperforms other dataset augmentation techniques both on in- and out-of-distribution test tasks. Furthermore,  we provide insights into when classic in-training augmentations (random rotations, flips, and crops) should be used in conjunction with dataset augmentations.
\end{itemize}

While we showcase the benefit of our data augmentation method for the task of monocular depth estimation, we envision the widespread usage of our approach for a variety of computer vision tasks such as semantic segmentation and surface normals estimation due to its simplicity and effectiveness.

\section{Related work}

\PAR{Monocular Depth Estimation.} 
This task involves predicting depth values for each pixel in a 2D image. Due to the inherent difficulty in predicting the correct 3D reprojection to a 2D image, well-performing models employ sophisticated data-driven methods to infer the depth values. Various approaches have been developed, including CNN-based encoder-decoder architectures \cite{ronnberger_2015_unet, eigen2014depth} and hybrid models combining CNNs with vision transformers (ViTs) \cite{dosovitskiy_2021_ViT}. Some methods treat MDE as a regression problem, while others approach it as an ordinal regression task \cite{depth-ordinal-regression}. Notable examples include AdaBins \cite{bhat_adabins_2021}, which uses adaptive bin centers and sizes, DepthFormer \cite{li2023depthformer}, which directly encodes images using ViTs, and BinsFormer \cite{li2022binsformer}, which combines ordinal regression on binned depth with transformer networks. Our work doesn't introduce a new MDE architecture but demonstrates the versatility of our method using these existing models \cite{bhat_adabins_2021, li2023depthformer, li2022binsformer}.

\PAR{Neural Radiance Fields (NeRFs).}
NeRFs are a scene representation method introduced in the seminal work \cite{mildenhall_nerf_2020} by Mildenhall~\etal. Using a sparse set of input views, they optimize an underlying volumetric scene function, taking advantage of a differentiable rendering pipeline. Novel views as well as depth data can then be synthesized by querying the network with novel camera poses. Nerfstudio \cite{tancik_nerfstudio_2023} facilitates development of NeRF-based architectures. Nerfacto, an extension of NeRF, incorporates recent advancements like camera pose refinement, per-image appearance conditioning, proposal sampling, scene contraction, and hash encoding. \texttt{depth-nerfacto} adds depth supervision \cite{deng_depth-supervised_2022}, while \texttt{nerfacto-huge} offers the highest fidelity among \texttt{nerfacto} variants.

\PAR{Zero-shot out-of-distribution model testing.}
While zero-shot model testing has been a common practice in the field of natural language processing \cite{yin_benchmarking_2019, rohrbach_evaluating_2011, thakur_beir_2021} and other subfields in computer vision \cite{xian_zero-shot_2019, zhu_zero_2020}, only recently has this trend reached monocular depth estimation. Van Dijk \etal\cite{van_dijk_how_2019} analyze the inherent biases towards the training dataset learned by MDEs by introducing synthetic objects into the testing dataset and checking how well the models react to them. MiDaS \cite{ranftl_towards_2020} is one of the works combining different datasets to improve the generalizability of MDEs to unseen data. Yin \etal\cite{yin_metric3d_2023} design a model architecture that allows predicting depth for arbitrary images by projecting different camera setups into a canonical space before estimating the depth and subsequently projecting back. Guizilini \etal\cite{guizilini_towards_2023} train their intrinsics-aware MDE on five datasets and evaluate it on four other datasets and show impressive results. However, they are not able to match current state-of-the-art models trained on these evaluation datasets.

\PAR{Offline and online data augmentation.}
Data augmentation is a well-known regularization technique in Machine Learning that increases generalizability by applying label-preserving transforms to input samples, either during training (online) \cite{simard_cnnaugment_2003, krizhevsky_imagenet_2012} or before training (offline) \cite{CycleGAN2017,Yang_2020_fda_DA,Gatys_2016_style-transfer-DA, atapour-abarghouei_real-time_2018}. Classical methods for CNNs typically involve simple transformations like random translations, rotations, or flips \cite{simard_cnnaugment_2003, krizhevsky_imagenet_2012}. Recent advancements in generative modeling, particularly GANs \cite{CycleGAN2017, sandfort_cycleganaugment_2019, besnier_nonexistentdataset_2019,atapour-abarghouei_real-time_2018} and diffusion models \cite{azizi_synthetic_2023}, have enabled the offline generation of photo-realistic synthetic data.
Various approaches have been developed to improve model performance. Sandfort \etal \cite{sandfort_cycleganaugment_2019} use CycleGAN for medical CT segmentation, improving performance even for out-of-distribution samples. Atapour-Abarghouei \etal \cite{atapour-abarghouei_real-time_2018} train MDEs on synthetic video game data and evaluate on KITTI using style transfer. Gasperini \etal \cite{gasperini_robust_2023} demonstrate online augmentation for MDE in adverse driving conditions. While recent methods often transfer styles in the image domain \cite{CycleGAN2017, Yang_2020_fda_DA, atapour-abarghouei_real-time_2018}, our approach uniquely augments the pose space while maintaining image space consistency.

\PAR{Synthetic datasets and virtual views for MDE.}
Rajpal \etal show in \cite{rajpal_gtaVmde_2023} that it is possible to improve monocular depth estimation performance on real-world data using a fully synthetic custom dataset.  
Mancini \etal\cite{TDI} use two synthetic outdoor datasets to train an MDE and subsequently evaluate it on KITTI, where they achieve results comparable to Eigen \etal \cite{eigen2014depth}, as well as their own synthetic datasets.
Bauer \etal\cite{bauer_nvs-monodepth_2021} use in-training novel view synthesis to improve depth prediction. %
In contrast to \cite{rajpal_gtaVmde_2023, TDI}, we do not use a fully synthetic dataset to improve monocular depth estimators but propose to use an existing dataset and extend it with virtual samples like Bauer \etal \cite{bauer_nvs-monodepth_2021}. However, our pipeline allows for more diverse poses and can be used independently of the training procedure as we generate samples before the training of the MDE network.
\section{Method}

\begin{figure}
    \centering
    \includegraphics[width=\linewidth]{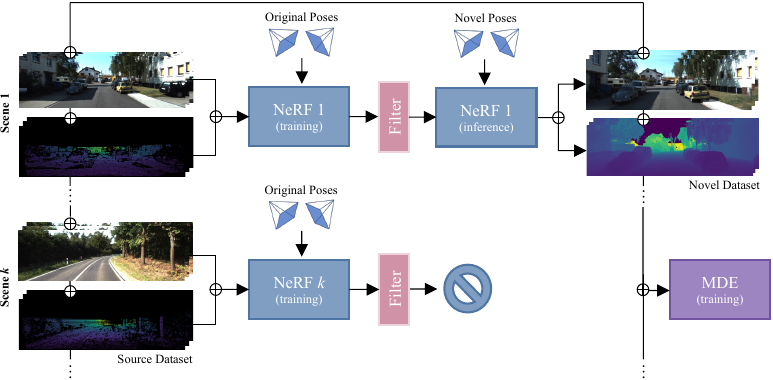}
    \caption{\textbf{Our proposed pipeline:} (1) Train NeRFs for each scene in MDE dataset, reserving images for quality evaluation. (2) Filter out subpar NeRFs. (3) Render novel views by perturbing original poses. (4) Combine novel and original views to create \textit{NeRFmented} training dataset for MDE network. Source dataset: KITTI \cite{geiger_kitti_2013}}
    \label{fig:pipeline}
\end{figure}

We propose a novel dataset augmentation technique for Monocular Depth Estimation which leverages the fact that many real-world image datasets have a sequence-like structure, meaning that the same environment is captured from many slightly different viewpoints. Using the provided pose information, we can build 3D models of the scenes in a dataset, which can subsequently be used to generate new RGB-D pairs from slightly different viewpoints, increasing the size of the original dataset.

\subsection{Proposed pipeline}

We split each scene in the source dataset into sub-scenes and train a separate depth-supervised NeRF using RGB-D images and their corresponding poses to encode a 3D representation. To ensure a high reconstruction quality for each NeRF, we withhold a small part of the input data from the training. Once the training has concluded, we use the withheld data as a small validation split to potentially filter out scenes with subpar image- or depth reconstruction quality. 
We construct novel views by applying minor 3D rigid body transformations to the original poses. Then, we use the trained NeRFs to render dense RGB-D images from these novel viewpoints.
Finally, we combine the RGB-D images from our novel views with the source dataset to form an augmented dataset and train the downstream monocular depth estimation network on it. The overall pipeline is depicted in \cref{fig:pipeline}.

\subsection{Scene representation model}
\label{sec:method_novel_view_synth}
\begin{figure*}[t]
    \centering
    \scriptsize
    \setlength{\tabcolsep}{1pt}
	\renewcommand{\arraystretch}{0.8}
	\newcommand{\sz}{0.48}
	\newcommand{\sh}{1.2cm}
	\newcommand{\gcs}{\hspace{8pt}}  %
	\begin{tabular}{ccc}
        & Depth-Nerfacto & Depth-Nerfacto-Huge \\
        \rotatebox[origin=c]{90}{RGB} &
        \makecell{\includegraphics[width=\sz\linewidth]{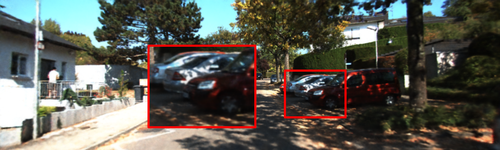}} &
        \makecell{\includegraphics[width=\sz\linewidth]{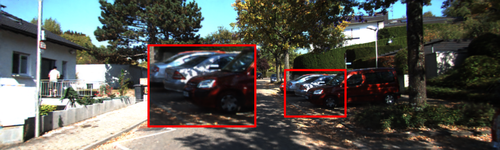}} \\
        \rotatebox[origin=c]{90}{Depth} & 
        \makecell{\includegraphics[width=\sz\linewidth]{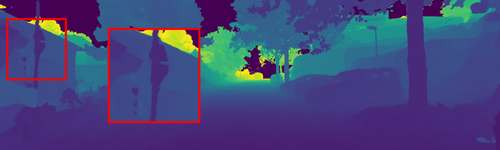}} &
        \makecell{\includegraphics[width=\sz\linewidth]{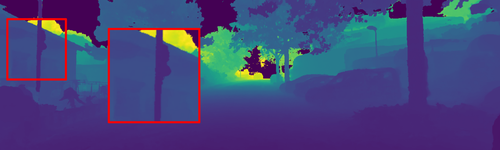}} \\
    \end{tabular}
    \caption{\textbf{Qualitative comparison of \texttt{depth-nerfacto} vs. \texttt{depth-nerfacto-huge} reconstruction on KITTI~\cite{geiger_kitti_2013}.} The figure shows reconstructions of training images using both \texttt{depth-nerfacto} and \texttt{depth-nerfacto-huge}. On average, \texttt{depth-nerfacto-huge} outputs exhibit higher levels of sharpness and better accuracy in both the generated RGB and depth images compared to \texttt{depth-nerfacto}.}
    \label{fig:depth_vs_huge}
\end{figure*}

For the NeRFs of our pipeline, we use the off-the-shelf Nerfstudio framework \cite{tancik_nerfstudio_2023}, which provides convenient access to various NeRF methods and extensions.
Because our method requires high-quality depth reconstruction and image reconstruction, we base our model on a combination of the \texttt{depth-nerfacto} and \texttt{nerfacto-huge} architectures, which were both introduced by Tancik \etal in \cite{tancik_nerfstudio_2023}.
This way, we benefit from depth supervision that is included in the \texttt{depth-nerfacto} training while increasing the learning capacity of the network by replacing the MLP of \texttt{depth-nerfacto} with the larger MLP of \texttt{nerfacto-huge}.
\cref{fig:depth_vs_huge} shows the qualitative performance gain of our \texttt{depth-nerfacto-huge} compared to the vanilla \texttt{depth-nerfacto}.

\section{Experiments}

\subsection{Datasets}

We consider the following two datasets in our experiments.

\PAR{KITTI~\cite{geiger_kitti_2013}.} 
The KITTI dataset is an outdoor dataset designed for diverse autonomous driving scenarios, featuring stereo-image pairs and annotated LiDAR ground-truth depth data. The KITTI Eigen split \cite{eigen2014depth} is a popular dataset split for monocular depth estimation with approximately 23,000 stereo-image pairs for training which are separated into \textbf{32} scenes for training and validation. The corresponding validation split consists of \textbf{888} stereo-image pairs. The test split contains \textbf{697} images from \textbf{29} different scenes.    

\PAR{Waymo Open Dataset \cite{sun_scalability_2020}.} 
The Waymo Open Dataset, developed by Waymo LLC for autonomous driving tasks, incorporates data from five high-resolution cameras and five LiDARs. It offers wider camera viewpoints and denser depth maps compared to datasets like KITTI, enabling more robust model training. To evaluate our \textit{NeRFmented} models' robustness, we test them on a subset of the Waymo test dataset, using only front, front-left, and front-right facing camera images. This selection minimizes domain shift from standard forward-facing camera MDE datasets like KITTI.

\subsection{Dataset augmentation baselines}
We evaluate the benefits of using our \textit{NeRFmented} KITTI dataset for zero-shot MDE on the Waymo test dataset by comparing it with the domain-adaptation methods Fourier Domain Analysis (FDA) \cite{Yang_2020_fda_DA} and style transfer \cite{atapour-abarghouei_real-time_2018}. 

For FDA, we only use a single Fourier-domain value of $\beta=0.0025$, which corresponds to replacing the amplitudes of the lowest 0.25\,\% of the source image frequencies (KITTI) with the amplitudes of the lowest 0.25\,\% of the target frequencies (Waymo). This way the domain gap is reduced slightly, but no significant artifacts are introduced. 
Regarding style transfer, we augment the original KITTI Eigen train split with style-transferred RGBs as described in \cite{atapour-abarghouei_real-time_2018}, leading to a total dataset size of 29,158 images (+26\,\%). The ground truth depth maps remain unchanged. 
Furthermore, we also ablate the effect of using classical data augmentation techniques in the training pipeline consisting of random rotations, flips, and crops.

\subsection{Novel view synthesis strategies for data augmentation}
We propose and evaluate five data augmentation strategies: 
(\textit{i}) \textbf{Reconstruction}: Re-rendering the exact training poses used to train the NeRF. This completely densifies the sparse depth map from the source dataset.
(\textit{ii}) \textbf{Interpolated}: Creating novel poses by interpolating between pairs of training poses. As NeRFs interpolate between embeddings, these renders benefit from good depth and RGB supervision. 
(\textit{iii}) \textbf{Angled}: Rendering two additional views for each training pose, rotated horizontally by $\pm 3^{\circ}$. This simulates plausible motion performed by a driving car. 
(\textit{iv, v}) \textbf{Translated horizontally, vertically}: Generating two additional views for each original pose by translating the camera horizontally/vertically $\pm 30\,cm$. 
For the KITTI Eigen train split \cite{eigen2014depth, geiger_kitti_2013} this results in dataset sizes of \textit{i}) 29,836 (+28\,\%), \textit{ii}) 29,744 (+28\,\%), \textit{iii}) 32,132 (+39\,\%), \textit{iv}) 34,949 images (+51\,\%), and \textit{v}) 34,949 images (+51\,\%). An ablation on the optimal amount of \textit{NeRFmented} data to add is provided in the supplementary.

The motivation behind these data augmentation strategies is to analyze the influence of depth-densification and pose diversification. With this diverse collection, we can evaluate whether it is sufficient to use NeRFs as depth completion networks or whether certain novel views of training scenes can improve performance on evaluation datasets.

\begin{figure}[!t]
    \centering
    \scriptsize
    \setlength{\tabcolsep}{1pt}
	\renewcommand{\arraystretch}{0.8}
	\newcommand{\sz}{0.46}
	\newcommand{\sh}{1.2cm}
	\newcommand{\gcs}{\hspace{8pt}}  %
\resizebox{0.8\textwidth}{!}{
	\begin{tabular}{cc@{\gcs}}
        \rotatebox{90}{\hspace{6pt} GT RGB}
        \includegraphics[width=\sz\columnwidth]{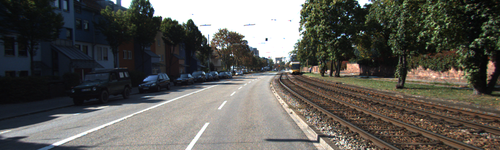} &
        \includegraphics[width=\sz\columnwidth]{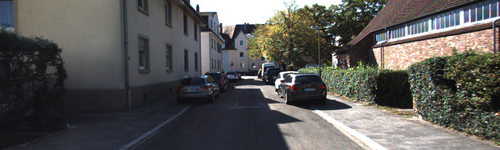} \\
        \rotatebox{90}{\hspace{6pt} Our RGB}
        \includegraphics[width=\sz\columnwidth]{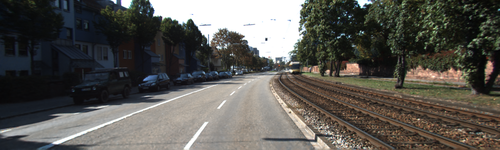} &
        \includegraphics[width=\sz\columnwidth]{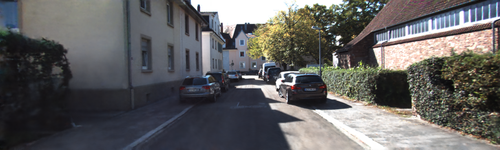} \\
        \rotatebox{90}{\hspace{2pt} GT Depth}
        \includegraphics[width=\sz\columnwidth]{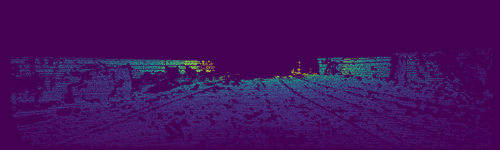} &
        \includegraphics[width=\sz\columnwidth]{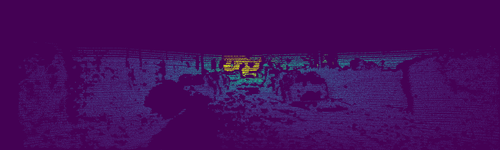} \\
        \rotatebox{90}{\hspace{2pt} Our Depth}
        \includegraphics[width=\sz\columnwidth]{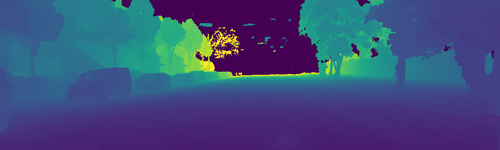} &
        \includegraphics[width=\sz\columnwidth]{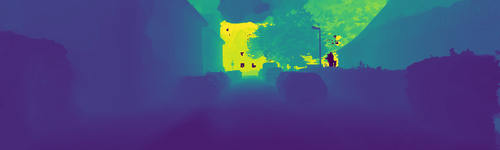} \\
    \end{tabular}}
    \caption{\textbf{Qualitative NeRF reconstruction results on KITTI~\cite{geiger_kitti_2013}}. Original KITTI images are compared with those generated by trained and filtered NeRFs for matching camera poses. The reconstructed RGB images closely resemble the originals, while NeRFs also complete sparse ground truth depth maps.}
    \label{fig:nerf_reconstruction}
\end{figure}

\subsection{Training of NeRFs}
We train a \texttt{depth-nerfacto-huge} model for each scene of the KITTI Eigen \cite{geiger_kitti_2013, eigen2014depth} train split and retain 10\,\% of the training data as a validation split.
As we experimentally find that the reconstruction quality of NeRFs drastically reduces as the size of a scene increases, the varying sizes of the scenes in the KITTI dataset require splitting them into sub-scenes.
However, it is unfavorable to decrease the scene size excessively, as this would increase the computational complexity without a gain in quality. To achieve a good balance, we ensure that no pose in each sub-scene is further than 50\,m away from the first pose in that sub-scene. For each sub-scene, we train a separate NeRF.

For each NeRF, we tune the weight of the depth loss and whether the poses are refined or not by running a grid search for 10,000 steps. 
We observe that when using NeRFs for 3D reconstruction, there is a strong trade-off between RGB quality and accuracy of the geometry, which in turn affects the accuracy of the depth maps. Since we require both to be of high quality, we use the small validation split that we withheld from the training and re-render the RGB-D images for each of their poses using all NeRFs from our grid search. Then, we select the NeRF that minimizes a trade-off validation measure $ T_\text{RGBD}=\sqrt{\alpha \cdot \text{Abs. Rel}^2 + \mathrm{LPIPS}^2} $, where we use the absolute relative error of the reprojected depth maps as a quality measure of the geometry and the LPIPS score \cite{zhang_unreasonable_2018} as a quality measure of the RGB reconstruction. During our experiments, we set $\alpha = 10$.

Finally, we continue training the best run of the hyperparameter tuning of each NeRF until we reach 30,000 iterations. Then, we apply a filtering step that eliminates NeRFs that fail to meet a specified dataset-dependent reconstruction quality standard. To achieve this, we utilize the trained NeRF for each scene to render the RGB-D images from the poses of the small withheld validation split and compare them to the corresponding ground-truth RGB-D images. This involves computing the absolute relative error to assess the depth reconstruction quality and the LPIPS score \cite{zhang_unreasonable_2018} to evaluate the image reconstruction quality. We discard all NeRFs that do not achieve an LPIPS score of less than $0.22$ or an absolute relative error of less than $0.05$ on the 3D reconstructions. The threshold for LPIPS is chosen empirically by observing which value produces minimally visually acceptable RGB images. The threshold for depth absolute relative error is chosen because it is an approximate lower bound of state-of-the-art MDE networks. We filter any scenes with a higher score so as not to introduce extraneous noise into the training dataset.

Training a \texttt{depth-nerfacto-huge} model on a sub-scene takes approximately 45~min on an RTX 4090. In our case, we obtained 434 sub-scenes from KITTI. Therefore, our total GPU compute for training the NeRFs was around 2 weeks. Generating a novel view at the native KITTI resolution takes around 1.4 seconds per frame. We note that both training the NeRFs and generating novel views can easily be parallelized on a sub-scene level when multiple GPUs are available. As this is a data augmentation strategy, there is no performance penalty when performing inference.

\begin{table*}[!tb]
 \centering
    \caption{\textbf{Comparison of data augmentation methods for models trained on augmented KITTI datasets evaluated on the Waymo dataset.} \small We showcase the performance of baseline MDEs and compare them to the \textit{NeRFmented} (Ours) versions of the same models evaluated against the unseen Waymo dataset. For \textit{NeRFmentation}, the NeRFm. Vert. $\pm 30\,cm$ strategy is shown. CAug indicates whether classic augmentations were used during training. The best results are in \colorbox{colorFst}{\textbf{bold}}. The second best results are \colorbox{colorSnd}{\underline{underlined}}, and the \colorbox{colorTrd}{baseline} is highlighted.}  
    \scalebox{0.8}{
    \begin{tabular}{clccccccc} 
        \toprule
     
        & Method & $\delta_{1}\uparrow$ & $\delta_{2}\uparrow$    & $\delta_{3}\uparrow$ & REL $\downarrow$ & SQ. REL $\downarrow$ & RMS $\downarrow$  & RMS$_{LOG}$ $\downarrow$ \\
        \midrule

        \multirow{4}{*}{\rotatebox{90}{CAug. \cmark}} & \textbf{Adabins}  &   \rd{0.561} & \rd{0.851} & \rd{0.933} & \rd{0.224} & \rd{2.201} & \rd{8.027} & \rd{0.328} \\
        & \textcolor{orange}{\spacedbullet}FDA~\cite{Yang_2020_fda_DA}      &  0.597 & \nd{0.884} & \nd{0.952} & 0.210 & \nd{1.906} & \nd{7.467} & 0.297 \\
                                    
        & \textcolor{orange}{\spacedbullet}Style Transfer~\cite{atapour-abarghouei_real-time_2018} &  \nd{0.618} & \fs{0.895} & \fs{0.957} & \fs{0.205} & \fs{1.836} & \fs{7.207} & \fs{0.287} \\ 
        & \textcolor{cyan}{\spacedbullet}NeRFmentation & \fs{0.666} & 0.866 & 0.937 & \nd{0.209} & 2.317 & 7.667 & \nd{0.293} \\
        \midrule
        \multirow{4}{*}{\rotatebox{90}{CAug. \xmark}} & \textbf{Adabins}                             & 0.543 & 0.854 & \nd{0.938} & 0.228 & 2.248 & 8.562 & 0.330 \\
        & \textcolor{orange}{\spacedbullet}FDA~\cite{Yang_2020_fda_DA}  & 0.584 & \nd{0.861} & \nd{0.938} & 0.226 & \fs{2.180} & \nd{7.983} & \nd{0.320} \\
        & \textcolor{orange}{\spacedbullet}Style Transfer~\cite{atapour-abarghouei_real-time_2018}  &  \nd{0.591} & 0.847 & 0.927 & \nd{0.225} & 2.300 & 8.215 & 0.329 \\
        & \textcolor{cyan}{\spacedbullet}NeRFmentation & \fs{0.629} & \fs{0.873} & \fs{0.946} & \fs{0.220} & \nd{2.189} & \fs{7.770} & \fs{0.295} \\
        \midrule
        \multirow{4}{*}{\rotatebox{90}{CAug. \cmark}} & \textbf{DepthFormer} & \nd{0.545} & \rd{0.885} & \rd{0.958} & \rd{0.230} & \rd{1.920} & \nd{7.776} & \rd{0.313} \\
        & \textcolor{orange}{\spacedbullet}FDA~\cite{Yang_2020_fda_DA}                       & 0.479 & \nd{0.891} & \nd{0.964} & \nd{0.225} & \nd{1.895} & 7.793 & \nd{0.308} \\
       & \textcolor{orange}{\spacedbullet}Style Transfer~\cite{atapour-abarghouei_real-time_2018}           & 0.470 & 0.880 & 0.958 & 0.230 & 1.931 & 7.867 & 0.319 \\
        &\textcolor{cyan}{\spacedbullet}NeRFmentation                & \fs{0.724} & \fs{0.926} & \fs{0.975} & \fs{0.170} & \fs{1.489} & \fs{6.777} & \fs{0.238} \\
        \midrule
                \multirow{4}{*}{\rotatebox{90}{CAug. \xmark}} &\textbf{DepthFormer} & \nd{0.479} & \nd{0.882} & \nd{0.961} & \nd{0.232} & \nd{1.987} & \nd{8.006} & \nd{0.313} \\
        & \textcolor{orange}{\spacedbullet}FDA~\cite{Yang_2020_fda_DA}  &  0.445 & 0.873 & 0.960 & 0.238 & 2.052 & 8.209 & 0.322 \\
        & \textcolor{orange}{\spacedbullet}Style Transfer~\cite{atapour-abarghouei_real-time_2018}   &  0.464 & 0.864 & 0.954 & 0.236 & 2.155 & 8.544 & 0.328 \\
        &\textcolor{cyan}{\spacedbullet}NeRFmentation & \fs{0.764} & \fs{0.937} & \fs{0.979} & \fs{0.160} & \fs{1.355} & \fs{6.365} & \fs{0.222} \\
        \midrule
        \multirow{4}{*}{\rotatebox{90}{CAug. \cmark}} & \textbf{BinsFormer}         &  \rd{0.571} & \rd{0.901} & \rd{0.966} & \rd{0.208} & \rd{1.782} & \rd{7.547} & \rd{0.289} \\
        & \textcolor{orange}{\spacedbullet}FDA~\cite{Yang_2020_fda_DA}  & \nd{0.599} & \nd{0.914} & \nd{0.971} & 0.201 & \nd{1.676} & \nd{7.209} & \nd{0.276} \\
        & \textcolor{orange}{\spacedbullet}Style Transfer~\cite{atapour-abarghouei_real-time_2018}            & 0.545 & 0.891 & 0.961 & \nd{0.214} & 1.837 & 7.718 & 0.301 \\
        & \textcolor{cyan}{\spacedbullet}NeRFmentation                & \fs{0.744} & \fs{0.928} & \fs{0.975} & \fs{0.168} & \fs{1.527} & \fs{6.617} & \fs{0.231} \\
        \midrule
        \multirow{4}{*}{\rotatebox{90}{CAug. \xmark}} & \textbf{BinsFormer}  & \nd{0.615} & \nd{0.905} & \nd{0.968} & \nd{0.202} & \nd{1.746} & \nd{7.505} & \nd{0.279} \\
        & \textcolor{orange}{\spacedbullet}FDA~\cite{Yang_2020_fda_DA} &  0.600 & 0.903 & 0.967 & 0.206 & 1.794 & 7.601 & 0.284 \\
        & \textcolor{orange}{\spacedbullet}Style Transfer~\cite{atapour-abarghouei_real-time_2018}             & 0.571 & 0.894 & 0.963 & 0.210 & 1.840 & 7.882 & 0.294 \\
        & \textcolor{cyan}{\spacedbullet}NeRFmentation                & \fs{0.770} & \fs{0.938} & \fs{0.978} & \fs{0.166} & \fs{1.416} & \fs{6.181} & \fs{0.220} \\
        \bottomrule
    \end{tabular}}
  \label{tab:waymo_datset_augmentation}
\end{table*}

\cref{fig:nerf_reconstruction} shows the images reconstructed by NeRFs alongside ground-truth images present in the KITTI dataset. Apart from a low LPIPS score, visual inspection of the reconstructed RGB images also suggests that they are of comparable quality to the original. Additionally, the reconstructed depth maps are dense, whereas the ground-truth ones are sparse. The low absolute relative error of the reconstructed depth maps indicates a high agreement between the original and reconstructed depth maps as well.

\begin{table*}[!tb]
 \centering

    \caption{\textbf{Ablation of \textit{NeRFmentation} strategies for models trained on the augmented KITTI dataset evaluated on the Waymo dataset and the KITTI Eigen [10,14] test split.} \small We showcase the performance of AdaBins \cite{bhat_adabins_2021}, DepthFormer \cite{li2023depthformer}, and BinsFormer \cite{li2022binsformer} models trained on the KITTI Eigen \cite{eigen2014depth, geiger_kitti_2013} train split and compare \textit{NeRFmentation} strategies. No classic augmentation is used. The best \textit{NeRFmentation} strategy to use depends on the chosen depth architecture. The best results are in \colorbox{colorFst}{\textbf{bold}}. The second best results are \colorbox{colorSnd}{\underline{underlined}}.} 
\resizebox{\textwidth}{!}{
    \begin{tabular}{cl|ccccccc|ccccccc} 
        \toprule
        \multicolumn{2}{c}{Dataset}&\multicolumn{7}{c}{\textbf{Waymo}}&\multicolumn{7}{c}{\textbf{KITTI}}\\
        \midrule     
        & Augmentation & $\delta_{1}^\uparrow$ & $\delta_{2}^\uparrow$    & $\delta_{3}^\uparrow$ & REL$^\downarrow$ & SQ. REL$^\downarrow$ & RMS$^\downarrow$  & RMS$_{LOG}^\downarrow$ &  $\delta_{1}^\uparrow$ & $\delta_{2}^\uparrow$    & $\delta_{3}^\uparrow$ & REL$^\downarrow$ & SQ. REL$^\downarrow$ & RMS$^\downarrow$  & RMS$_{LOG}^\downarrow$  \\
        \midrule
        \multirow{5}{*}{\rotatebox[origin=c]{90}{Adabins}}&\textcolor{cyan}{\spacedbullet}NeRFm. Reconstruction & 0.667 & {0.893} & \fs{0.956} & 0.201 & \fs{1.932} & \fs{7.335} & \fs{0.276} & \fs{0.958} & \nd{0.994} & 0.999 & \nd{0.062} & \fs{0.210} & \fs{2.506} & \fs{0.094}\\
        &\textcolor{cyan}{\spacedbullet}NeRFm. Interpolation &  \nd{0.680} & {0.875} & 0.944 & \fs{0.197} & \nd{2.024} & \nd{7.352} & 0.281 & \nd{0.957} & \fs{0.995} & 0.999 & \fs{0.061} & \nd{0.213} & 2.524 & \fs{0.094}\\
        &\textcolor{cyan}{\spacedbullet}NeRFm. Angled $\pm 3^{\circ}$ & {0.669} & \fs{0.886} & \nd{0.953} & {0.204} & 2.040 & 7.470 & {0.280} & \fs{0.958} & \nd{0.994} & 0.999 & \nd{0.062} & \nd{0.213} & \nd{2.513} & \nd{0.095} \\
        &\textcolor{cyan}{\spacedbullet}NeRFm. Horiz. $\pm 30\,cm$ & \fs{0.684} & \nd{0.884} & 0.952 & 0.202 & 2.039 & 7.593 & \nd{0.279} & 0.955 & 0.993 & 0.999 & 0.065 & 0.234 & 2.595 & 0.098\\
        &\textcolor{cyan}{\spacedbullet}NeRFm. Vert. $\pm 30\,cm$ & 0.629 & 0.873 & 0.946 & {0.220} & 2.189 & 7.770 & 0.295 & \fs{0.958} & \nd{0.994} & 0.999 & \nd{0.062} & 0.215 & 2.534 & \nd{0.095}\\
        \midrule
        \multirow{5}{*}{\rotatebox[origin=c]{90}{DepthFormer}}&\textcolor{cyan}{\spacedbullet}NeRFm. Reconstruction &  {0.735} & 0.932 & \nd{0.977} & \nd{0.167} & 1.433 & {6.540} & {0.234} & \fs{0.974} & 0.997 & \nd{0.999} & \nd{0.054} & 0.160 & 2.178 & \nd{0.081} \\
        &\textcolor{cyan}{\spacedbullet}NeRFm. Interpolation &  \nd{0.739} & \nd{0.936} & \fs{0.979} & {0.168} & {1.420} & \nd{6.418} & \nd{0.231} & \nd{0.973} & 0.997 & \nd{0.999} & \fs{0.053} & \nd{0.158} & 2.178 & \nd{0.081}\\
        &\textcolor{cyan}{\spacedbullet}NeRFm. Angled $\pm 3^{\circ}$ & {0.721} & {0.933} & \nd{0.977} & 0.171 & \nd{1.418} & 6.569 & {0.238} & \nd{0.973} & 0.997 & \nd{0.999} & \nd{0.054} & \nd{0.158} & 2.165 & \nd{0.081} \\
        &\textcolor{cyan}{\spacedbullet}NeRFm. Horiz. $\pm 30\,cm$ & 0.730 & 0.929 & 0.976 & 0.168 & 1.451 & 6.740 & 0.238 & \fs{0.974} & 0.997 & \fs{1.000} & \nd{0.054} & \nd{0.158} & \nd{2.158} & \nd{0.081}\\
        &\textcolor{cyan}{\spacedbullet}NeRFm. Vert. $\pm 30\,cm$ & \fs{0.764} & \fs{0.937} & \fs{0.979} & \fs{0.160} & \fs{1.355} & \fs{6.365} & \fs{0.222} & \fs{0.974} & 0.997 & \fs{1.000} & \fs{0.053} & \fs{0.156} & \fs{2.145} & \fs{0.080}\\
        \midrule
        \multirow{5}{*}{\rotatebox[origin=c]{90}{BinsFormer}}&\textcolor{cyan}{\spacedbullet}NeRFm. Reconstruction &  {0.744} & {0.933} & {0.977} & \nd{0.171} & {1.447} & {6.476} & {0.230} & \nd{0.973} & 0.997 & 0.999 & \nd{0.055} & \nd{0.159} & {2.181} & 0.082\\
        &\textcolor{cyan}{\spacedbullet}NeRFm. Interpolation &  0.727 & {0.933} & {0.977} & {0.174} & {1.462} & {6.481} & 0.236 & \nd{0.973} & 0.997 & 0.999 & \fs{0.054} & \nd{0.159} & 2.191 & \nd{0.081}\\
        &\textcolor{cyan}{\spacedbullet}NeRFm. Angled $\pm 3^{\circ}$ & \nd{0.764} & \fs{0.940} & \fs{0.979} & \fs{0.166} & \fs{1.395} & \nd{6.228} & \nd{0.224} & \fs{0.975} & 0.997 & 0.999 & \fs{0.054} & \fs{0.154} & \fs{2.131} & \fs{0.080}\\
        &\textcolor{cyan}{\spacedbullet}NeRFm. Horiz. $\pm 30\,cm$ & 0.736 & 0.933 & \nd{0.978} & {0.177} & 1.501 & 6.483 & 0.234 & \nd{0.973} & 0.997 & 0.999 & \fs{0.054} & 0.160 & \nd{2.173} & \nd{0.081}\\
        &\textcolor{cyan}{\spacedbullet}NeRFm. Vert. $\pm 30\,cm$ & \fs{0.770} & \nd{0.938} & \nd{0.978} & \fs{0.166} & \nd{1.416} & \fs{6.181} & \fs{0.220} & 0.972 & 0.997 & 0.999 & \nd{0.055} & 0.161 & 2.187 & 0.082\\
        \bottomrule
    \end{tabular}}
  \label{tab:waymo_dataset_strategies}
\end{table*}

\subsection{Training of monocular depth estimation models}
To demonstrate that our augmentation technique is model-agnostic, we show it in combination with three popular MDE architectures: AdaBins \cite{bhat_adabins_2021}, DepthFormer \cite{li2023depthformer}, and BinsFormer \cite{li2022binsformer}. We use the Monocular-Depth-Estimation-Toolbox \cite{lidepthtoolbox2022}, an open-source monocular depth estimation toolbox aimed at collecting several state-of-the-art MDE models into a single environment. We train the described off-the-shelf MDE networks on the augmented KITTI Eigen \cite{geiger_kitti_2013, eigen2014depth} train split and compare the results to Classic, FDA, and Style Transfer augmentation results. Due to computational constraints, we cannot use the original batch size used for DepthFormer and BinsFormer. Hence, following the linear scaling rule described in \cite{goyal2018accurate}, we decrease the learning rate and increase the number of iterations accordingly. Other hyperparameters are kept as described in the corresponding papers to show the influence of the augmentation methods. For the evaluation of the MDE models, we use the same metrics as in \cite{bauer_nvs-monodepth_2021}. For their definitions, we refer to the supplementary material.

\begin{figure}[!ht]
        \hspace{-0.4cm}
    \centering
    \scriptsize
    \setlength{\tabcolsep}{1pt}
	\renewcommand{\arraystretch}{0.8}
	\newcommand{\sz}{0.32}
	\newcommand{\sh}{1.65cm}
\resizebox{0.8\textwidth}{!}{
    \begin{tabular}{cccc}
         RGB & AdaBins~\cite{bhat_adabins_2021} & NeRFmented AdaBins~\cite{bhat_adabins_2021} \\
        \includegraphics[width=\sz\linewidth]{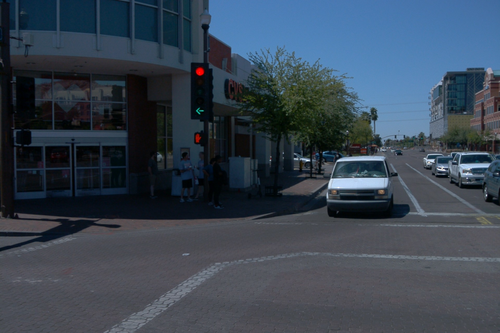}&
        \includegraphics[width=\sz\linewidth]{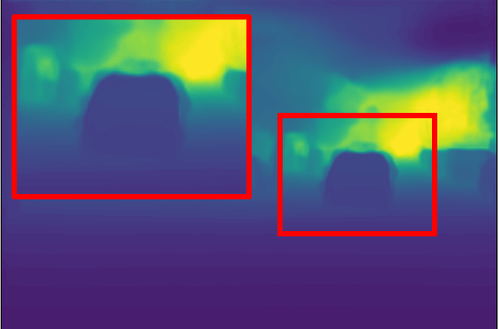}&
        \includegraphics[width=\sz\linewidth]{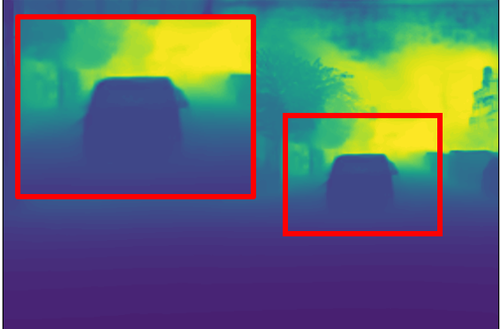}\\
        
        \includegraphics[width=\sz\linewidth]{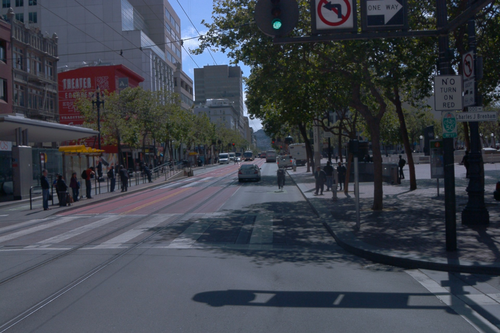}&
        \includegraphics[width=\sz\linewidth]{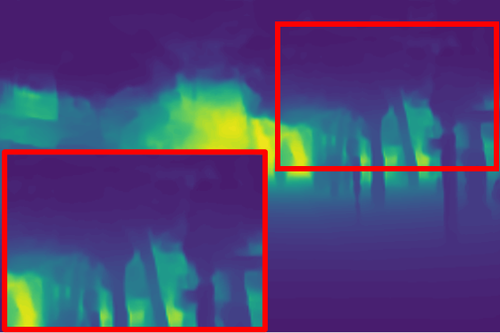}&
        \includegraphics[width=\sz\linewidth]{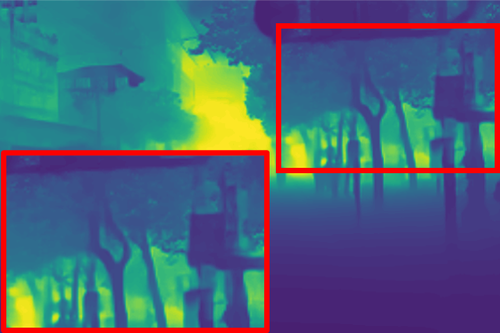}\\
        
        \includegraphics[width=\sz\linewidth]{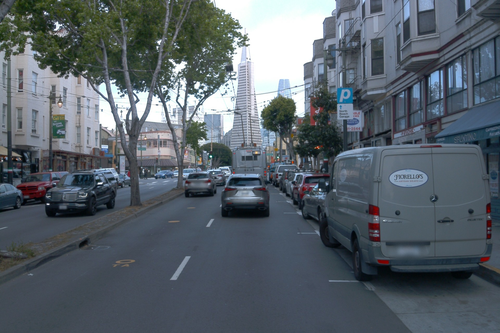}&
        \includegraphics[width=\sz\linewidth]{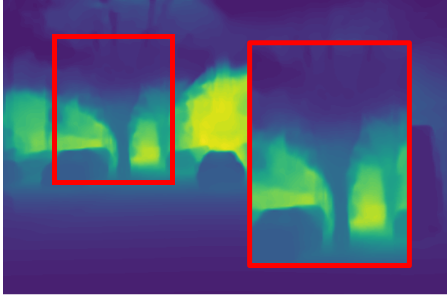}&
        \includegraphics[width=\sz\linewidth]{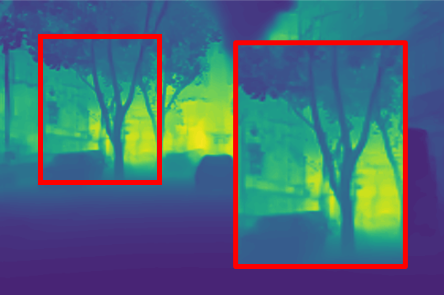}\\
    \end{tabular} }
    \caption{\textbf{Qualitative results on the Waymo~\cite{sun_scalability_2020} dataset, focusing on close-up details.} We show the qualitative close-ups of the performance of the vanilla-trained AdaBins~\cite{bhat_adabins_2021} vs our proposed \textit{NeRFmented} AdaBins, demonstrating the capability of our method to recover fine grain details in the prediction that the baseline is not able to predict. Color scale: 0 (purple) to 80 meters (yellow).}
    \label{fig:waymo_qualitative}
\end{figure}

\subsection{Evaluation of zero-shot dataset transfer to Waymo}
Evaluation on a different dataset provides a good insight into the models' generalization capabilities. The models are strictly trained on KITTI and KITTI-based \textit{NeRFmented} images in this comparison. \cref{tab:waymo_datset_augmentation} shows that the \textit{NeRFmented} runs deliver massive uplifts during zero-shot evaluation on the Waymo dataset, beating baseline performance in a large majority of metrics for every single augmentation strategy. Note that the \textit{NeRFmented} results shown in \cref{tab:waymo_datset_augmentation} correspond to the augmentation regime \textit{(iv)}, vertical translation. We furthermore note that \textit{NeRFmentation} performs better when classic augmentations are turned off. 

\begin{figure}[tb]
        \hspace{-0.4cm}
    \centering
    \scriptsize
    \setlength{\tabcolsep}{1pt}
	\renewcommand{\arraystretch}{0.8}
	\newcommand{\sz}{0.44}
	\newcommand{\sh}{1.63cm}
    \begin{tabular}{ccc}
        \rotatebox{90}{\hspace{15pt}\tiny RGB} & \includegraphics[width=\sz\linewidth]{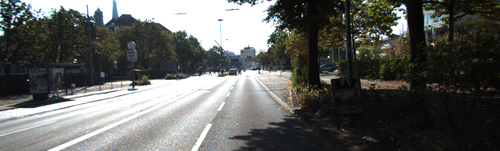}&
        \includegraphics[width=\sz\linewidth]{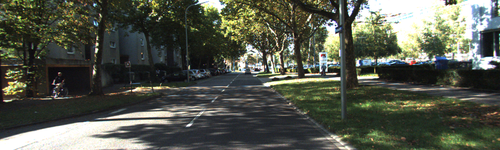} \\
         \rotatebox{90}{\hspace{4pt}\tiny DepthFormer} & \includegraphics[width=\sz\linewidth]{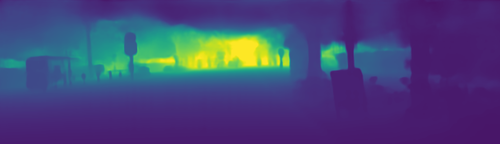}&
        \includegraphics[width=\sz\linewidth]{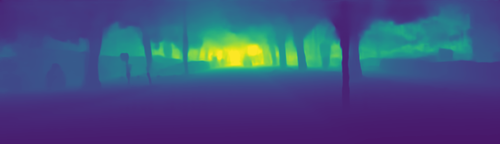} \\
        \rotatebox{90}{\hspace{0pt}\tiny \shortstack{\textbf{NeRFmented} \\ DepthFormer}} & \includegraphics[width=\sz\linewidth]{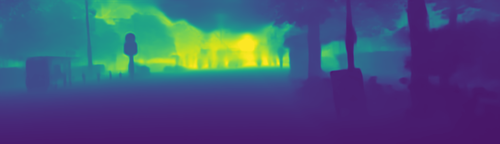}&
        \includegraphics[width=\sz\linewidth]{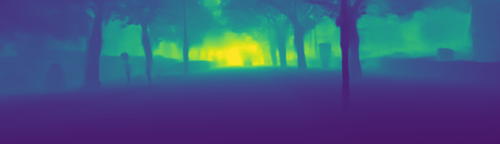} \\
    \end{tabular} 
    \caption{\textbf{Qualitative results on the KITTI~\cite{geiger_kitti_2013} dataset.} We show the qualitative depth predictions of the vanilla-trained DepthFormer~\cite{li2023depthformer} vs our proposed \textit{NeRFmented} DepthFormer, demonstrating the capability of our method to recover fine-grain details in the prediction that the baseline is not able to predict. Color scale: 0 (purple) to 80 meters (yellow).}
    \label{fig:kitti_qualitative}
\end{figure} 

As shown in \cref{tab:waymo_dataset_strategies}, out of all strategies, the vertical translation strategy performs best. We believe that it forces the MDE to overfit less to a specific camera height, allowing better performance for the different car heights in Waymo. Generally, the largest difference in performance lies between \textit{NeRFmented} and non-\textit{NeRFmented} runs, whereas it doesn't seem to matter as much which \textit{NeRFmentation} strategy is picked. We hypothesize that since all of the \textit{NeRFmentation} runs offer depth completion, it is this additional dense supervision that helps these models perform. However, we also note that in most cases vertical translation outperforms reconstructing the RGB-D pairs. Therefore, we argue that it is not only the depth completion offered by NeRFs that increases model robustness but also the additional viewpoints. The augmentation's quantitative success is mirrored by the qualitative results displayed in \cref{fig:waymo_qualitative}. An increase in fine detail prediction in street signs, stoplights, streetlamps, and foliage is clearly visible. Car outlines become sharper and background building details are much improved as well.

\begin{table*}[!tb]
 \centering

\caption{\textbf{Comparison of data augmentation methods on the KITTI Eigen~\cite{geiger_kitti_2013, eigen2014depth} test split.} \small We showcase the performance of AdaBins~\cite{bhat_adabins_2021}, DepthFormer~\cite{li2023depthformer} and BinsFormer~\cite{li2022binsformer} models trained on the KITTI Eigen~\cite{geiger_kitti_2013, eigen2014depth} train split and compare different augmentation methods to our \textit{NeRFmentation} method. CAug indicates whether classic augmentations were used during training. For \textit{NeRFmentation}, the NeRFm. Vert. $\pm 30cm$ strategy is shown. The best results are in \colorbox{colorFst}{\textbf{bold}}. The second best results are \colorbox{colorSnd}{\underline{underlined}}, and the \colorbox{colorTrd}{baseline} is highlighted.}  
\scalebox{0.8}{
    \begin{tabular}{clccccccc} 
        \toprule
     
        & Method & $\delta_{1}\uparrow$ & $\delta_{2}\uparrow$    & $\delta_{3}\uparrow$ & REL $\downarrow$ & SQ. REL $\downarrow$ & RMS $\downarrow$  & RMS$_{LOG}$ $\downarrow$ \\
        \midrule 
        
        \multirow{4}{*}{\rotatebox{90}{CAug. \cmark}} & \textbf{Adabins} \cite{bhat_adabins_2021} & \fs{0.964} & \fs{0.995} & \rd{0.999} & \fs{0.058} & \fs{0.190} & \fs{2.360} & \fs{0.088} \\
        & \textcolor{orange}{\spacedbullet}FDA~\cite{Yang_2020_fda_DA}      & \nd{0.962} & \fs{0.995} & 0.999 & \nd{0.059} & \nd{0.200} & \nd{2.385} & \nd{0.090} \\
                                    
        & \textcolor{orange}{\spacedbullet}Style Transfer~\cite{atapour-abarghouei_real-time_2018} & 0.960 & \nd{0.994} & 0.999 & 0.060 & 0.211 & 2.482 & 0.093 \\
        & \textcolor{cyan}{\spacedbullet}NeRFmentation & 0.960 & \nd{0.994} & 0.999 & 0.062 & 0.210 & 2.427 & 0.094 \\
        \midrule
        \multirow{4}{*}{\rotatebox{90}{CAug. \xmark}} & \textbf{Adabins} \cite{bhat_adabins_2021} & \fs{0.962} & \fs{0.995} & \fs{0.999} & \fs{0.060} & \fs{0.204} & \fs{2.449} & \fs{0.092} \\
        & \textcolor{orange}{\spacedbullet}FDA~\cite{Yang_2020_fda_DA}  & 0.956 & \nd{0.994} & \fs{0.999} & \nd{0.062} & 0.219 & 2.542 & 0.096 \\
        & \textcolor{orange}{\spacedbullet}Style Transfer~\cite{atapour-abarghouei_real-time_2018}  & 0.948 & 0.992 & \nd{0.998} & 0.066 & 0.253 & 2.798 & 0.103 \\
        & \textcolor{cyan}{\spacedbullet}NeRFmentation & \nd{0.958} & \nd{0.994} & \fs{0.999} & \nd{0.062} & \nd{0.215} & \nd{2.534} & \nd{0.095} \\
        \midrule
        \multirow{4}{*}{\rotatebox{90}{CAug. \cmark}} & \textbf{DepthFormer} \cite{li2023depthformer} & \nd{0.975} & \rd{0.997} & \rd{0.999} & \rd{0.052} & \rd{0.158} & \rd{2.143} & \nd{0.079} \\
        & \textcolor{orange}{\spacedbullet}FDA~\cite{Yang_2020_fda_DA}                       & 0.973 & 0.997 & 0.999 & 0.052 & \nd{0.154} & \nd{2.123} & \nd{0.079} \\
       & \textcolor{orange}{\spacedbullet}Style Transfer~\cite{atapour-abarghouei_real-time_2018}           & 0.972 & 0.997 & 0.999 & 0.052 & 0.160 & 2.192 & 0.082 \\
        &\textcolor{cyan}{\spacedbullet}NeRFmentation                & \fs{0.976} & 0.997 & 0.999 & 0.052 & \fs{0.150} & \fs{2.088} & \fs{0.078} \\
        \midrule
                \multirow{4}{*}{\rotatebox{90}{CAug. \xmark}} &\textbf{DepthFormer} \cite{li2023depthformer} & \nd{0.972} & \fs{0.997} & \nd{0.999} & \nd{0.054} & \nd{0.159} & \nd{2.172} & \nd{0.082} \\
        & \textcolor{orange}{\spacedbullet}FDA~\cite{Yang_2020_fda_DA}  & 0.971 & \fs{0.997} & \nd{0.999} & \nd{0.054} & 0.164 & 2.213 & \nd{0.082} \\
        & \textcolor{orange}{\spacedbullet}Style Transfer~\cite{atapour-abarghouei_real-time_2018}   & 0.965 & \nd{0.995} & \nd{0.999} & 0.057 & 0.194 & 2.525 & 0.089 \\
        &\textcolor{cyan}{\spacedbullet}NeRFmentation & \fs{0.974} & \fs{0.997} & \fs{1.000} & \fs{0.053} & \fs{0.156} & \fs{2.145} & \fs{0.080} \\
        \midrule
        \multirow{4}{*}{\rotatebox{90}{CAug. \cmark}} & \textbf{BinsFormer}  \cite{li2022binsformer}       & \fs{0.974} & \rd{0.997} & \rd{0.999} & \fs{0.052} & \fs{0.151} & \fs{2.098} & \fs{0.079} \\
        & \textcolor{orange}{\spacedbullet}FDA~\cite{Yang_2020_fda_DA}  & \fs{0.974} & 0.997 & 0.999 & \nd{0.053} & \nd{0.155} & \nd{2.117} & \nd{0.080} \\
        & \textcolor{orange}{\spacedbullet}Style Transfer~\cite{atapour-abarghouei_real-time_2018}            & 0.971 & 0.997 & 0.999 & 0.054 & 0.165 & 2.226 & 0.083 \\
        & \textcolor{cyan}{\spacedbullet}NeRFmentation                & \nd{0.973} & 0.997 & 0.999 & 0.054 & 0.156 & 2.124 & 0.081 \\
        \midrule
        \multirow{4}{*}{\rotatebox{90}{CAug. \xmark}} & \textbf{BinsFormer} \cite{li2022binsformer}  & \fs{0.973} & \fs{0.997} & 0.999 & \fs{0.054} & 0.159 & {2.191} & \fs{0.081} \\
        & \textcolor{orange}{\spacedbullet}FDA~\cite{Yang_2020_fda_DA} & \fs{0.973} & \fs{0.997} & 0.999 & \fs{0.054} & \fs{0.159} & \fs{2.178} & \fs{0.081} \\
        & \textcolor{orange}{\spacedbullet}Style Transfer~\cite{atapour-abarghouei_real-time_2018}             & {0.968} & \nd{0.996} & 0.999 & {0.058} & {0.184} & 2.411 & 0.088 \\
        & \textcolor{cyan}{\spacedbullet}NeRFmentation                & \nd{0.972} & \fs{0.997} & 0.999 & \nd{0.055} & \nd{0.161} & \nd{2.187} & \nd{0.082} \\
        \bottomrule
    \end{tabular}}
  \label{tab:kitti_dataset_augmentation}
\end{table*}

\subsection{Evaluation on KITTI}
\label{sec:training_kitti}
When evaluating \textit{NeRFmentation} on the KITTI Eigen \cite{geiger_kitti_2013, eigen2014depth} test split, large qualitative improvements can be seen (\cref{fig:kitti_qualitative}). As a result of the densification of the depth maps, the MDE architecture gains much-needed supervision here. This effect is apparent by the increased clarity in the upper parts of the predicted depth maps. We also notice increased detail in the entire image for objects such as street signs, poles, and trees.   

The quantitative results on the KITTI Eigen test split are displayed in \cref{tab:kitti_dataset_augmentation}. For all \textit{NeRFmentation} results shown, we choose the vertical translation strategy for consistency. \textit{NeRFmented} models perform slightly worse than baseline on average. Better small detail prediction or improved performance in the upper part of the image are not measured as the LiDAR-generated depth maps do not offer any ground truth information in the upper parts of the image (see \cref{fig:nerf_reconstruction}). Therefore, areas, where NeRFmentation provides denser depth information, are not captured by the evaluation metrics on this dataset. To eliminate this factor in evaluation, we also evaluate our models on our own densified (\emph{NeRFmented}) KITTI test set. Here, \cref{tab:nerf_kitti_dataset} of the main paper shows NeRFmentation significantly outperforming the baseline. In \cref{tab:waymo_dataset_strategies} we notice that the same \textit{NeRFmentation} strategies which work well on Waymo continue to do so for KITTI. We also note that contrary to the case when evaluated on Waymo, classic augmentations increase the performance when using \textit{NeRFmentation}. Interestingly some models benefit more from the additional viewpoints and dense depth information, while others seem to be more sensitive to the slight domain shift introduced by the synthetic data.

\begin{table*}[!tb]
 \centering

    \caption{\textbf{Comparison of performances on the NeRF-generated KITTI dataset.} \small This table shows the performance of baseline and \textit{NeRFmented} KITTI trained models evaluated on a NeRF rendered subset of the KITTI Eigen~\cite{eigen2014depth, geiger_kitti_2013} test split scenes. \textit{NeRFmentation} improves upon baseline performance independent of the architecture used across all strategies. No classic augmentation is used. The best results are in \colorbox{colorFst}{\textbf{bold}}. The second best results are \colorbox{colorSnd}{\underline{underlined}}, and the \colorbox{colorTrd}{baseline} is highlighted.} 
\scalebox{0.8}{
    \begin{tabular}{clccccccc} 
        \toprule
     
        & Augmentation & $\delta_{1}\uparrow$ & $\delta_{2}\uparrow$    & $\delta_{3}\uparrow$ & REL $\downarrow$ & SQ. REL $\downarrow$ & RMS $\downarrow$  & RMS$_{LOG}$ $\downarrow$ \\
        \midrule
        \multirow{7}{*}{\rotatebox{90}{\textbf{Adabins} \cite{bhat_adabins_2021}}}&\textcolor{orange}{\spacedbullet}No Augm. & 0.776 & 0.964 & 0.989 & 0.182 & 1.011 & 4.104 & 0.205 \\
        &\textcolor{orange}{\spacedbullet}Classic & \rd{0.761} & \rd{0.961} & \rd{0.988} & \rd{0.183} & \rd{0.987} & \rd{4.248} & \rd{0.211} \\
        &\textcolor{cyan}{\spacedbullet}NeRFm. Reconstruction & \fs{0.909} & \fs{0.986} & \fs{0.995} & \nd{0.126} & \nd{0.639} & \nd{3.380} & \fs{0.137} \\
        &\textcolor{cyan}{\spacedbullet}NeRFm. Interpolation & \nd{0.908} & \fs{0.986} & \nd{0.994} & \fs{0.121} & \fs{0.615} & \fs{3.377} & \fs{0.137} \\
        &\textcolor{cyan}{\spacedbullet}NeRFm. Angled $\pm 3^{\circ}$ & 0.898 & \nd{0.985} & \nd{0.994} & 0.140 & 0.821 & 3.488 & 0.142 \\
        &\textcolor{cyan}{\spacedbullet}NeRFm. Horiz. $\pm 30cm$ & 0.902 & \nd{0.985} & \fs{0.995} & 0.140 & 0.823 & 3.454 & 0.141 \\
        &\textcolor{cyan}{\spacedbullet}NeRFm. Vert. $\pm 30cm$ & 0.850 & 0.984 & \fs{0.995} & 0.170 & 0.852 & 3.646 & 0.172 \\
        \midrule
        \multirow{7}{*}{\rotatebox{90}{\textbf{DepthFormer} \cite{li2023depthformer}}}&\textcolor{orange}{\spacedbullet}No Augm. & 0.801 & 0.977 & 0.992 & 0.168 & 0.805 & 3.783 & 0.189 \\
        &\textcolor{orange}{\spacedbullet}Classic & \rd{0.791} & \rd{0.978} & \rd{0.992} & \rd{0.171} & \rd{0.840} & \rd{3.804} & \rd{0.193} \\
        &\textcolor{cyan}{\spacedbullet}NeRFm. Reconstruction & \nd{0.922} & \fs{0.990} & \fs{0.997} & \nd{0.109} & \nd{0.492} & \fs{3.073} & \fs{0.125} \\
        &\textcolor{cyan}{\spacedbullet}NeRFm. Interpolation & \fs{0.923} & \nd{0.989} & \nd{0.996} & \fs{0.104} & \fs{0.455} & \fs{3.073} & \fs{0.125} \\
        &\textcolor{cyan}{\spacedbullet}NeRFm. Angled $\pm 3^{\circ}$ & 0.919 & \nd{0.989} & \nd{0.996} & 0.112 & 0.521 & \nd{3.119} & \nd{0.127} \\
        &\textcolor{cyan}{\spacedbullet}NeRFm. Horiz. $\pm 30cm$ & \nd{0.922} & 0.989 & \nd{0.996} & 0.114 & 0.532 & 3.186 & 0.128 \\
        &\textcolor{cyan}{\spacedbullet}NeRFm. Vert. $\pm 30cm$ & 0.905 & \fs{0.990} & \nd{0.996} & 0.143 & 0.583 & 3.350 & 0.153 \\
        \midrule
        \multirow{7}{*}{\rotatebox{90}{\textbf{BinsFormer} \cite{li2022binsformer}}}&\textcolor{orange}{\spacedbullet}No Augm. & 0.799 & 0.979 & \nd{0.992} & 0.172 & 0.846 & 3.744 & 0.190 \\
        &\textcolor{orange}{\spacedbullet}Classic & \rd{0.807} & \rd{0.979} & \nd{0.992} & \rd{0.168} & \rd{0.831} & \rd{3.743} & 0\rd{.189} \\
        &\textcolor{cyan}{\spacedbullet}NeRFm. Reconstruction & 0.920 & \fs{0.989} & \fs{0.996} & \nd{0.109} & \nd{0.476} & 3.066 & 0.127 \\
        &\textcolor{cyan}{\spacedbullet}NeRFm. Interpolation & \nd{0.925} & \fs{0.989} & \fs{0.996} & \fs{0.106} & \fs{0.454} & \fs{3.010} & \fs{0.124} \\
        &\textcolor{cyan}{\spacedbullet}NeRFm. Angled $\pm 3^{\circ}$ & 0.921 & 0.988 & \fs{0.996} & 0.113 & 0.514 & \nd{3.061} & 0.127 \\
        &\textcolor{cyan}{\spacedbullet}NeRFm. Horiz. $\pm 30cm$ & \fs{0.926} & \fs{0.990} & \fs{0.996} & 0.111 & 0.509 & 3.101 & \nd{0.125} \\
        &\textcolor{cyan}{\spacedbullet}NeRFm. Vert. $\pm 30cm$ & 0.897 & 0.988 & \fs{0.996} & 0.145 & 0.585 & 3.334 & 0.158 \\
        \bottomrule
    \end{tabular}}
  \label{tab:nerf_kitti_dataset}
\end{table*}

\subsection{Ablation on NeRF-generated KITTI dataset}
We believe our evaluation on KITTI and Waymo is limited by their sparse ground truth and limited number of viewing directions. To fully capture the improved generalization capabilities by \textit{NeRFmentation}, we design our own synthetic test set with dense ground truth based on KITTI. We train NeRFs on 111 sub-scenes from the KITTI Eigen test split. Then, we filter out subpar scene representations by setting the LPIPS threshold to 0.3 and abs. rel. threshold to 0.05, leaving 15 scenes for rendering. Finally, we render novel views whose poses are randomly translated and rotated along each axis. This yields a test set with \textbf{1,218} dense RGB-D pairs.

In \cref{tab:nerf_kitti_dataset} we see significant improvement in all metrics for all evaluated depth architectures using \textit{NeRFmented} data. This demonstrates that a non-ideal dense test set can better capture the value that is provided to the depth architectures by the \textit{NeRFmented} data.

\section{Discussion}

In this paper, we explore the idea of improving the accuracy and generalizability of monocular depth estimators by synthesizing new data from existing data via neural radiance fields.
Our evaluation shows that our method \textit{NeRFmentation} greatly improved performance on the unseen Waymo and densified and perturbed KITTI dataset while retaining solid results on the base KITTI dataset. 
As classic augmentations hurt the performance when using \textit{NeRFmentation} for out-of-distribution tasks, we argue that our presented method, \textit{NeRFmentation}, should be considered a replacement for classic augmentations.

\PAR{Limitations.}
\textit{NeRFmentation} improves monocular depth estimators by addressing sparse depth maps, limited viewing angles, and scene diversity issues. However, it shows no improvement for datasets already rich in these aspects, such as NYU-Depth V2 \cite{silberman2012indoor}. This indoor dataset, captured with Microsoft Kinect, provides dense depth maps from varied viewpoints. We hypothesize that these characteristics limit \textit{NeRFmentation}'s potential for improvement (see supplementary material for evidence).

\PAR{Future work.}
NeRFs struggle with unobserved regions during training, producing noise in novel view renderings that unfairly penalize the MDE. A masking strategy for observed areas is provided in the supplementary material. Additionally, the static world assumption incorrectly reconstructs dynamic objects like cars and people. Dynamic NeRF methods \cite{pumarola2020dnerf, song2023nerfplayer, park2023temporal} could potentially address this limitation, enabling higher fidelity data synthesis.

\newpage
\PAR{Acknowledgements.} One of the authors is funded by 2023/00027/001 under the
``Campaign financed by the European Union Next Generation - Recovery Plan - Competent Ministry - Spanish Government''.


\clearpage
\begin{center}    
    \textbf{\Large{Supplementary Material}}
\end{center}
\counterwithin{figure}{section}
\counterwithin{table}{section}

\appendix %

\begin{itemize}
    \item[$\diamond$] \cref{sec:Evaluation} provides details about the evaluation metrics of the monocular depth estimation algorithms.
    \item[$\diamond$] \cref{sec:supp_saturation} shows an additional ablation about the nerfmented views saturation ablation on the KITTI and Waymo datasets.
    \item[$\diamond$] \cref{sec:supp_qualitative_kitti} gives additional qualitative results with state-of-the-art methods on the KITTI and Waymo dataset. Also, it compares NeRFmentation to DepthAnything.
    \item[$\diamond$] \cref{sec:angled} introduces a qualitative comparison between the reconstructed, interpolated, and angled augmentation methods.
    \item[$\diamond$] \cref{sec:supp_qualitative_nerf} shows scene meshes of the NeRFs trained on the KITTI dataset.
    \item[$\diamond$] \cref{sec:supp_nyu} presents a discussion about the proposed method applied to the NYU-Depth V2 dataset. The section discusses:
    \begin{itemize}
        \item[$\circ$] Training on NYU-Depth V2.
        \item[$\circ$] Zero-shot data transfer to Replica Dataset.
        \item[$\circ$] Ablation on NeRF-generated NYU-Depth V2 dataset.
    \end{itemize}
    \item[$\diamond$] Finally, \cref{sec:supp_masking} introduces the future works of NeRFmentation.
\end{itemize}

\section{Evaluation Metrics for MDE}
\label{sec:Evaluation}
We utilize the MDE evaluation metrics used in \cite{bauer_nvs-monodepth_2021}. The metrics are defined as follows, given prediction $\hat{y}_i$ and ground truth value $y_{i}$:\\
Relative Error (REL): $\frac{1}{N}\sum^{N}_{i=1}\frac{\lvert y_i-\hat{y}_i \rvert}{y_{i}}$ \\
Squared Relative Error (SQ. REL): $\frac{1}{N}\sum^{N}_{i=1}\frac{\lvert y_i-\hat{y}_i \rvert^2}{y_{i}}$ \\
Root Mean Squared Error (RMSE): $\sqrt{\frac{1}{N}\sum^{N}_{i=1}(y_{i}-\hat{y}_i)^2}$ \\
Log RMSE: $\sqrt{\frac{1}{N}\sum^{N}_{i=1}(\log y_{i}-\log \hat{y}_i)^2}$ \\
Threshold Accuracy ($\delta_{j}$): $\max(\frac{y_i}{\hat{y}_i},\frac{\hat{y}_i}{y_{i}}) = \delta < 1.25^{j}$ 

\section{NeRFmented views saturation ablation on the KITTI and Waymo dataset}
\label{sec:supp_saturation}

We investigate the effect of adding increasing amounts of NeRFmented views to the base KITTI dataset when using the interpolation strategy with Depthformer in \cref{tab:saturation}. We render views at 10 interpolated viewpoints between each pair of training poses. We then chose a random subset ranging between 1k to 50k NeRFmented images to add to the base 23k image KITTI dataset. Hyperparameters and iterations were kept constant for all runs. On the KITTI eigen test split, we see a small decrease in performance as more NeRFmented views are added. On the Waymo test set, we observe increasing performance as the added view count reaches 5-10k images and then a decrease in performance as the novel views become the majority of the training dataset. It is very interesting to note, that only 1k additional views (+4.3\%) already lead to great gains on Waymo. 

\begin{table}
\footnotesize
\setlength{\tabcolsep}{3pt}
    \centering
    \caption{\textbf{Comparison of NeRFmented view count evaluated on KITTI and Waymo.} We observe the effects of adding an increasing number of interpolated views to the base KITTI dataset. These results are for DepthFormer. Classic augmentation is used for 0k views and turned off for the others. We see a slight decrease in performance on KITTI as more views are added. The best results on Waymo are between 5k and 10k additional views. The best results are in \textbf{bold} dark green. The second best results are \underline{underlined} light green.}  
    \begin{tabular}{clccccccc}
        \toprule
        \small
        & \# Extra Views & $\delta_{1}\uparrow$ & $\delta_{2}\uparrow$    & $\delta_{3}\uparrow$ & REL $\downarrow$ & SQ. REL $\downarrow$ & RMS $\downarrow$  & RMS$_{LOG}$ $\downarrow$ \\
        \midrule
        \multirow{6}{*}{\rotatebox[origin=c]{90}{\textbf{KITTI} \cite{geiger_kitti_2013}}}&\textcolor{orange}{\spacedbullet}0k & \fs{0.975} & 0.997 & 0.999 & \fs{0.052} & \fs{0.158} & \fs{2.143} & \fs{0.079} \\
        &\textcolor{cyan}{\spacedbullet}1k & \nd{0.973} & 0.997 & 0.999 & \nd{0.053} & \fs{0.158} & 2.174 & \nd{0.081} \\
        &\textcolor{cyan}{\spacedbullet}5k & \nd{0.973} & 0.997 & \fs{1.000} & 0.054 & 0.162 & 2.185 & \nd{0.081} \\
        &\textcolor{cyan}{\spacedbullet}10k & \nd{0.973} & 0.997 & \fs{1.000} & 0.054 & \nd{0.160} & \nd{2.163} & \nd{0.081} \\
        &\textcolor{cyan}{\spacedbullet}20k & \nd{0.973} & 0.997 & 0.999 & \nd{0.053} & \nd{0.160} & 2.183 & \nd{0.081} \\
        &\textcolor{cyan}{\spacedbullet}50k & 0.972 & 0.997 & 0.999 & 0.055 & 0.165 & 2.226 & 0.083 \\
        \midrule
        \multirow{6}{*}{\rotatebox[origin=c]{90}{\textbf{Waymo} \cite{sun_scalability_2020}}}&\textcolor{orange}{\spacedbullet}0k & 0.545 & 0.885 & 0.958 & 0.217 & 1.949 & 7.911 & 0.304 \\
        &\textcolor{cyan}{\spacedbullet}1k & 0.690 & 0.922 & 0.974 & 0.177 & 1.522 & 6.976 & 0.251 \\
        &\textcolor{cyan}{\spacedbullet}5k & 0.718 & \fs{0.933} & \fs{0.978} & 0.172 & \fs{1.441} & \fs{6.570} &  \nd{0.238} \\
        &\textcolor{cyan}{\spacedbullet}10k & \fs{0.733} &  \nd{0.931} &  \nd{0.977} & \fs{0.167} & 1.443 &  \nd{6.683} & \fs{0.234} \\
        &\textcolor{cyan}{\spacedbullet}20k &  \nd{0.722} & 0.928 & 0.976 &  \nd{0.171} &  \nd{1.473} & 6.764 & 0.240 \\
        &\textcolor{cyan}{\spacedbullet}50k & 0.701 & 0.918 & 0.972 & 0.176 & 1.578 & 7.041 & 0.250 \\
        \bottomrule
    \end{tabular}  
    \label{tab:saturation}
\end{table}

\section{Qualitative comparison with state-of-the-art methods on the KITTI and Waymo dataset}
\label{sec:supp_qualitative_kitti}

\begin{figure}[htb!]
    \centering
    \tiny
    \setlength{\tabcolsep}{1pt}
    \renewcommand{\arraystretch}{0.8}
    \newcommand{\sz}{0.22}
    \hspace*{-25pt}
    \begin{tabular}{ccccc}
        \rotatebox{90}{\hspace{16pt} RGB} &
        \includegraphics[width=\sz\linewidth]{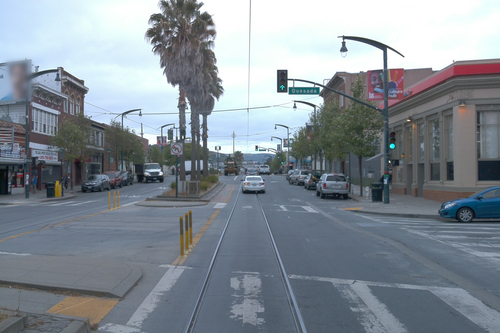} &
        \includegraphics[width=\sz\linewidth]{figures/QualitativeFigures/RGB07.png} &
        \includegraphics[width=\sz\linewidth]{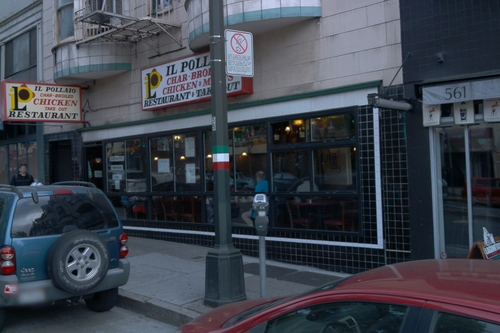} &
        \includegraphics[width=\sz\linewidth]{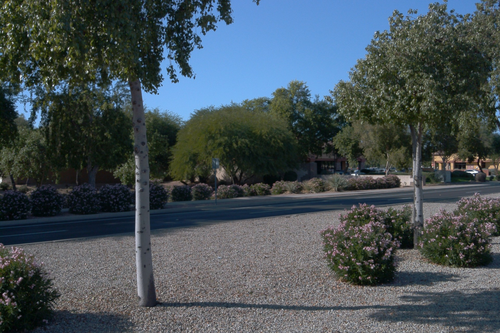}
\\
        \rotatebox{90}{\hspace{8pt}AdaBins~\cite{bhat_adabins_2021}} &
        \includegraphics[width=\sz\linewidth]{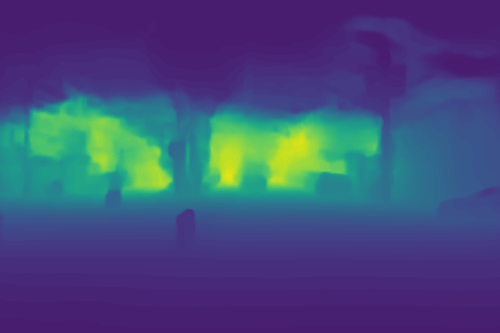} &
        \includegraphics[width=\sz\linewidth]{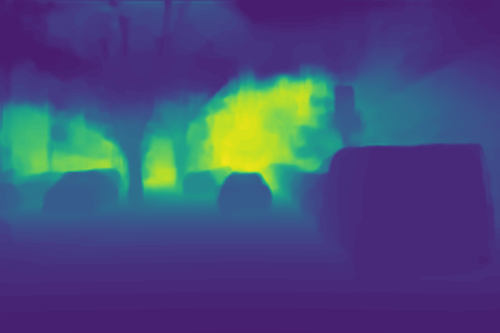} &
        \includegraphics[width=\sz\linewidth]{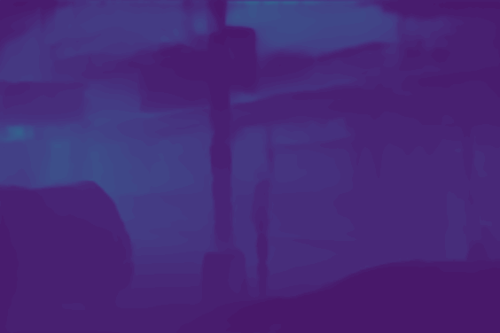} &
        \includegraphics[width=\sz\linewidth]{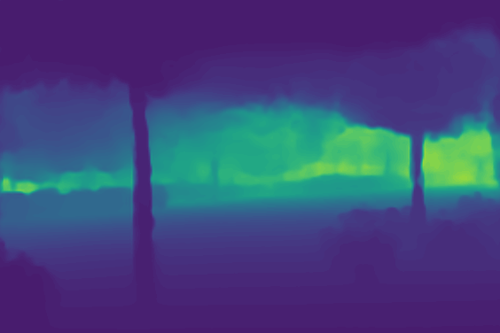}
\\
        \rotatebox{90}{\hspace{4pt}\shortstack{\textbf{NeRFmented}\\AdaBins~\cite{bhat_adabins_2021}}} &
        \includegraphics[width=\sz\linewidth]{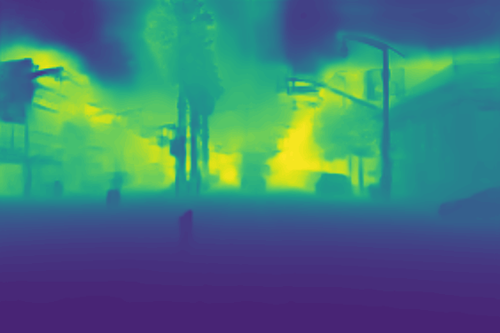} &
        \includegraphics[width=\sz\linewidth]{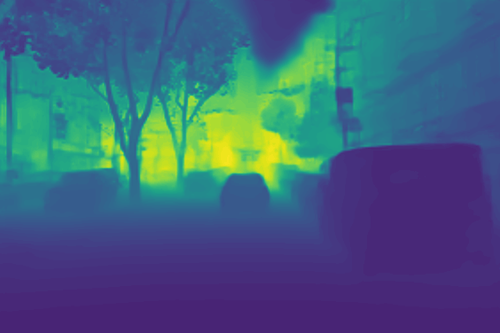} &
        \includegraphics[width=\sz\linewidth]{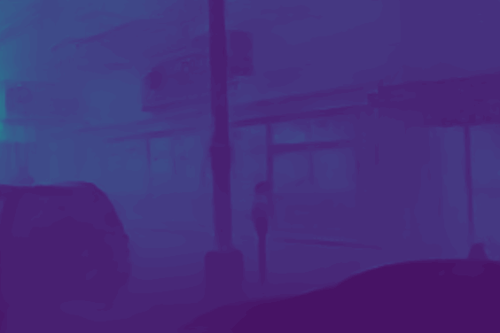} &
        \includegraphics[width=\sz\linewidth]{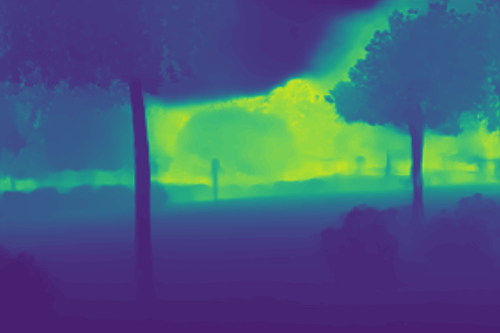}
\\
        \rotatebox{90}{\hspace{0pt} DepthFormer~\cite{li2023depthformer}} &
        \includegraphics[width=\sz\linewidth]{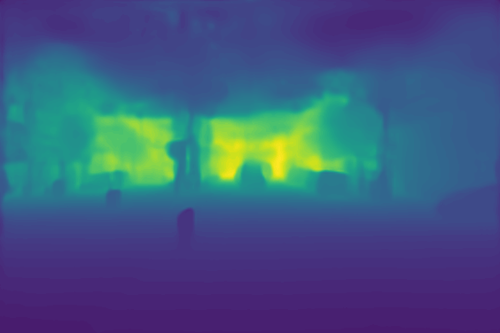} &
        \includegraphics[width=\sz\linewidth]{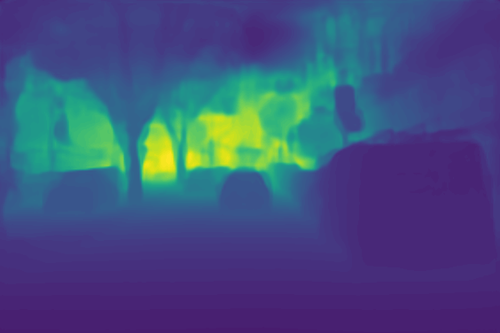} &
        \includegraphics[width=\sz\linewidth]{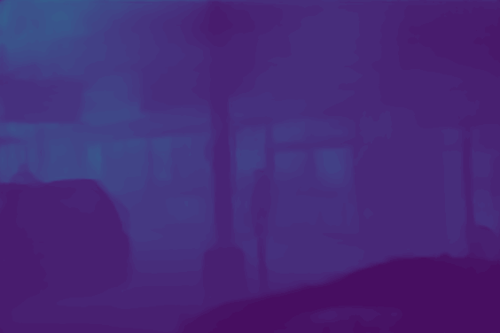} &
        \includegraphics[width=\sz\linewidth]{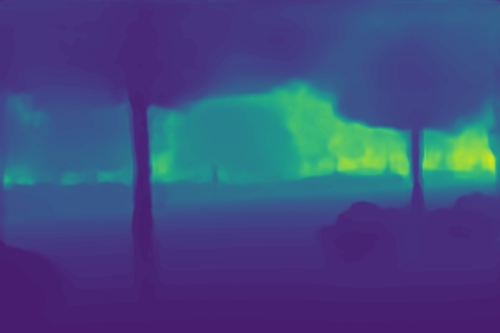}
\\      
        \rotatebox{90}{\hspace{0pt}\shortstack{\textbf{NeRFmented}\\DepthFormer~\cite{li2023depthformer}}} &
        \includegraphics[width=\sz\linewidth]{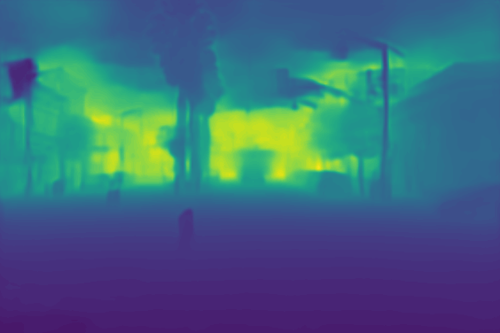} &
        \includegraphics[width=\sz\linewidth]{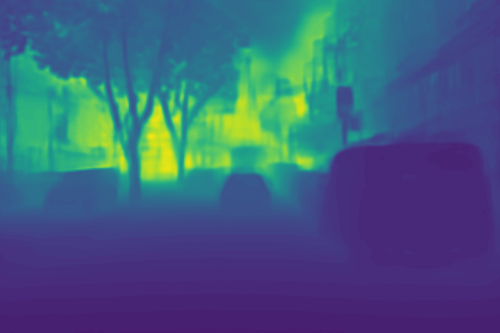} &
        \includegraphics[width=\sz\linewidth]{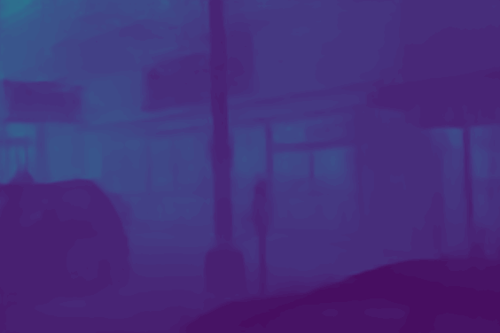} &
        \includegraphics[width=\sz\linewidth]{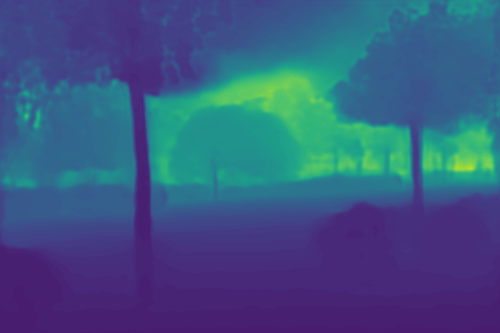}
\\
        \rotatebox{90}{\hspace{2pt} BinsFormer~\cite{li2022binsformer}} &
        \includegraphics[width=\sz\linewidth]{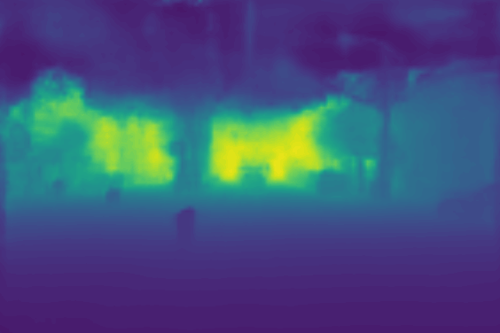} &
        \includegraphics[width=\sz\linewidth]{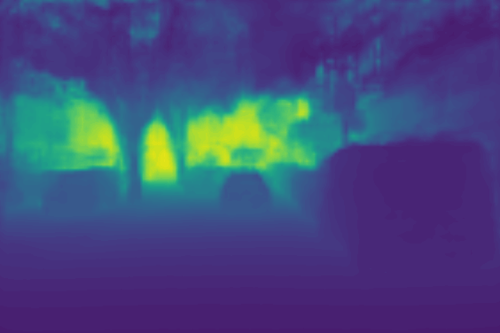} &
        \includegraphics[width=\sz\linewidth]{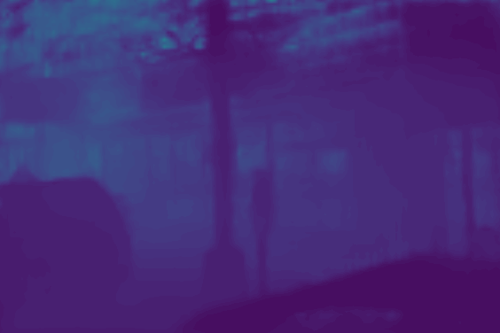} &
        \includegraphics[width=\sz\linewidth]{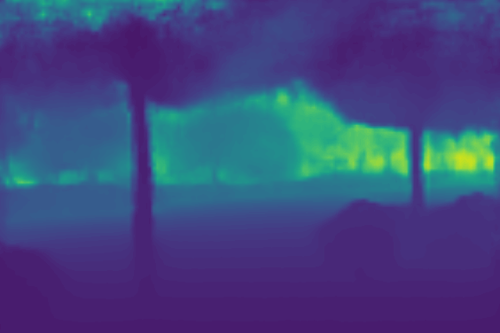}
\\
        \rotatebox{90}{\hspace{2pt} \shortstack{\textbf{NeRFmented}\\BinsFormer~\cite{li2022binsformer}}} &
        \includegraphics[width=\sz\linewidth]{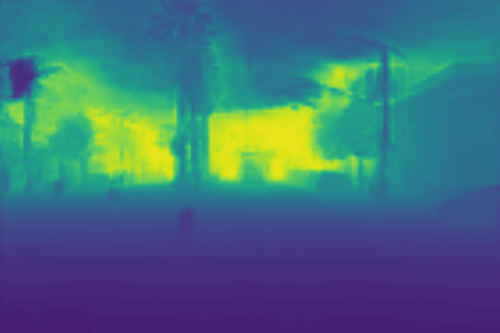} &
        \includegraphics[width=\sz\linewidth]{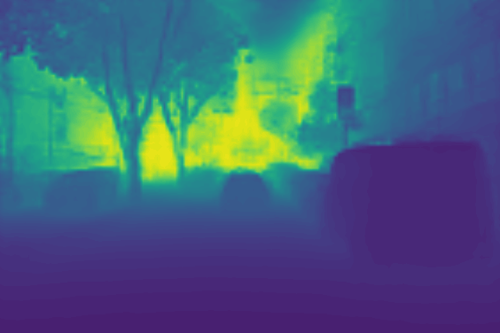} &
        \includegraphics[width=\sz\linewidth]{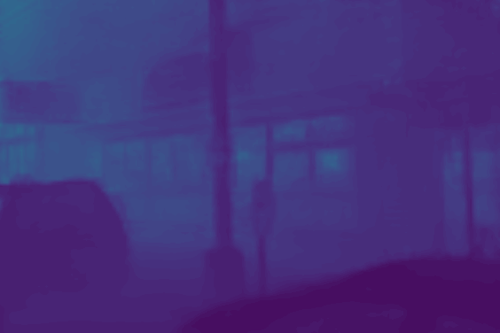} &
        \includegraphics[width=\sz\linewidth]{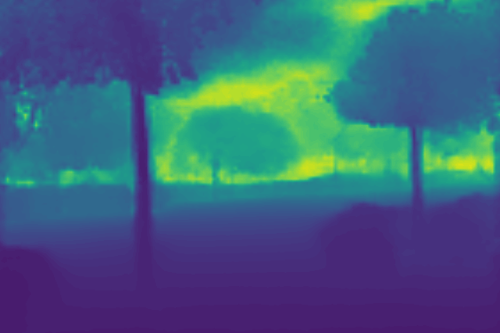}
\\
    \end{tabular}
    \caption{\textbf{Qualitative predictions on the Waymo \cite{sun_scalability_2020} dataset.} We show the performance of NeRFmentation (Ours) compared to AdaBins~\cite{bhat_adabins_2021}, DepthFormer~\cite{li2023depthformer} and BinsFormer~\cite{li2022binsformer}. The color scale goes from 0 (purple) to 80 meters (yellow). Best viewed on a monitor zoomed in.
    }
    \label{fig:qualitative_wamosup}
\end{figure}

\begin{sidewaysfigure}[htb!]
    \centering
    \tiny
    \setlength{\tabcolsep}{1pt}
    \renewcommand{\arraystretch}{0.8}
    \newcommand{\sz}{0.24}
    \newcommand{\sh}{2.5cm}
    \hspace*{-25pt}
    \begin{tabular}{ccccc}
        \rotatebox{90}{\hspace{11pt}RGB} &
        \includegraphics[width=\sz\linewidth]{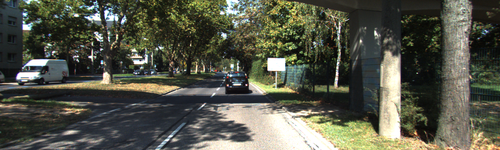} &
        \includegraphics[width=\sz\linewidth]{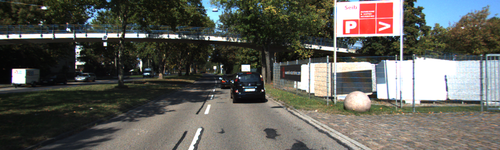} &
        \includegraphics[width=\sz\linewidth]{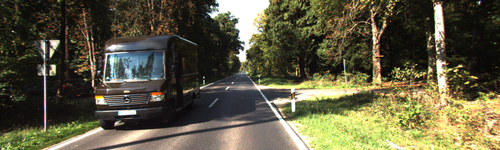} &
        \includegraphics[width=\sz\linewidth]{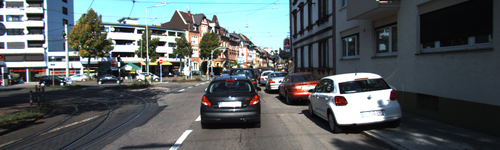}
\\
        \rotatebox{90}{\hspace{6pt}AdaBins} &
        \includegraphics[width=\sz\linewidth]{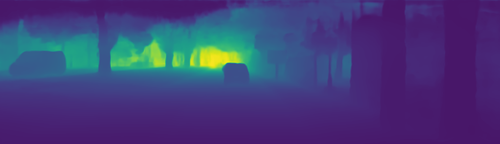} &
        \includegraphics[width=\sz\linewidth]{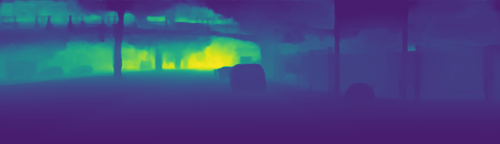} &
        \includegraphics[width=\sz\linewidth]{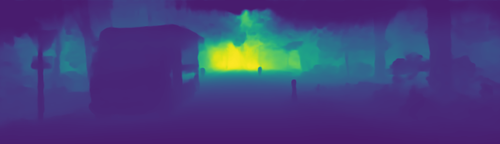} &
        \includegraphics[width=\sz\linewidth]{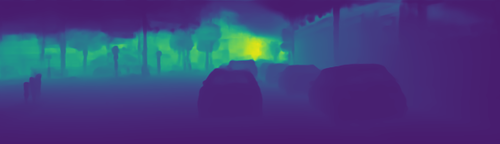}
\\
        \rotatebox{90}{\hspace{4pt}\shortstack{\textbf{NeRFm.}\\AdaBins}} &
        \includegraphics[width=\sz\linewidth]{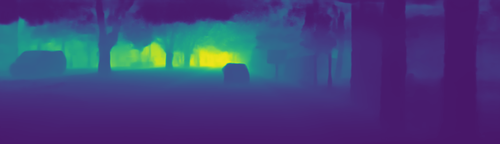} &
        \includegraphics[width=\sz\linewidth]{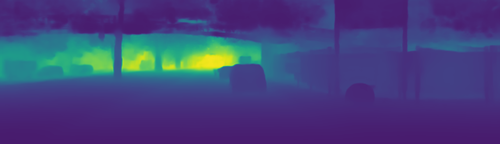} &
        \includegraphics[width=\sz\linewidth]{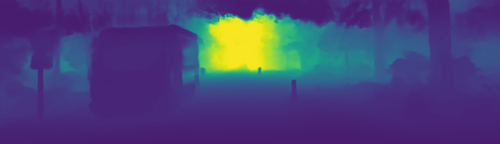} &
        \includegraphics[width=\sz\linewidth]{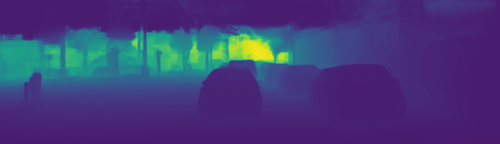}
\\
        \rotatebox{90}{\hspace{6pt}DepthF.} &
        \includegraphics[width=\sz\linewidth]{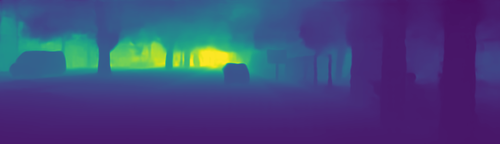} &
        \includegraphics[width=\sz\linewidth]{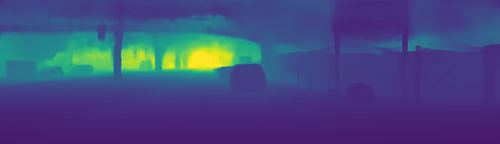} &
        \includegraphics[width=\sz\linewidth]{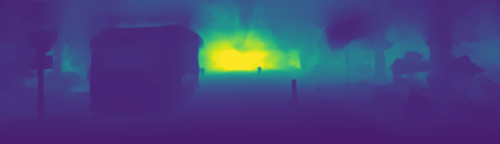} &
        \includegraphics[width=\sz\linewidth]{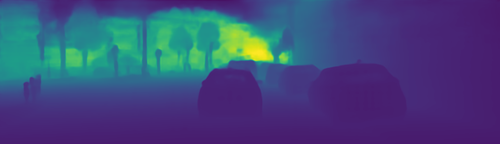}
\\
        
        \rotatebox{90}{\hspace{4pt}\shortstack{\textbf{NeRFm.}\\DepthF.}} &
        \includegraphics[width=\sz\linewidth]{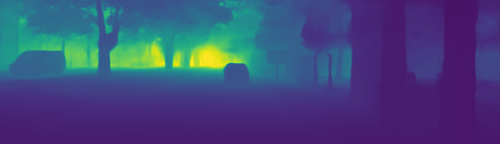} &
        \includegraphics[width=\sz\linewidth]{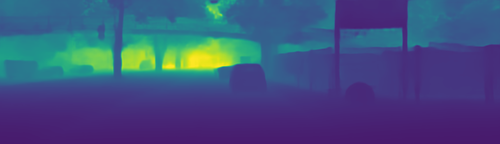} &
        \includegraphics[width=\sz\linewidth]{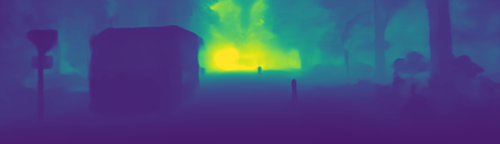} &
        \includegraphics[width=\sz\linewidth]{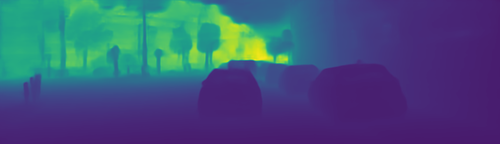}
\\
        \rotatebox{90}{\hspace{8pt}BinsF.} &
        \includegraphics[width=\sz\linewidth]{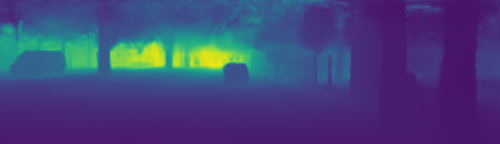} &
        \includegraphics[width=\sz\linewidth]{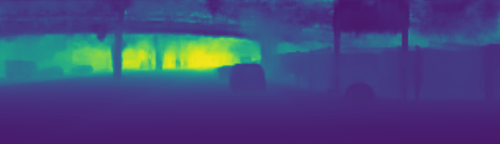} &
        \includegraphics[width=\sz\linewidth]{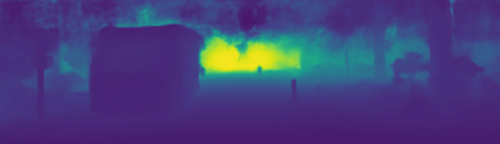} &
        \includegraphics[width=\sz\linewidth]{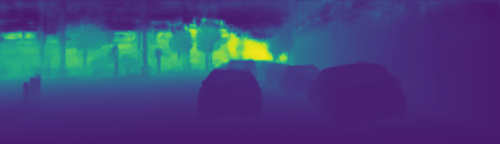}
\\
        \rotatebox{90} 
        {\hspace{4pt}\shortstack{\textbf{NeRFm.}\\BinsF.}} &
        \includegraphics[width=\sz\linewidth]{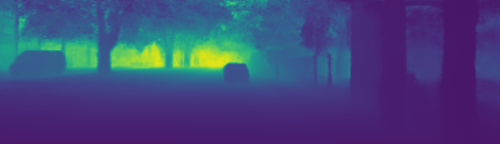} &
        \includegraphics[width=\sz\linewidth]{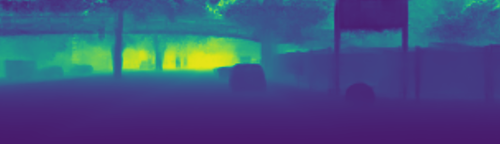} &
        \includegraphics[width=\sz\linewidth]{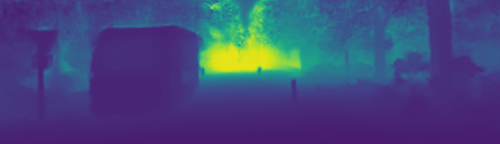} &
        \includegraphics[width=\sz\linewidth]{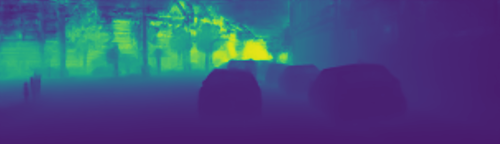}
\\
    \end{tabular}
    \caption{\textbf{Qualitative Results on the KITTI~\cite{geiger_kitti_2013} dataset.} We show the performance of NeRFmentation (Ours Interpolated) compared to vanilla-trained AdaBins~\cite{bhat_adabins_2021}, DepthFormer~\cite{li2023depthformer} and BinsFormer~\cite{li2022binsformer} leading to improved predictions by using NeRF augmented data for introducing additional viewpoints in the training images. Both the NeRFmented models and vanilla-trained models have been trained on the KITTI Eigen~\cite{eigen2014depth, geiger_kitti_2013} train split and evaluated on test split. Color scale: 0 (purple) to 80 meters (yellow).}
    \label{fig:qualitative_compared_sup}
\end{sidewaysfigure}

\cref{fig:qualitative_wamosup} and \cref{fig:qualitative_compared_sup} provide additional qualitative results compared to the state-of-the-art methods AdaBins~\cite{bhat_adabins_2021}, Depthformer~\cite{li2023depthformer} and BinsFormer~\cite{li2022binsformer} on the KITTI~\cite{geiger_kitti_2013} and Waymo~\cite{sun_scalability_2020} datasets. 
Qualitative improvements on KITTI are similar to those found on Waymo, our proposed method can predict increased details in objects such as trees and street signs, which can be especially noted in the top third part of the image. 

\newpage

\newpage

\FloatBarrier
\subsection{Comparison with DepthAnything}
DepthAnything is a promising series of novel foundation models for relative and metric depth estimation. For comparison, we choose DepthAnything v1 \cite{yang2024depthanythingunleashingpower} over v2 \cite{yang2024depthv2}, as weights for metric depth estimation fine-tuned on KITTI are only publicly available for v1. To evaluate NeRFmentation, we use DepthFormer trained on additional vertically translated views from KITTI but without classic augmentations.

\begin{table}[htb!]
\centering
\normalsize
\caption{Evaluation of DepthAnything v1 and NeRFmentation on KITTI and Waymo datasets. }
\label{tab:depth_comparison_grouped}
\begin{tabular}{lcc cc}
\toprule
                  & \multicolumn{2}{c}{KITTI} & \multicolumn{2}{c}{Waymo} \\
\cmidrule(lr){2-3} \cmidrule(lr){4-5}
                  & REL $\downarrow$ & $\delta_{1}\uparrow$  & REL $\downarrow$ & $\delta_{1}\uparrow$  \\
\midrule
DepthAnything v1 & \fs{0.046} & \fs{0.982} & 0.192 & 0.564 \\
NeRFmentation    & 0.053 & 0.974 & \fs{0.160} & \fs{0.724} \\
\bottomrule
\end{tabular}
\end{table}

\begin{figure}[htb!]
\centering
\begin{subfigure}{.49\textwidth}
  \centering
  \includegraphics[width=0.98\linewidth]{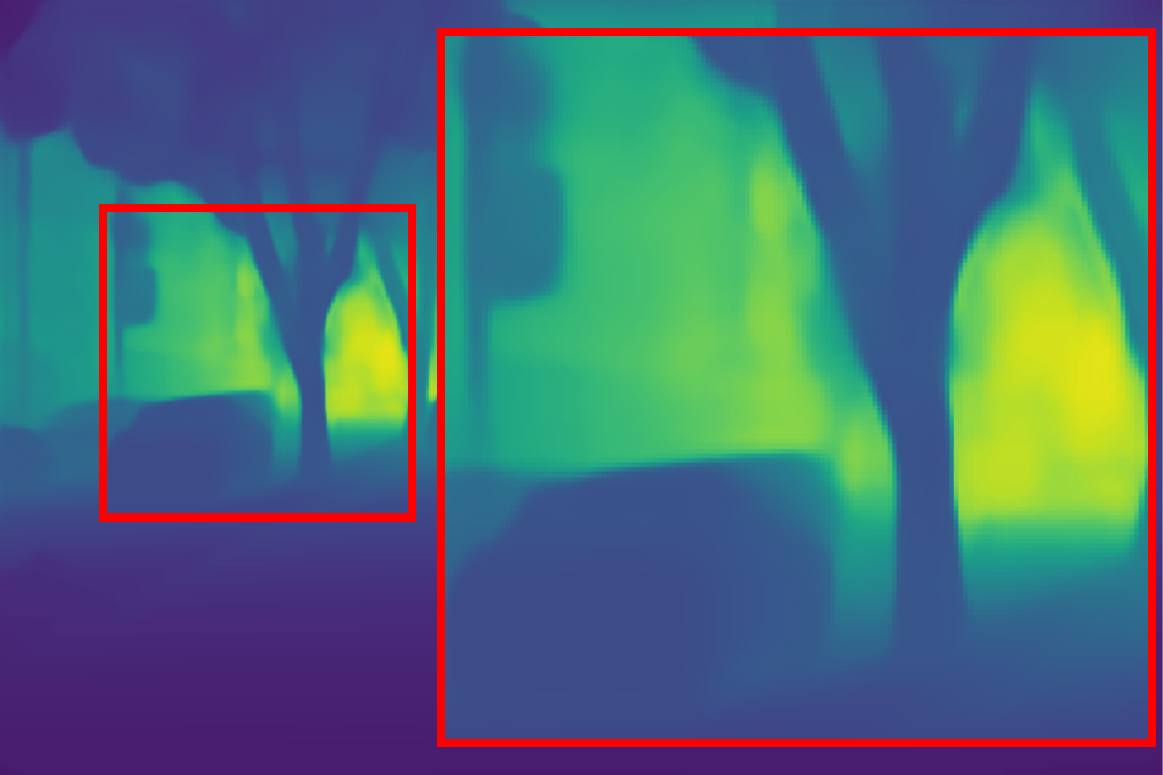}
\end{subfigure}%
\begin{subfigure}{.49\textwidth}
  \centering
  \includegraphics[width=0.98\linewidth]{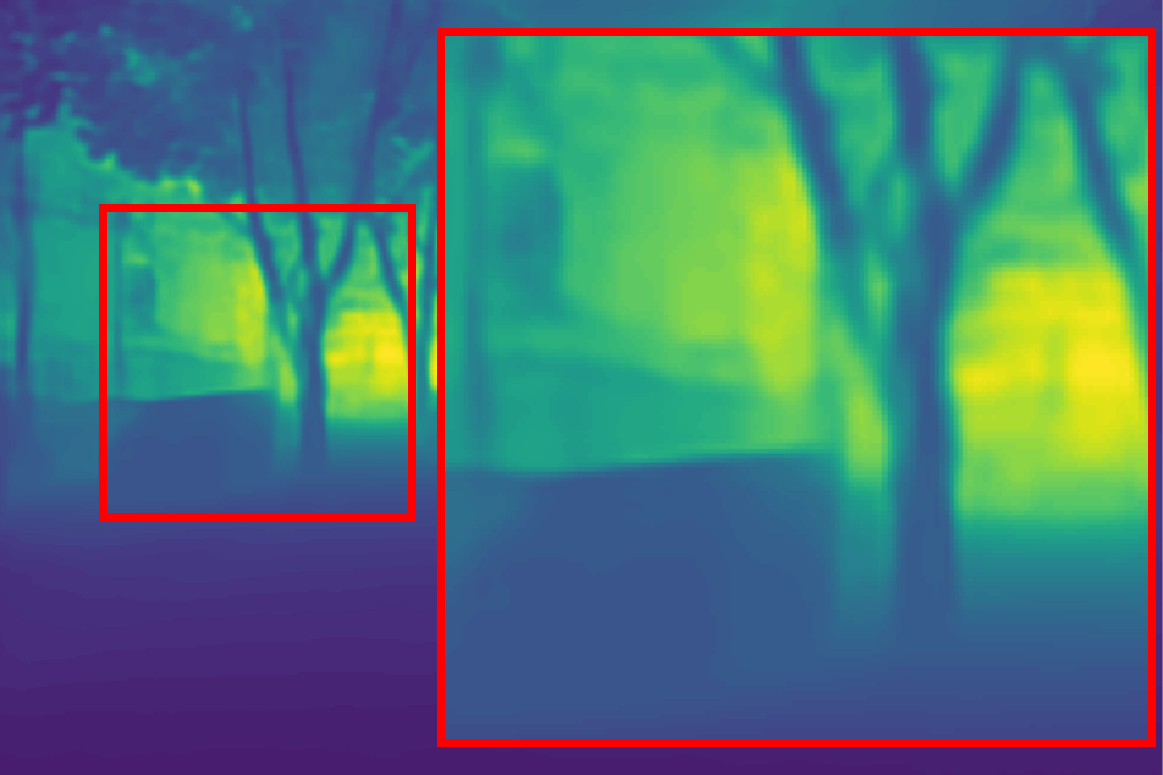}
\end{subfigure}
\caption{Zoomed visualizations of the predictions of DepthAnything v1 (left) and NeRFmentation (right) on the Waymo dataset.}
\label{fig:visual-comparison}
\end{figure}

As shown in  Table \ref{tab:depth_comparison_grouped}, the performance of DepthAnything on in-distribution tasks (KITTI) is impressive. However, the motivation for NeRFmentation is to avoid dataset-specific overfitting. When testing on Waymo, especially the large difference in $\delta_1$ scores suggests that DepthAnything's estimates are significantly worse more often than those of NeRFmentation. This is also supported by the finer details of NeRFmentation's predictions shown in Figure \ref{fig:visual-comparison}. Therefore, we argue that NeRFmentation is more robust to unseen data than DepthAnything.

We also want to highlight that finetuning DepthAnything could benefit from NeRFmentation as well. Unfortunately, this experiment was not possible due to time constraints.

\FloatBarrier
\section{Qualitative comparison between the Reconstructed, Interpolated, and Angled augmentation strategies}
\label{sec:angled}

\begin{sidewaysfigure}[htb!]
    \hspace{-0.4cm}
    \centering
    \tiny
    \setlength{\tabcolsep}{1pt}
    \renewcommand{\arraystretch}{0.8}
    \newcommand{\sz}{0.145\linewidth}
    \begin{tabular}{cccccc}
         RGB & DepthFormer & \makecell[c]{\textbf{NeRFm.} DepthFormer \\ (Reconstructed)} & \makecell[c]{\textbf{NeRFm.} DepthFormer \\ (Interpolated)} & \makecell[c]{\textbf{NeRFm.} DepthFormer \\ (Angled)} \\
        \includegraphics[width=\sz]{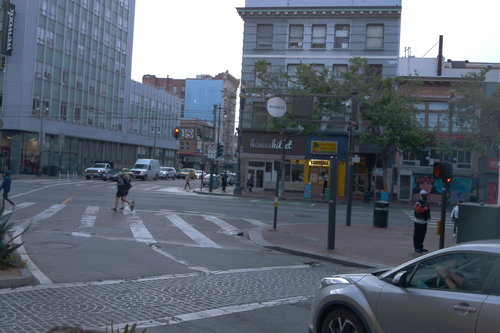}&
        \includegraphics[width=\sz]{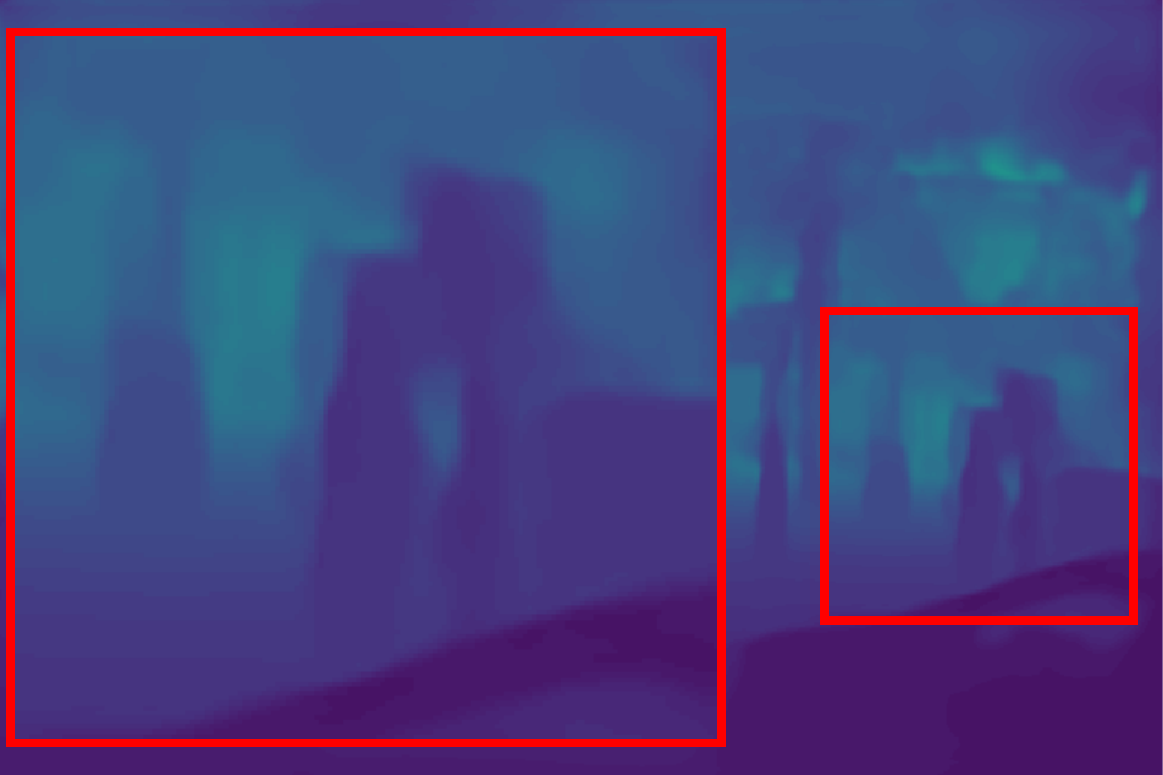}&
        \includegraphics[width=\sz]{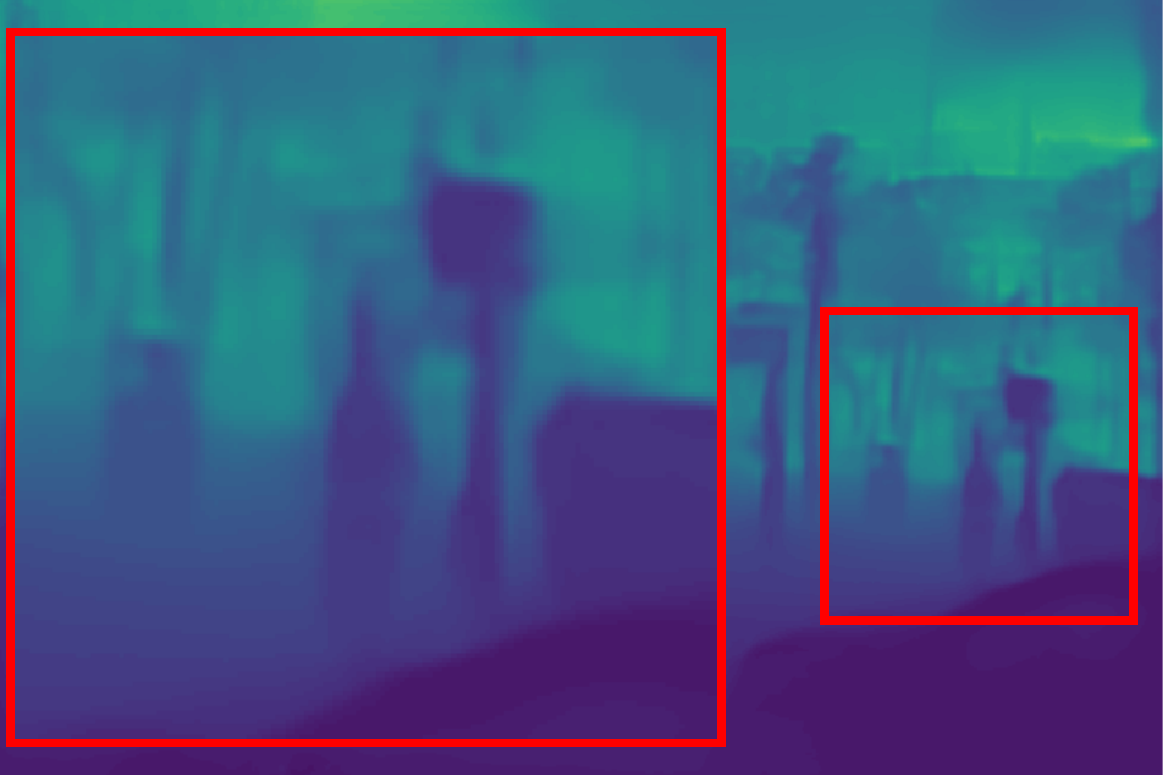}&
        \includegraphics[width=\sz]{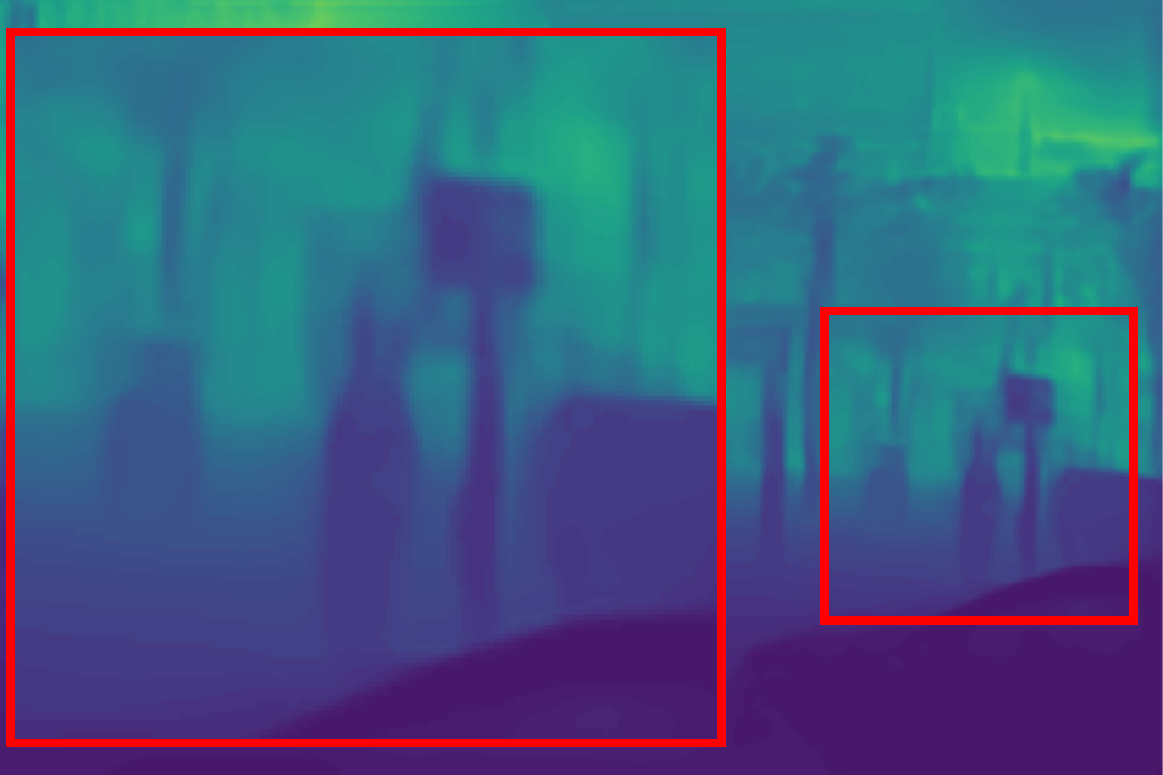}&
        \includegraphics[width=\sz]{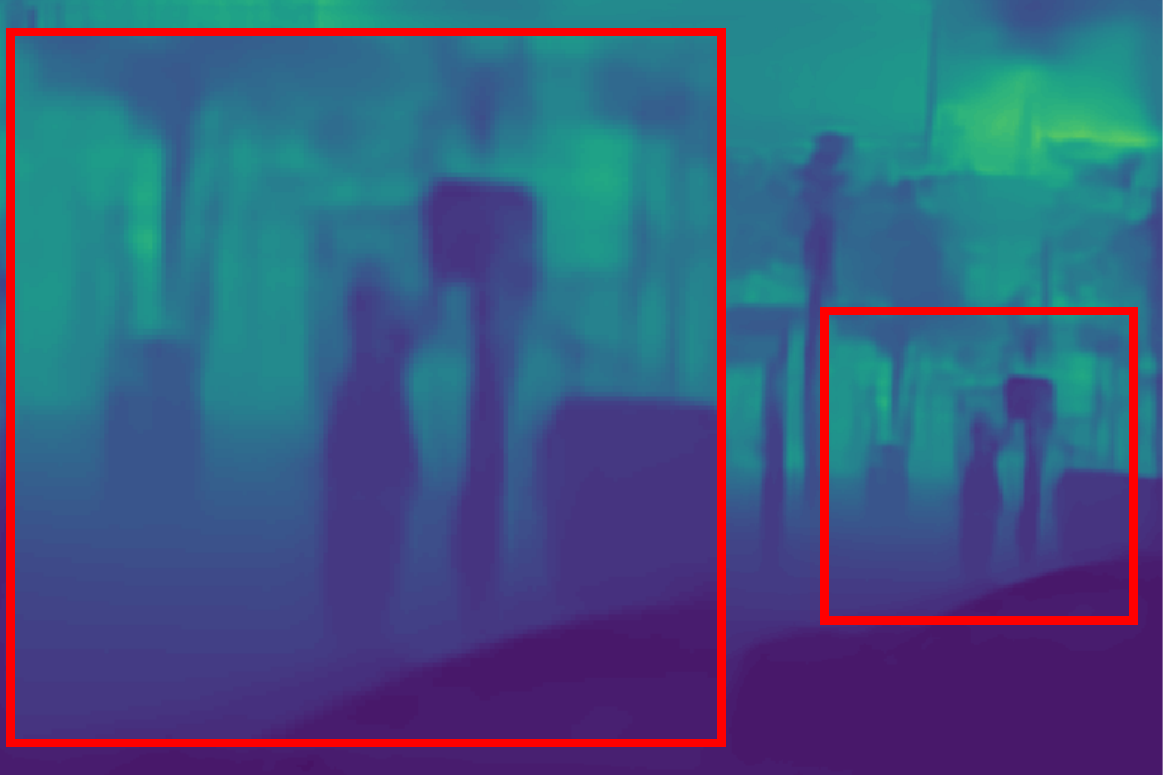}\\

        \includegraphics[width=\sz]{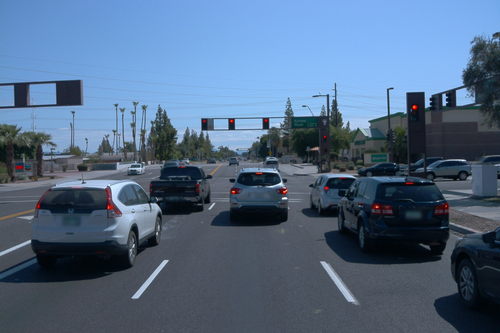}&
        \includegraphics[width=\sz]{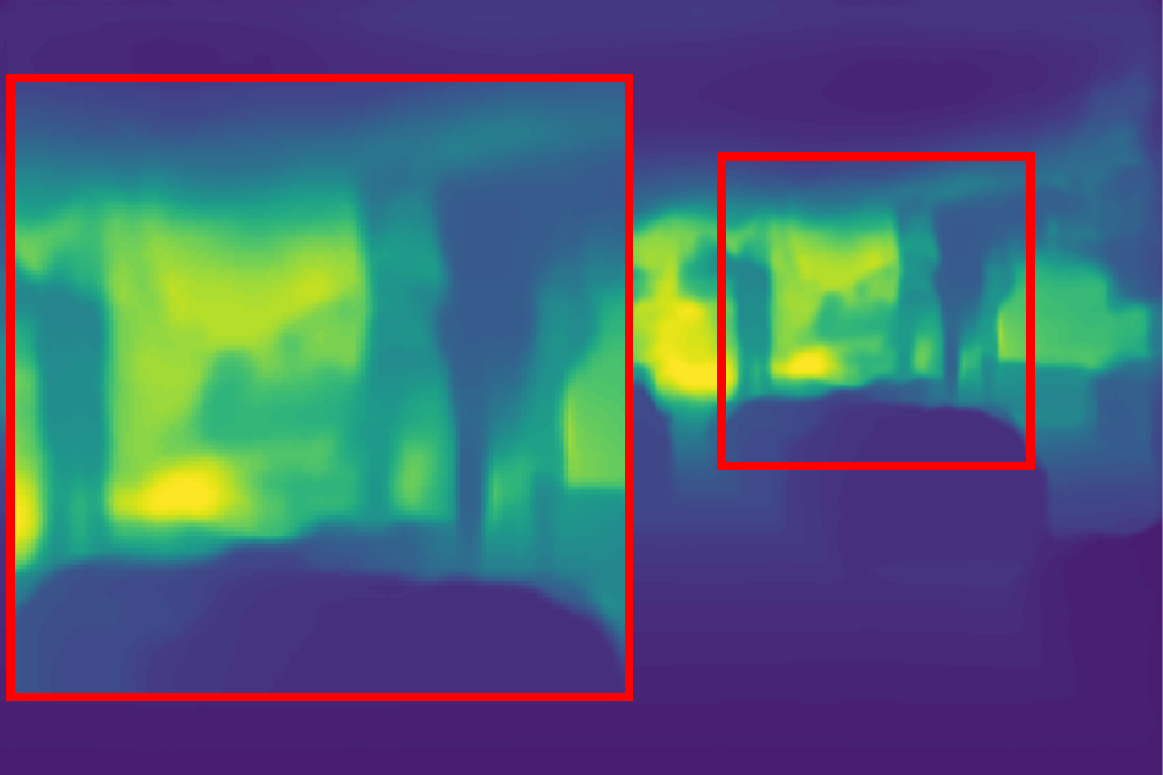}&
        \includegraphics[width=\sz]{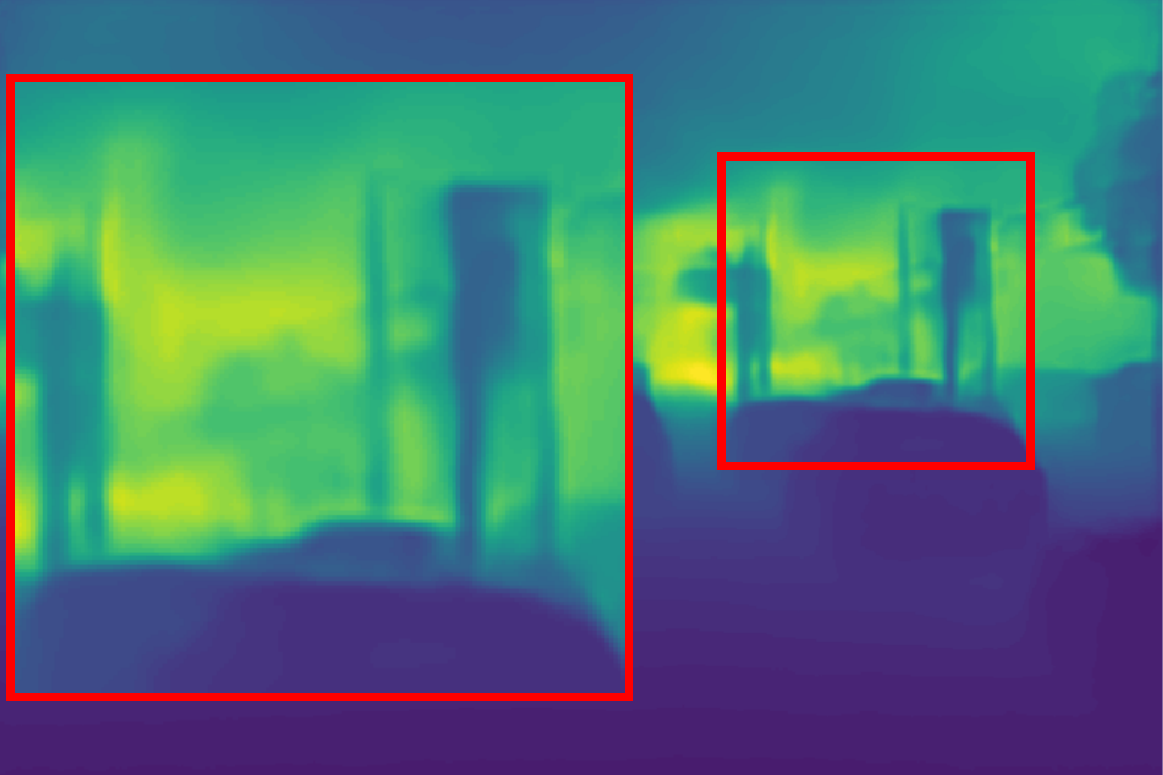}&
        \includegraphics[width=\sz]{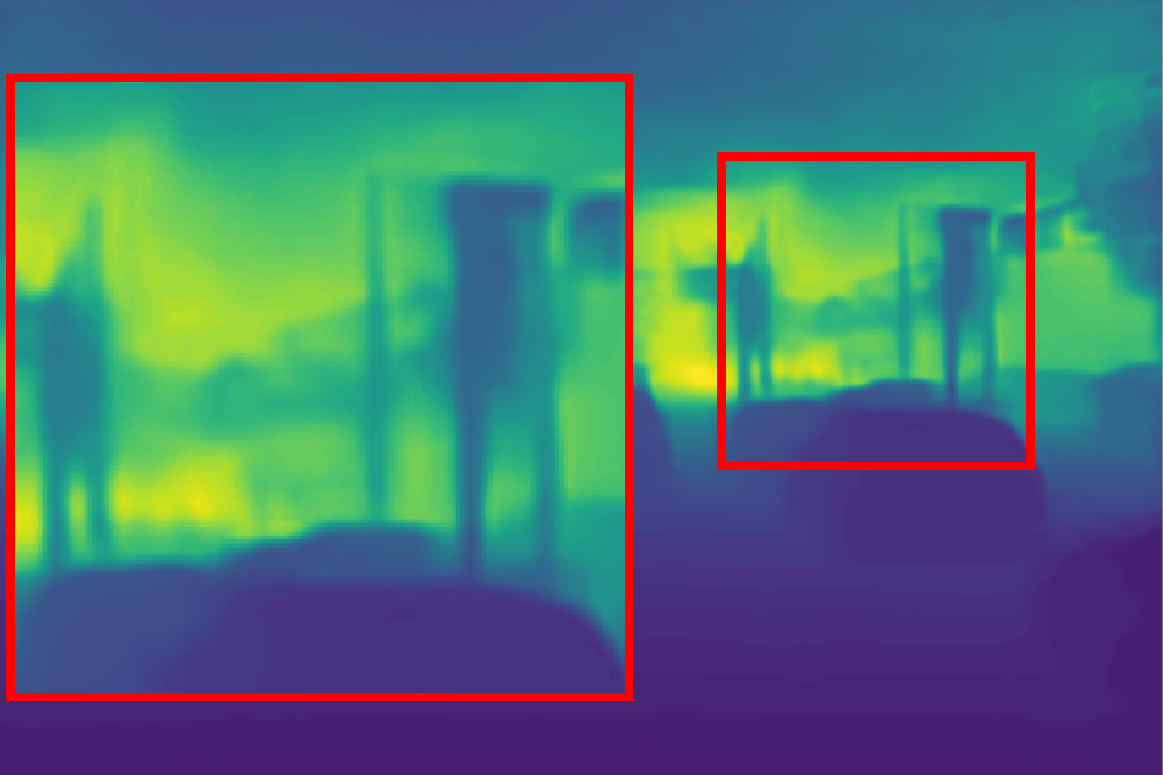}&
        \includegraphics[width=\sz]{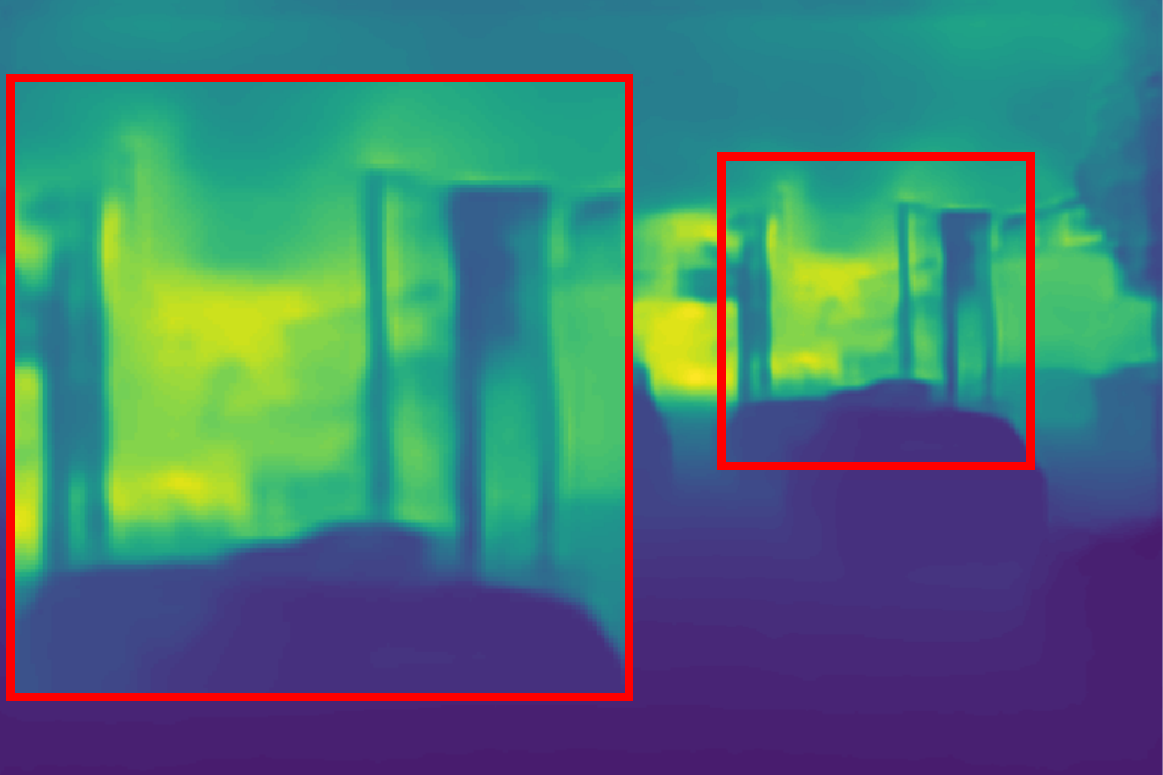}\\
        
        \includegraphics[width=\sz]{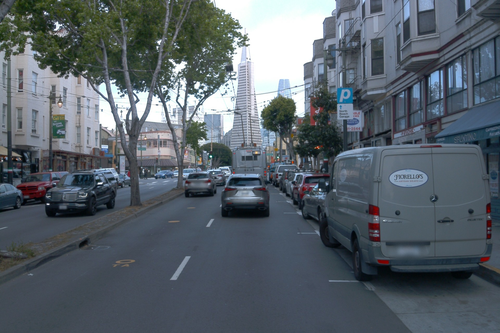}&
        \includegraphics[width=\sz]{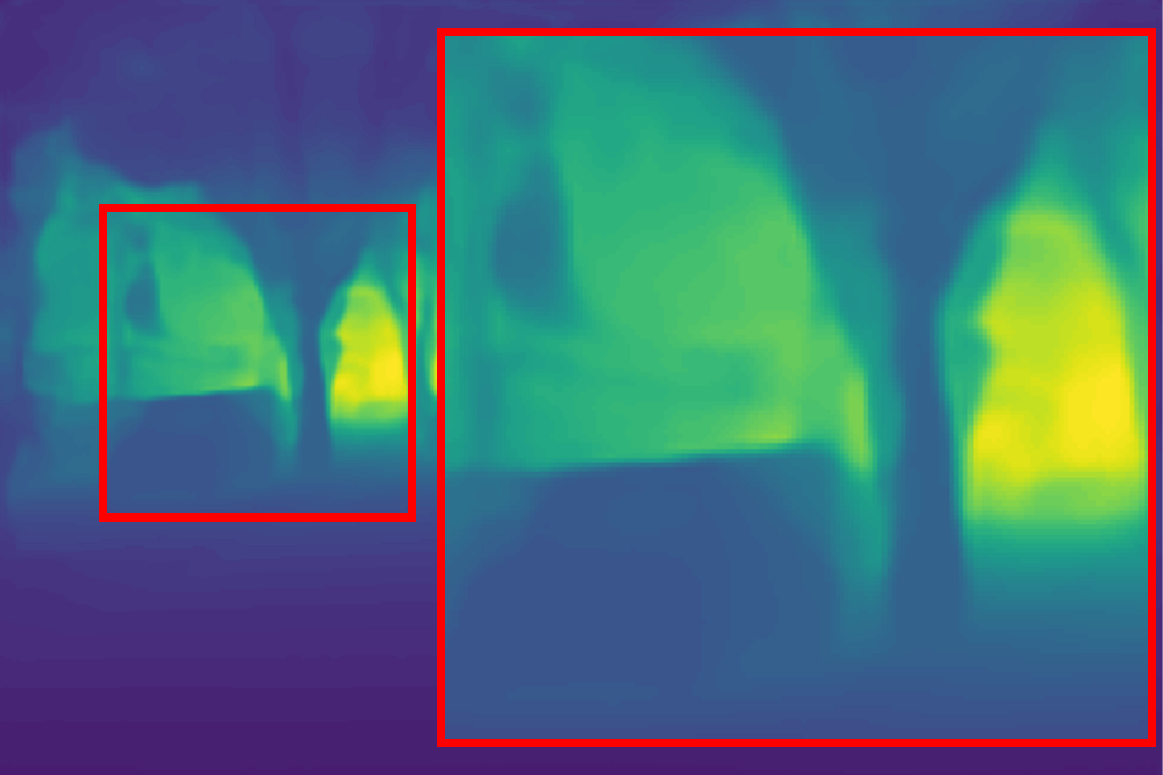}&
        \includegraphics[width=\sz]{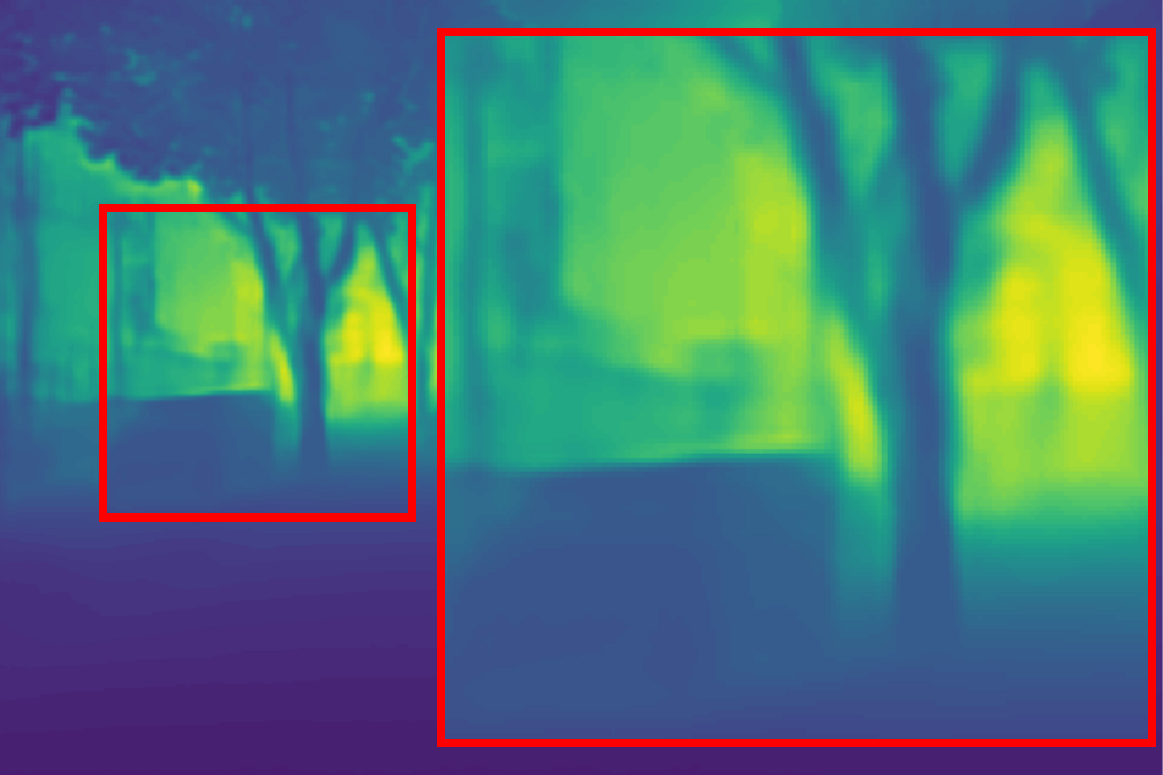}&
        \includegraphics[width=\sz]{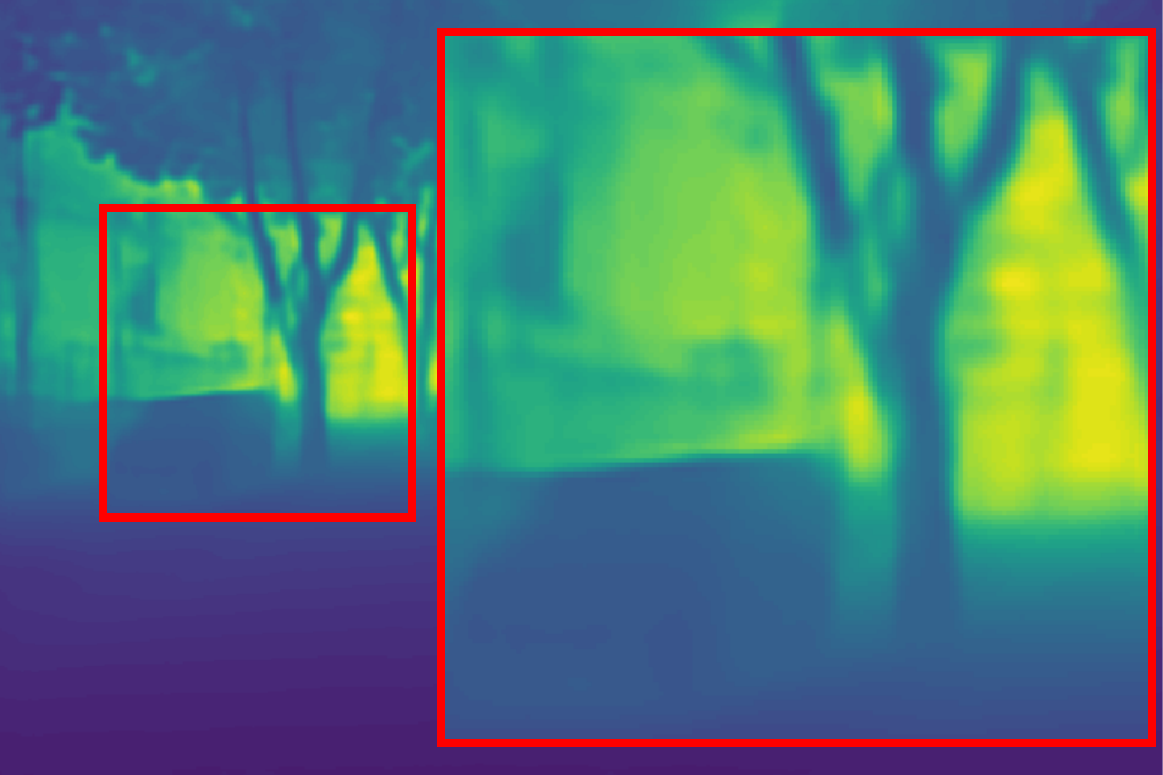}&
        \includegraphics[width=\sz]{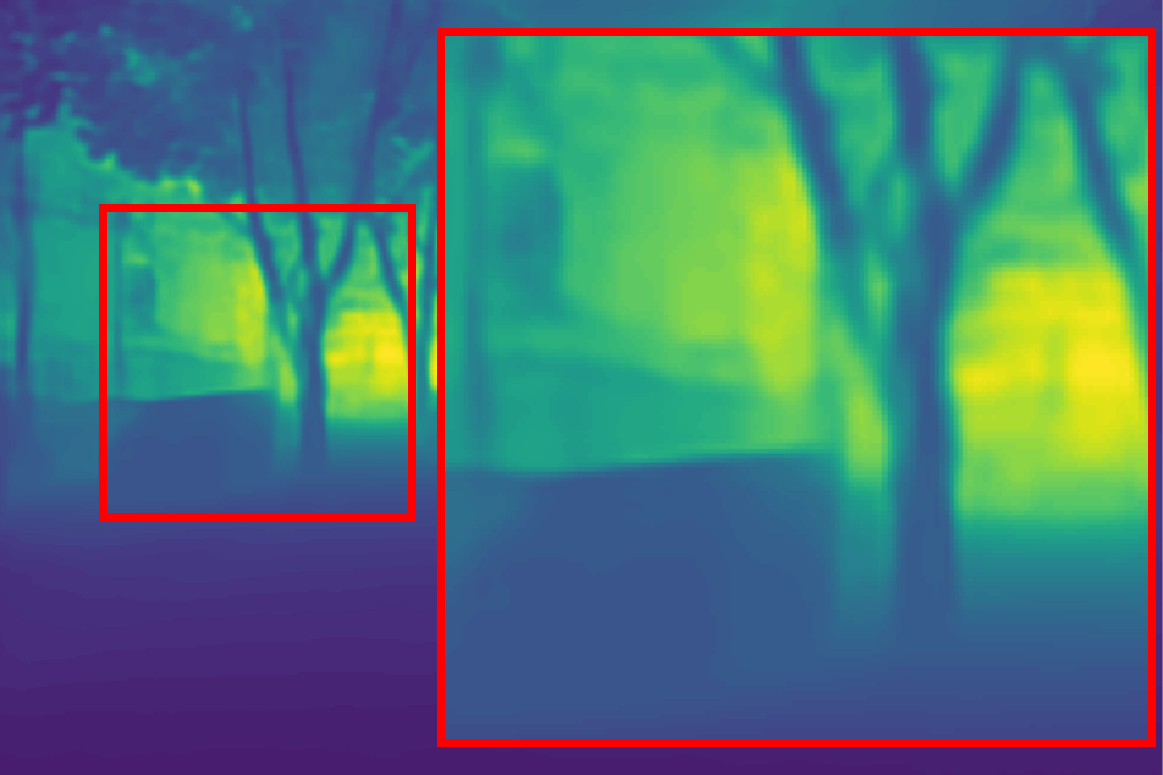}\\

        \includegraphics[width=\sz]{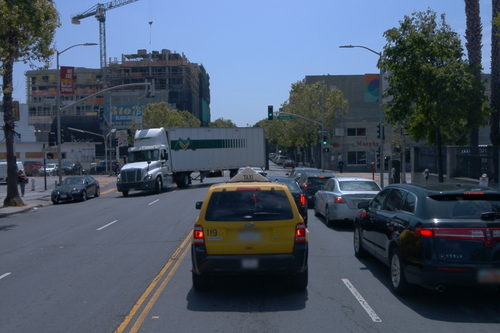}&
        \includegraphics[width=\sz]{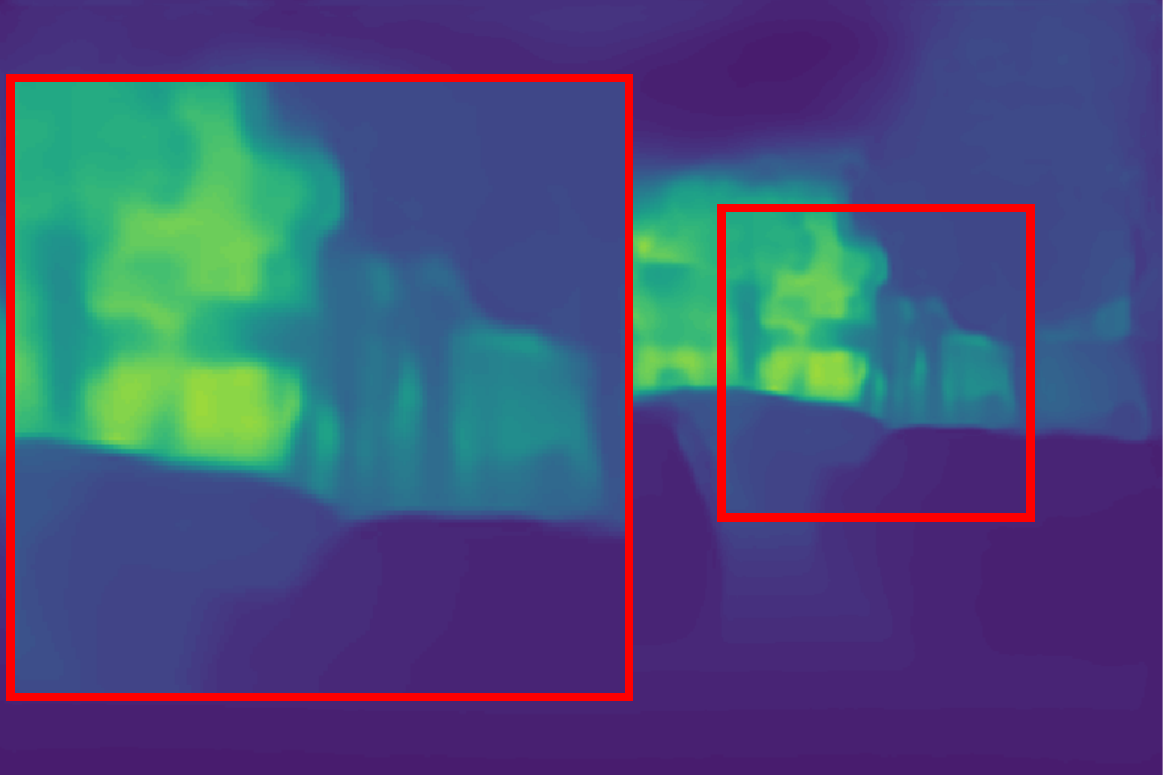}&
        \includegraphics[width=\sz]{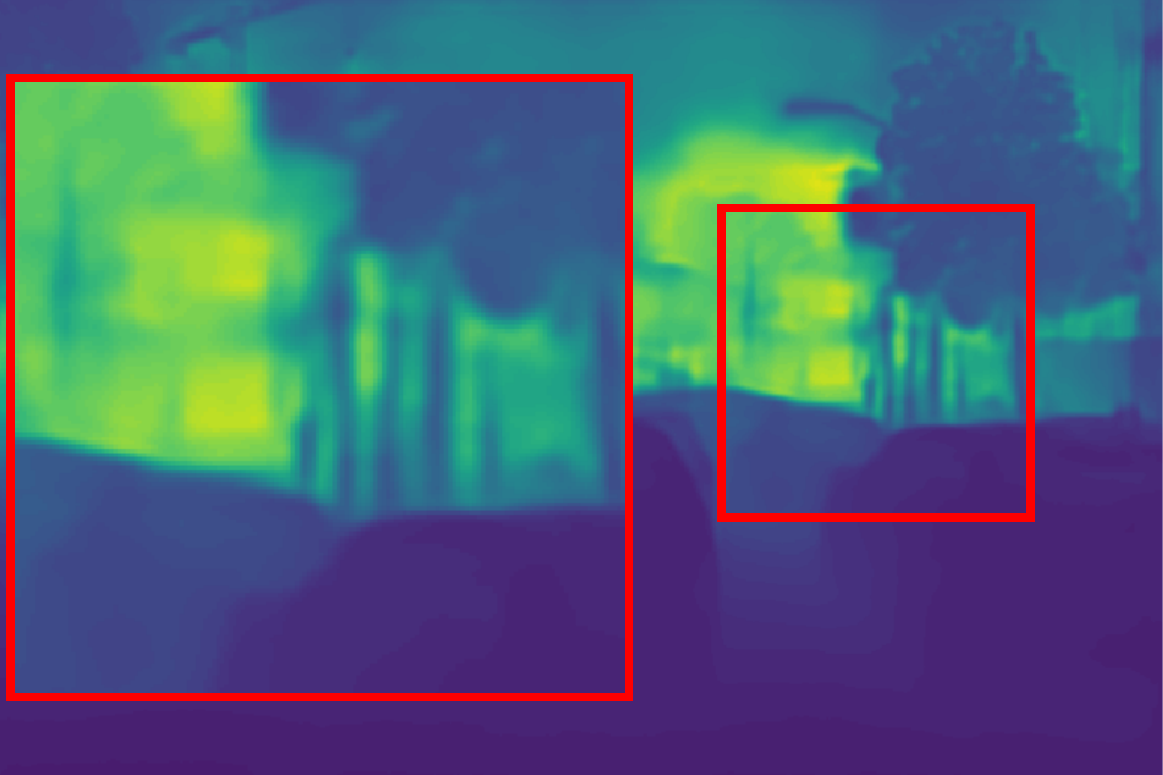}&
        \includegraphics[width=\sz]{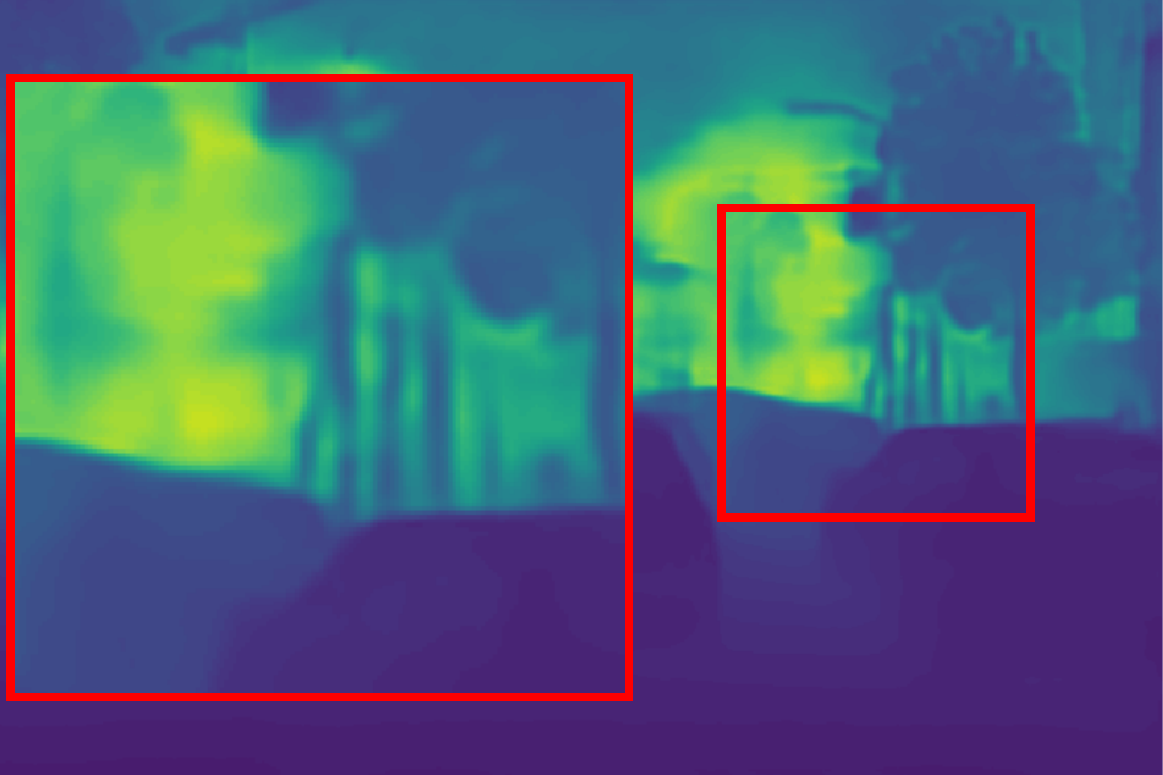}&
        \includegraphics[width=\sz]{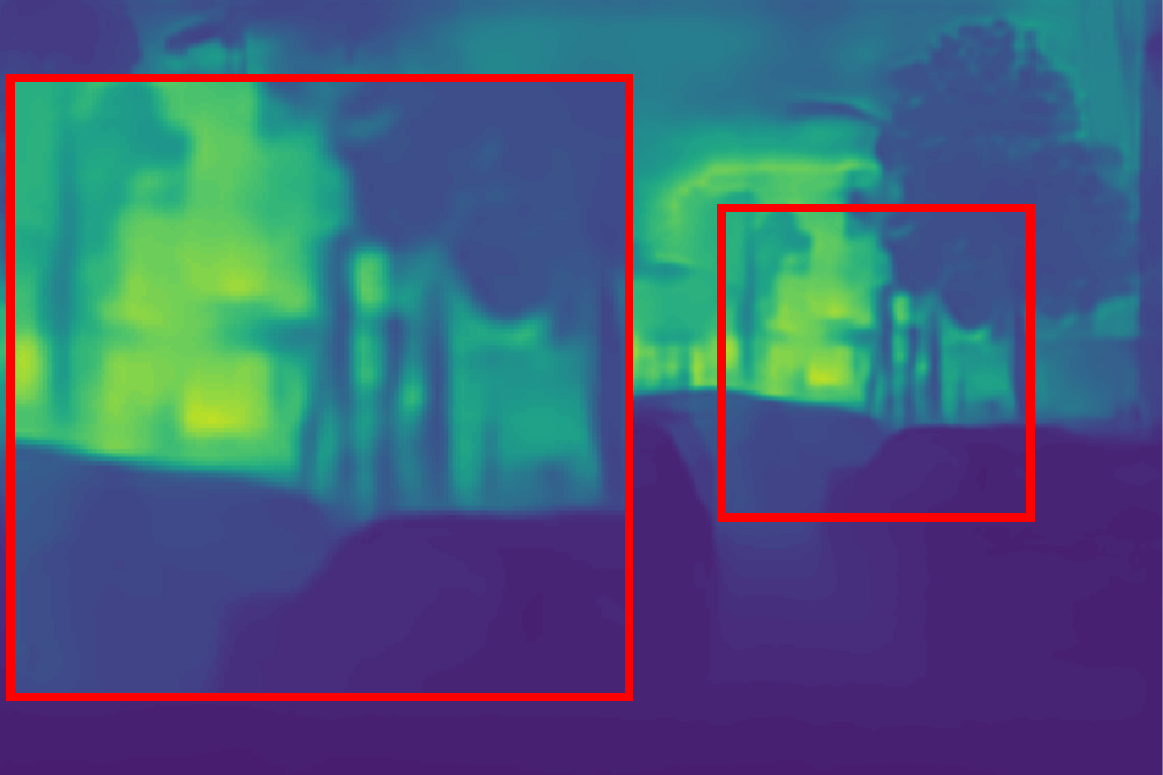}\\

        \includegraphics[width=\sz]{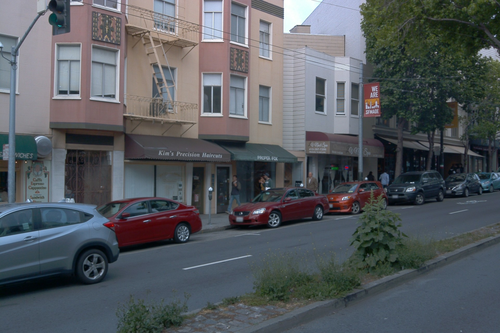}&
        \includegraphics[width=\sz]{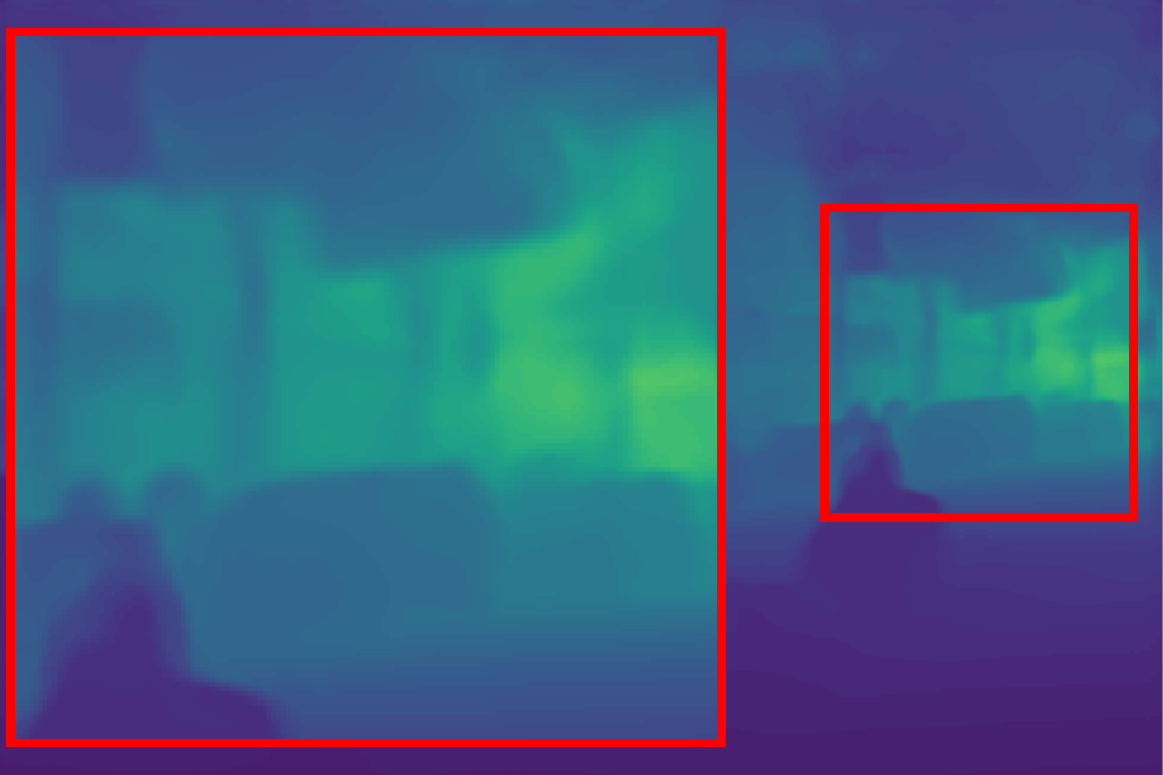}&
        \includegraphics[width=\sz]{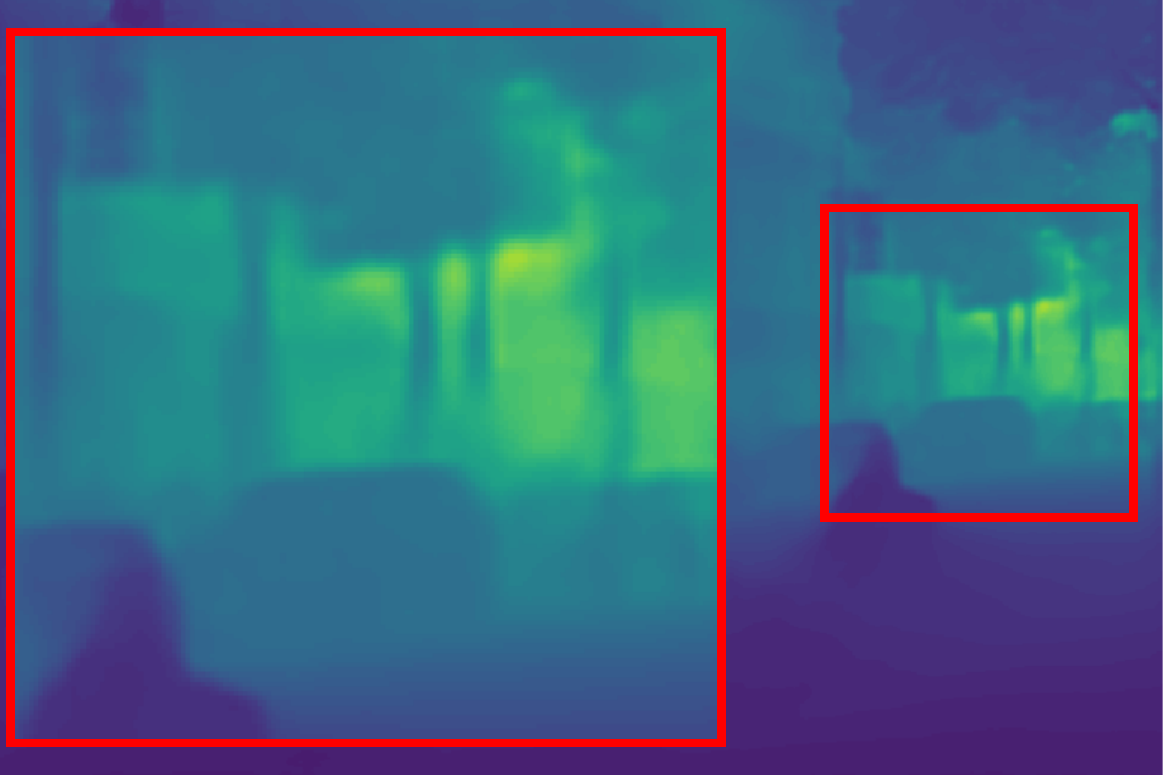}&
        \includegraphics[width=\sz]{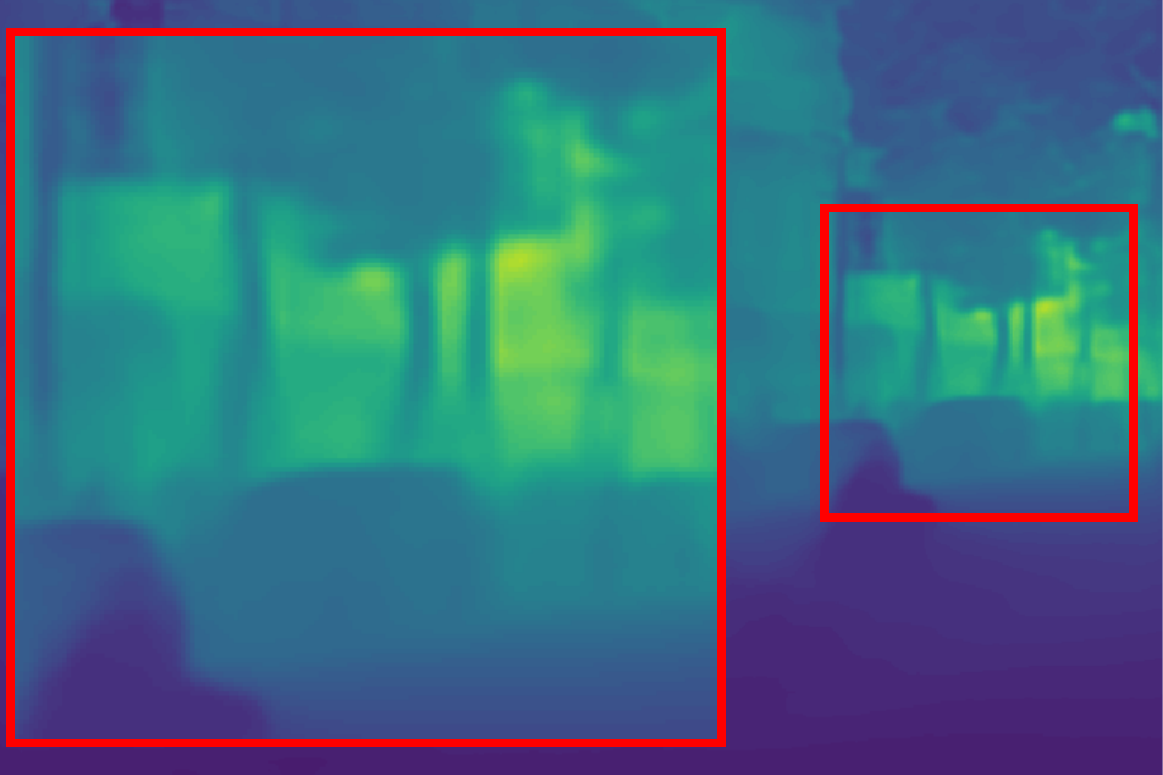}&
        \includegraphics[width=\sz]{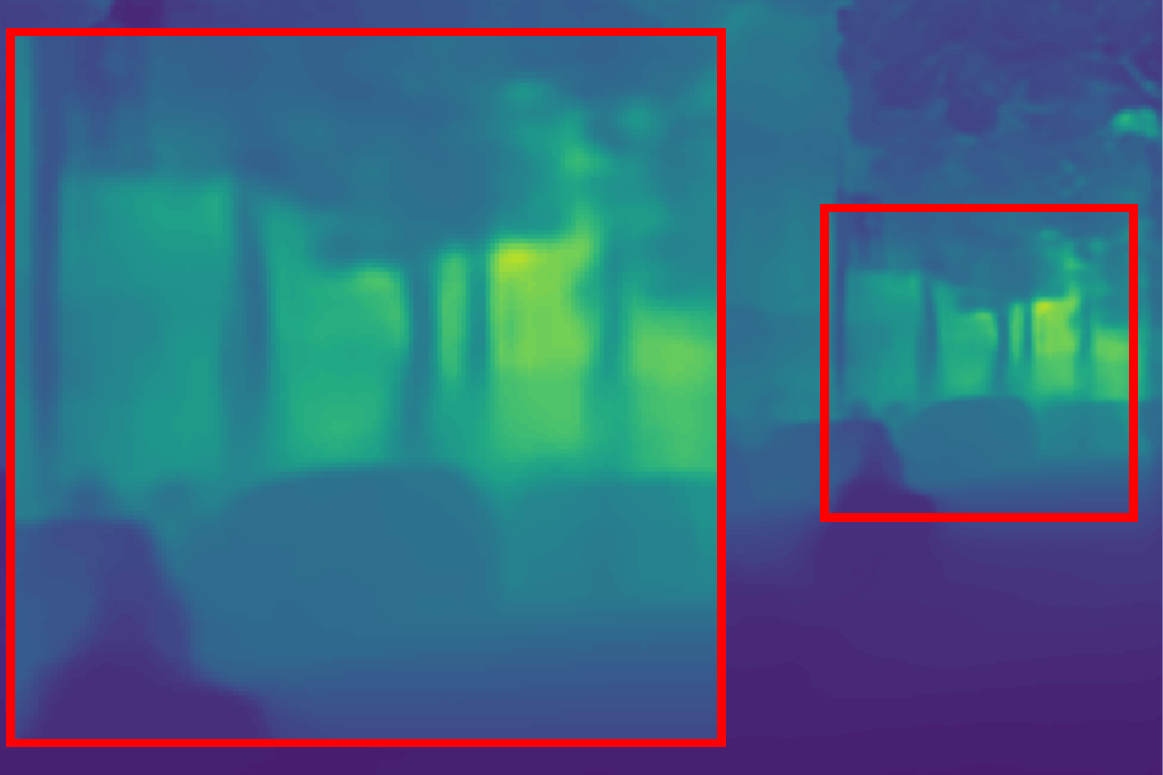}\\

        \includegraphics[width=\sz]{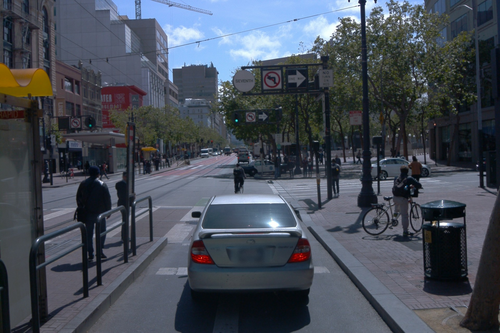}&
        \includegraphics[width=\sz]{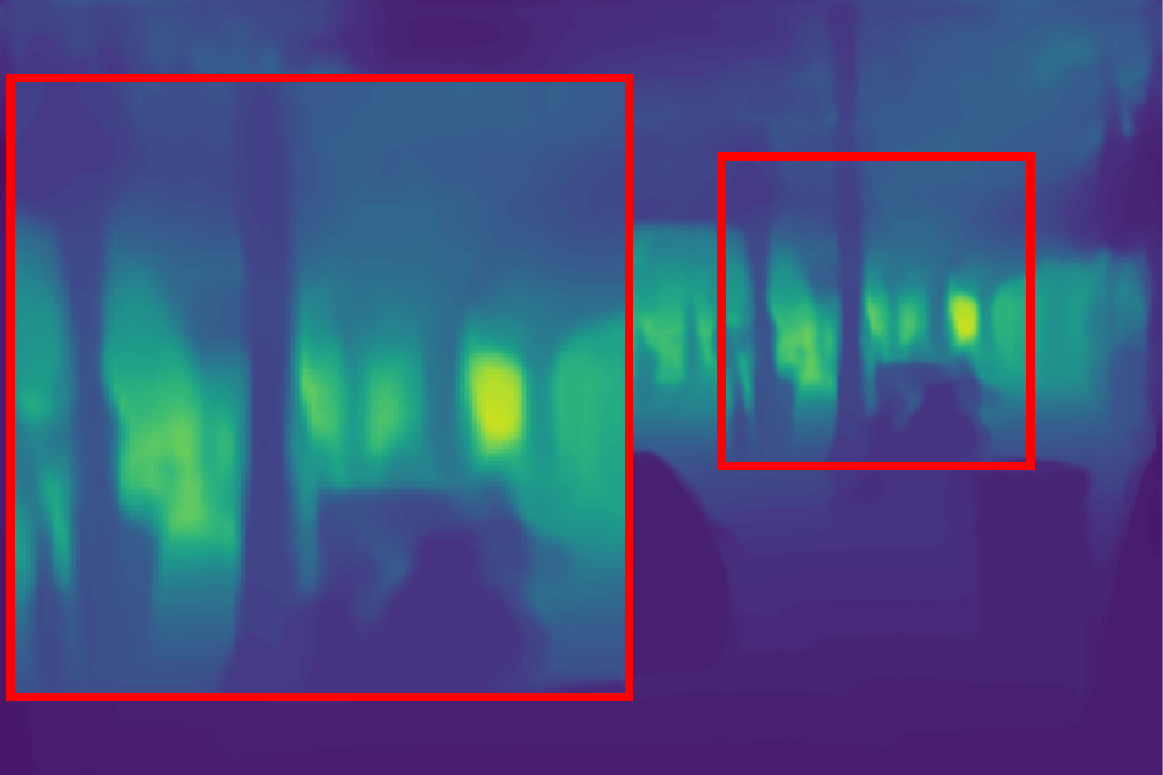}&
        \includegraphics[width=\sz]{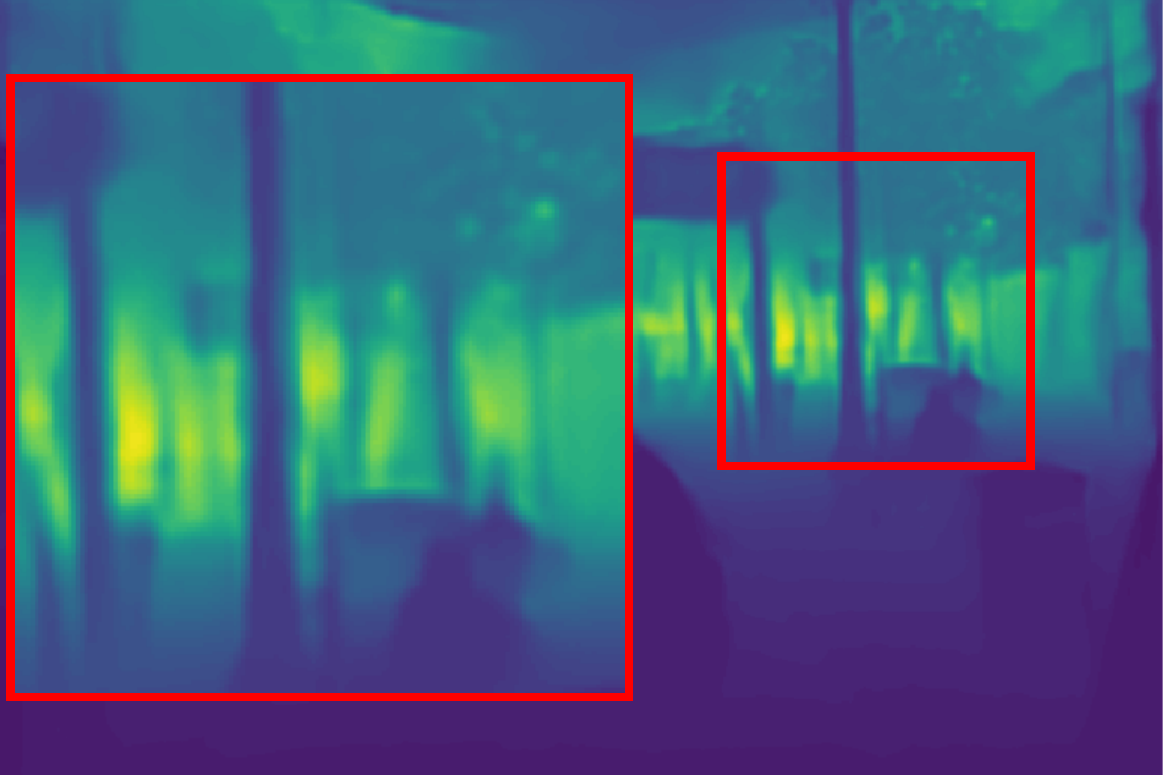}&
        \includegraphics[width=\sz]{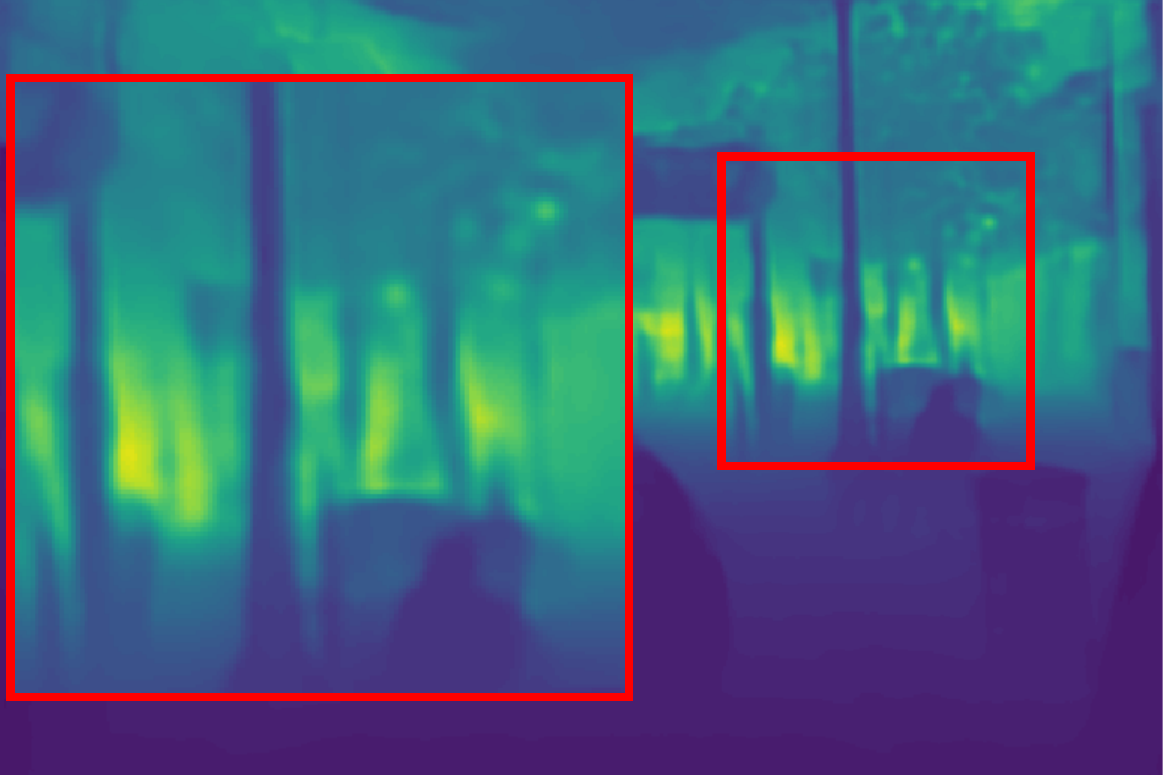}&
        \includegraphics[width=\sz]{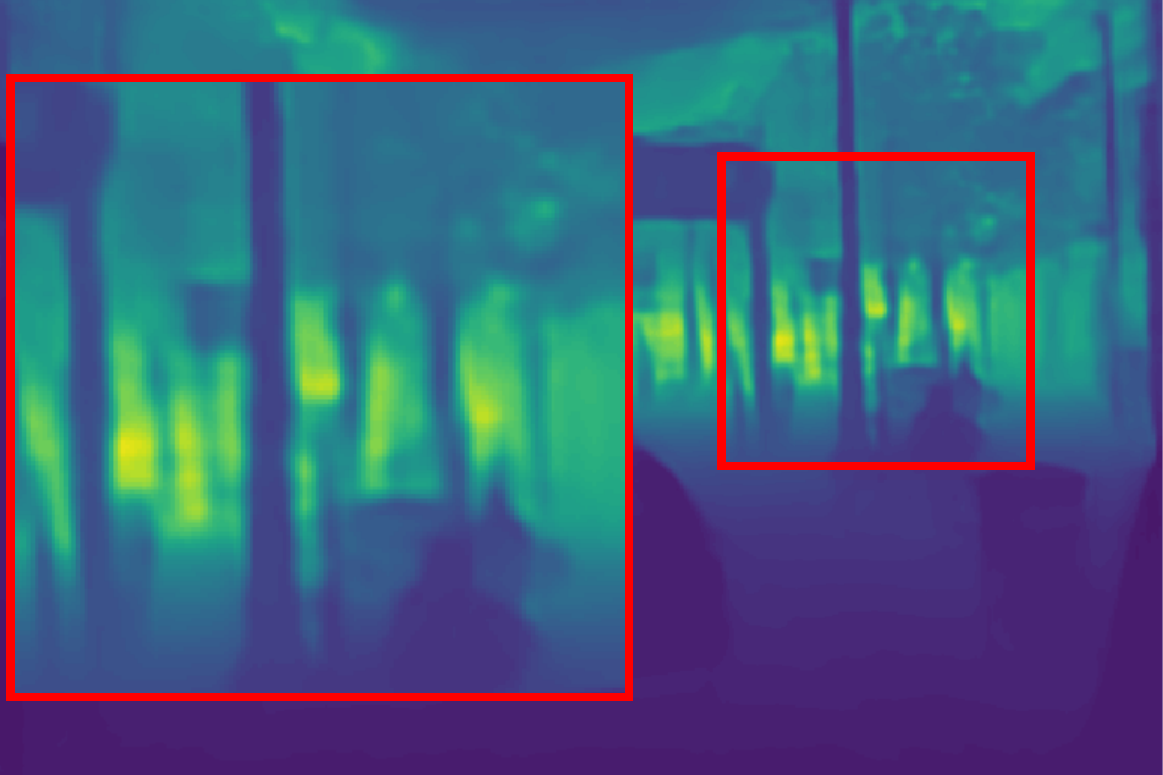}\\
    \end{tabular} 
    \caption{\textbf{Qualitative DepthFormer~\cite{li2023depthformer} results on the Waymo~\cite{sun_scalability_2020} dataset, focusing on close-up details.} Greatly improved detail compared to the baseline, but no apparent differences between reconstructed, interpolated, and angled data augmentation strategies. Color scale: 0 (purple) to 80 meters (yellow).}
    \label{fig:zoomedin_waymo_method_comparison}
\end{sidewaysfigure}

We compare the three proposed data augmentation strategies in conjunction with DepthFormer in \cref{fig:zoomedin_waymo_method_comparison}. The differences between the different NeRFmented depth predictions on the Waymo dataset show that they all perform similarly. This suggests that the depth completion offered by NeRFs is the significant driver behind the performance gains. 

\newpage

\newpage

\FloatBarrier
\section{Scene mesh of NeRFs trained on KITTI}
\label{sec:supp_qualitative_nerf}

\begin{figure}[htb!]
    \centering
    \scriptsize
    \setlength{\tabcolsep}{1pt}
	\renewcommand{\arraystretch}{0.8}
	\newcommand{\sz}{0.48}
	\newcommand{\sh}{1.2cm}
	\newcommand{\gcs}{\hspace{8pt}}  %
	\begin{tabular}{cc@{\gcs}}
        \includegraphics[width=\sz\linewidth]{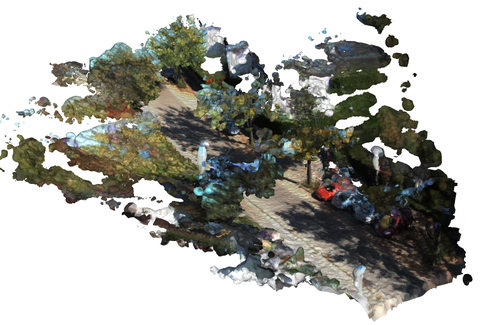} &
        \includegraphics[width=\sz\linewidth]{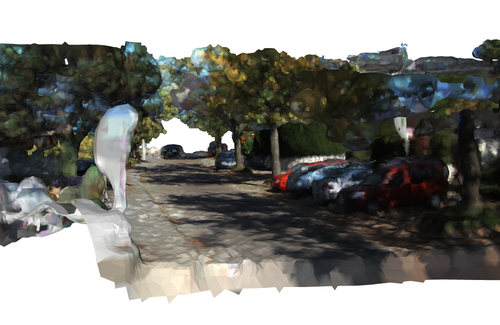} \\
        Bird's-eye view & Front-facing view \\
    \end{tabular}
    \caption{\textbf{Mesh generated with NeRF that was trained on KITTI~\cite{geiger_kitti_2013} sub-scene.} One can see that the underlying scene geometry is accurately learned. The bird's-eye view shows what a single sub-scene looks like after the scene has been split into smaller 50-meter sub-scenes.}
    \label{fig:mesh}
\end{figure}

Using the built-in Poisson mesh export in Nerfstudio \cite{tancik_nerfstudio_2023} we generate a mesh of the NeRF scene. \cref{fig:mesh} shows a bird's-eye view and a front-facing view of a 50-meter KITTI sub-scene. This 3D representation allows us to verify that the NeRF depth map predictions are sound and consistent. 

\newpage

\FloatBarrier
\section{Discussion: NeRFmentation on NYU-Depth V2}
\label{sec:supp_nyu}

Given the promising results produced using NeRFmentation on the task of MDE on outdoor scenes, investigating the efficacy of this method on MDE on indoor scenes becomes an obvious follow-up question. The tasks of indoor and outdoor MDE have differing constraints, use cases, and available resources. The lack of diverse poses in outdoor scene trajectories, for instance, may not translate to indoor scenes. Here a trajectory may be captured by a device like the Microsoft Kinect, which is able to operate with a wide range of viewing angles and short physical distances between captured frames. This is because a Kinect is not limited by the lanes of a road or the speed of a vehicle. Thus, the inherent problems present in outdoor MDE and the resulting issue of data scarcity that NeRFmentation aims to tackle may have limited value for the indoor setting. Our following investigation applies the same experiments and procedures previously followed for the outdoor KITTI Dataset to the indoor NYU-Depth V2 Dataset. The performance of NeRFmentation is evaluated using AdaBins~\cite{bhat_adabins_2021} and DepthFormer~\cite{li2023depthformer} models. 

 \paragraph{NYU-Depth V2 Dataset~\cite{silberman2012indoor}.} 
 NYU-Depth V2 is an indoor dataset consisting of various commercial and residential scenes captured using Microsoft Kinect. The dataset comprises a smaller subset with dense segmentation labels and preprocessed depth as well as the much larger raw output of the Kinect sensor in the form of raw RGB, depth, and accelerometer data. The raw data is used for the purposes of our work, including a training set of approximately \textbf{120K} samples and a test set of \textbf{654} samples~\cite{eigen2014depth}. As the RGB and depth images are not synchronized, they are preprocessed using the toolbox provided by the dataset authors. The images are additionally center cropped as described by Eigen \etal~\cite{eigen2014depth}. 

\subsection{Training on NYU-Depth V2} 

\begin{figure}[htb!]
    \centering
    \scriptsize
    \setlength{\tabcolsep}{1pt}
	\renewcommand{\arraystretch}{0.8}
	\newcommand{\sz}{0.40}
	\newcommand{\sh}{1.2cm}
	\newcommand{\gcs}{\hspace{8pt}}  %
	\begin{tabular}{cc@{\gcs}}
        \rotatebox{90}{\hspace{34pt} GT}
        \includegraphics[width=\sz\columnwidth]{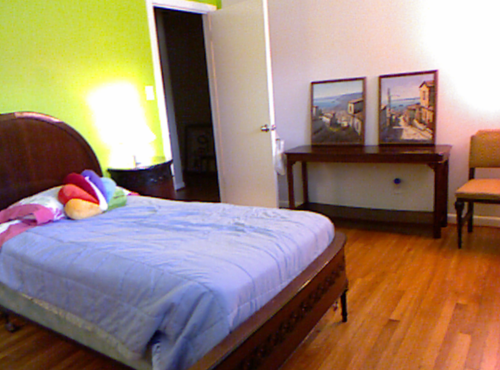} &
        \includegraphics[width=\sz\columnwidth]{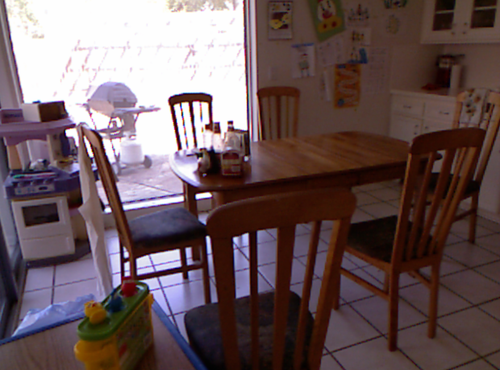} \\
        \rotatebox{90}{\hspace{32pt} Ours}
        \includegraphics[width=\sz\columnwidth]{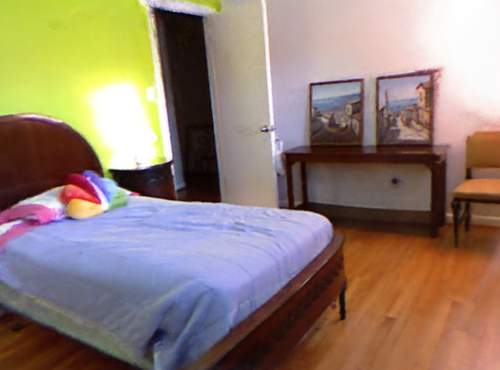} &
        \includegraphics[width=\sz\columnwidth]{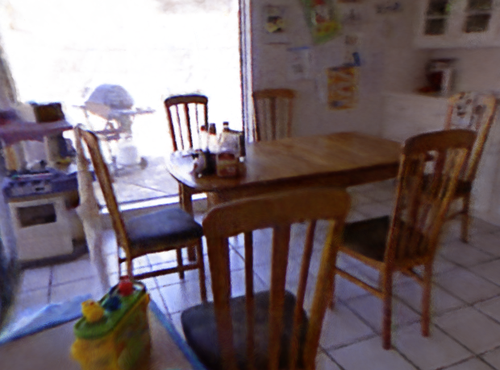} \\
        \rotatebox{90}{\hspace{34pt} GT}
        \includegraphics[width=\sz\columnwidth]{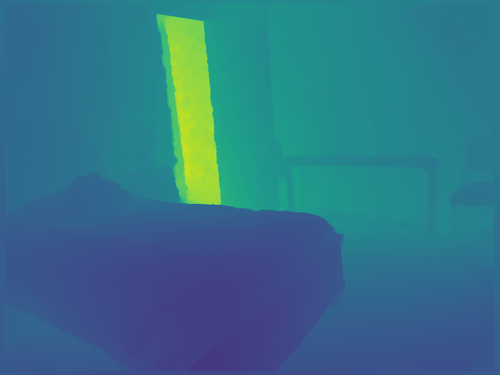} &
        \includegraphics[width=\sz\columnwidth]{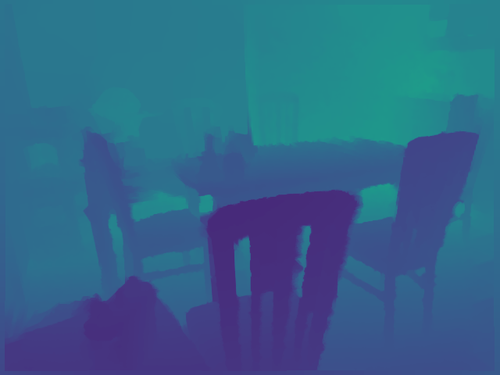} \\
        \rotatebox{90}{\hspace{32pt} Ours}
        \includegraphics[width=\sz\columnwidth]{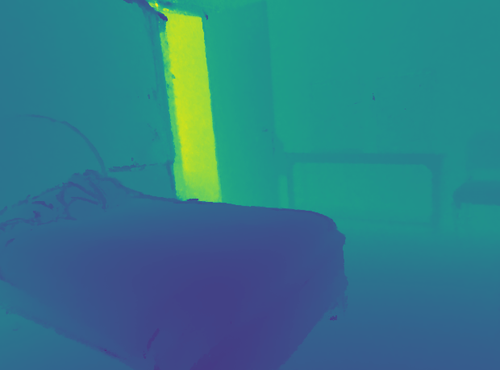} &
        \includegraphics[width=\sz\columnwidth]{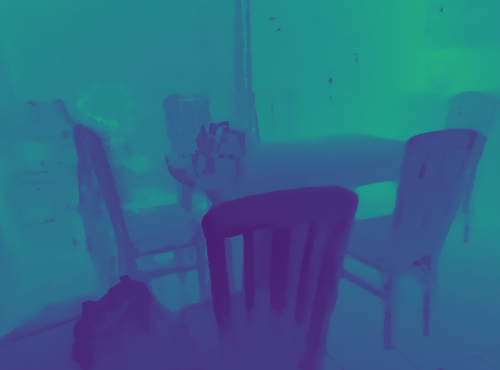} \\
    \end{tabular}
    \caption{\textbf{Qualitative NeRF reconstruction results on the NYU-Depth V2 Dataset~\cite{silberman2012indoor}} We show original images from the NYU-Depth V2 dataset and the corresponding images generated using the trained and filtered NeRFs for the same camera poses. The reconstructed RGB images appear very similar to their real counterparts. The reconstructed depth maps are also very close to the ground truth depth maps. However, there is some noise apparent on the edges of objects in the reconstructed depth maps. Color scale: 0 (purple) to 10 meters (yellow).}
    \label{fig:nerf_reconstruction_nyu}
\end{figure}

\begin{figure}[htb!]
    \centering
    \scriptsize
    \setlength{\tabcolsep}{1pt}
    \renewcommand{\arraystretch}{0.8}
    \newcommand{\sz}{0.23}
    \newcommand{\sh}{1.9cm}
    \begin{tabular}{ccccc}
        \rotatebox{90}{\hspace{22pt}RGB} &
        \includegraphics[width=\sz\linewidth]{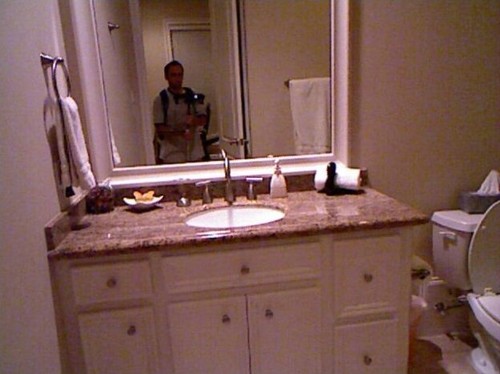} &
        \includegraphics[width=\sz\linewidth]{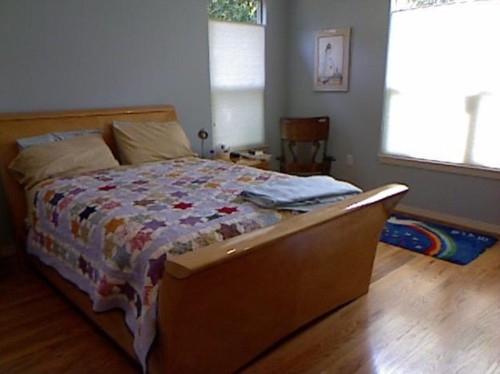} &
        \includegraphics[width=\sz\linewidth]{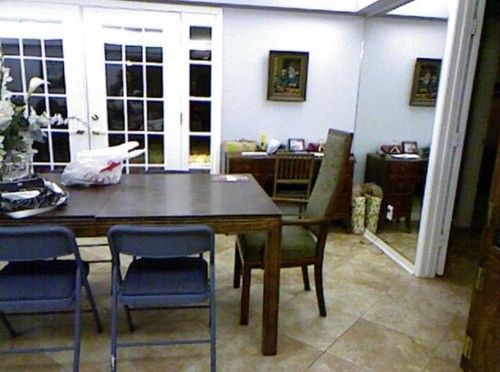} &
        \includegraphics[width=\sz\linewidth]{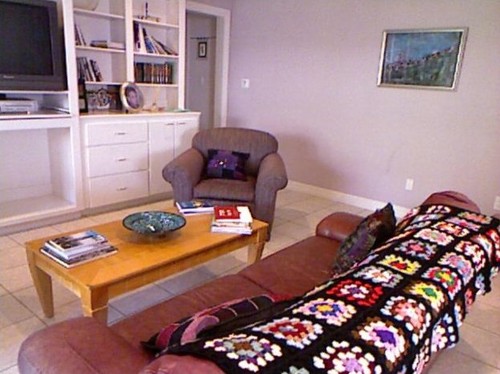}
        \\
        \rotatebox{90}{\hspace{14pt}GT Depth} &
        \includegraphics[width=\sz\linewidth]{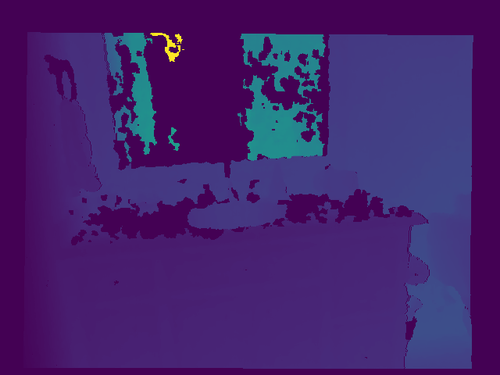} &
        \includegraphics[width=\sz\linewidth]{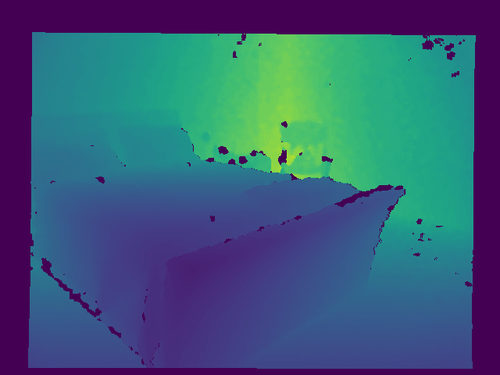} &
        \includegraphics[width=\sz\linewidth]{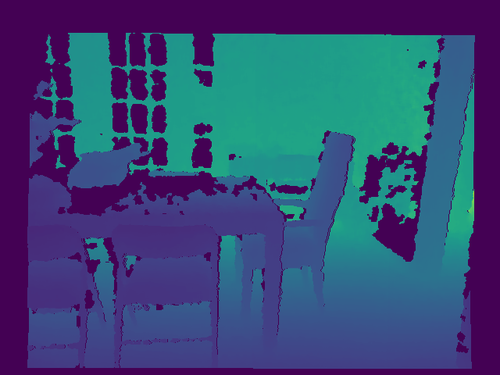} &
        \includegraphics[width=\sz\linewidth]{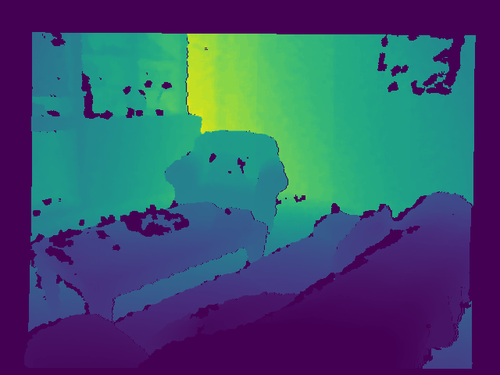}
        \\
        \rotatebox{90}{\hspace{12pt}AdaBins~\cite{bhat_adabins_2021}} &
        \includegraphics[width=\sz\linewidth]{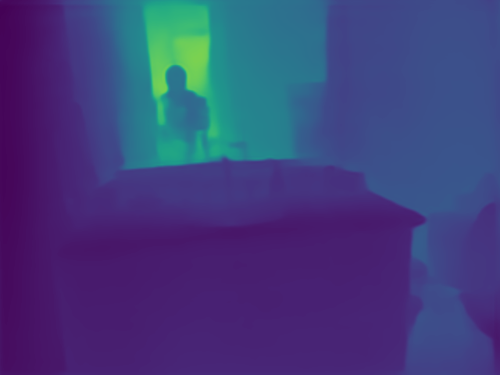} &
        \includegraphics[width=\sz\linewidth]{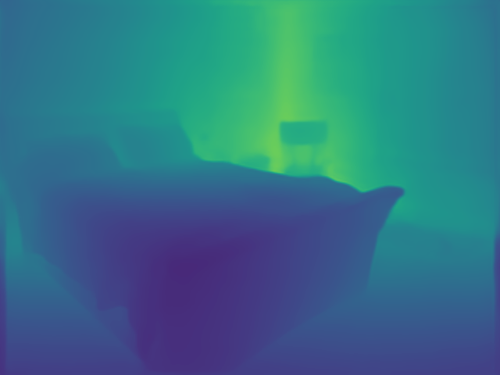} &
        \includegraphics[width=\sz\linewidth]{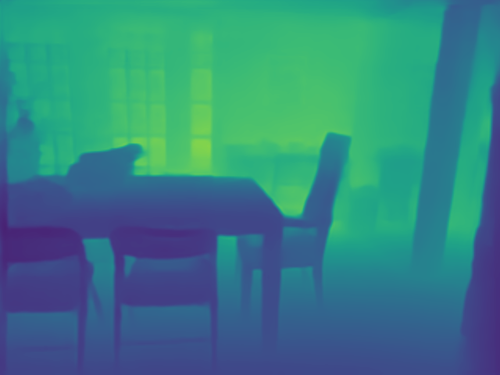} &
        \includegraphics[width=\sz\linewidth]{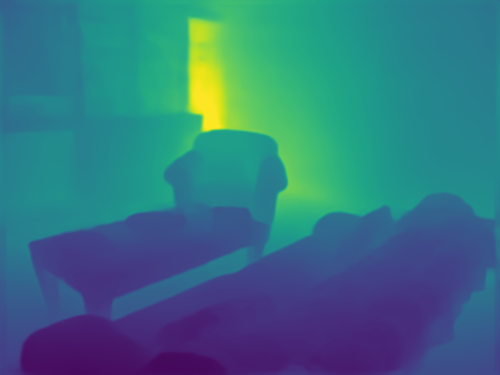}
          \\
        \rotatebox{90}{\hspace{4pt}\shortstack{\textbf{NeRFmented}\\AdaBins~\cite{bhat_adabins_2021}}} &
        \includegraphics[width=\sz\linewidth]{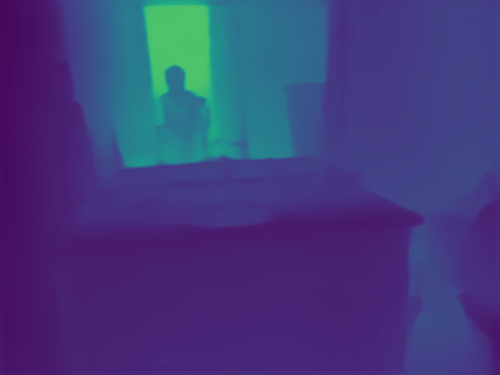} &
        \includegraphics[width=\sz\linewidth]{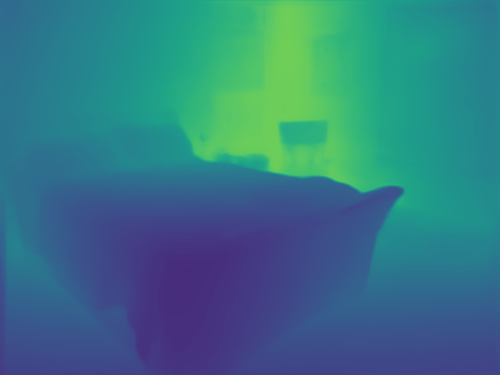} &
        \includegraphics[width=\sz\linewidth]{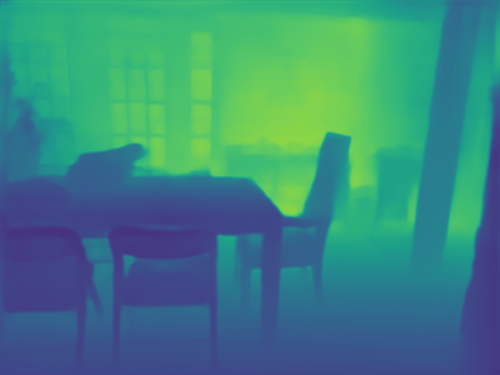} &
        \includegraphics[width=\sz\linewidth]{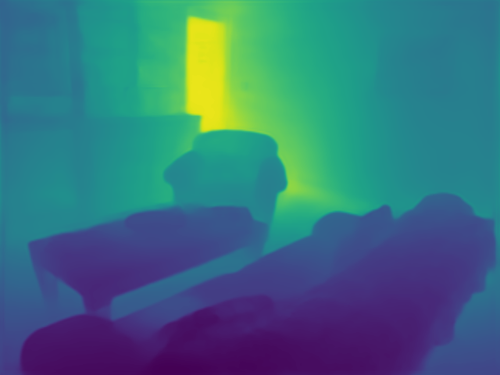}
        \\
        \rotatebox{90}{\hspace{2pt}DepthFormer~\cite{li2023depthformer}} &
        \includegraphics[width=\sz\linewidth]{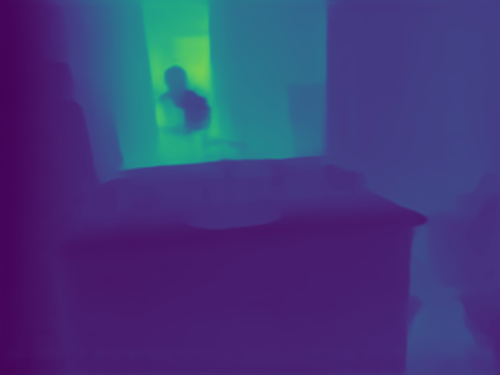} &
        \includegraphics[width=\sz\linewidth]{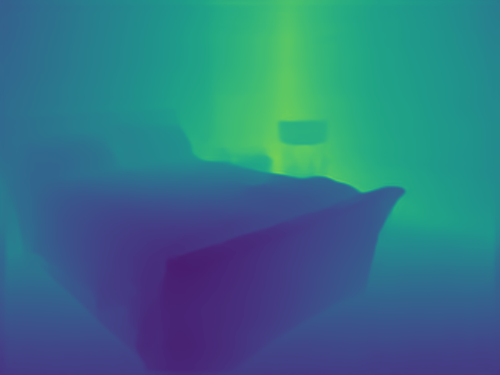} &
        \includegraphics[width=\sz\linewidth]{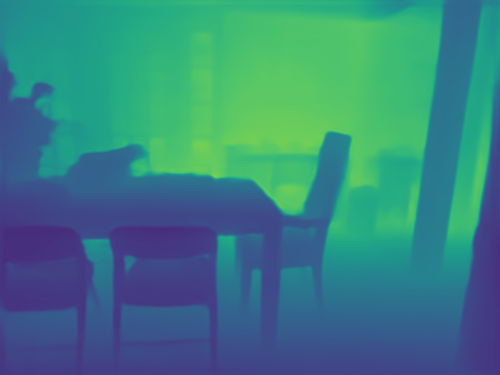} &
        \includegraphics[width=\sz\linewidth]{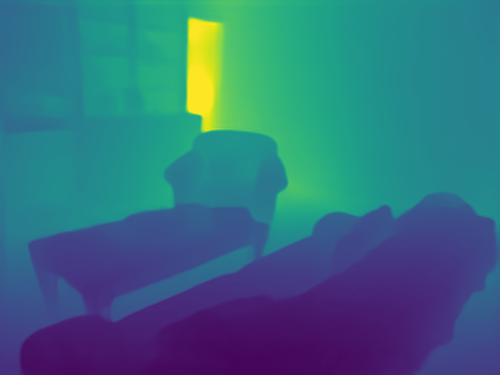}
        \\
        \rotatebox{90}{\hspace{0pt}\shortstack{\textbf{NeRFmented}\\DepthFormer~\cite{li2023depthformer}}} &
        \includegraphics[width=\sz\linewidth]{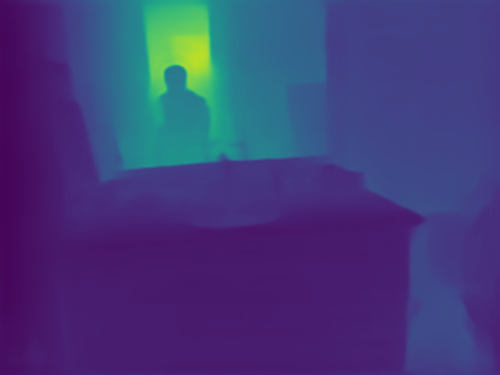} &
        \includegraphics[width=\sz\linewidth]{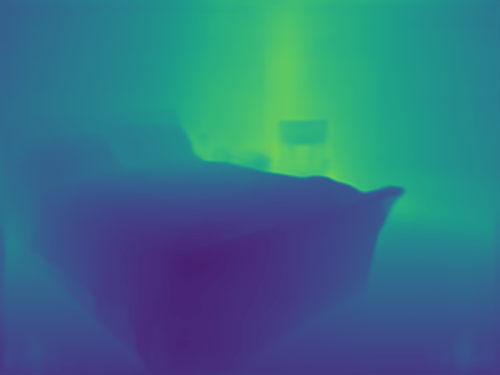} &
        \includegraphics[width=\sz\linewidth]{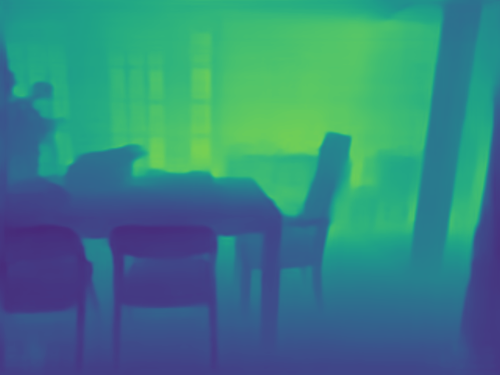} &
        \includegraphics[width=\sz\linewidth]{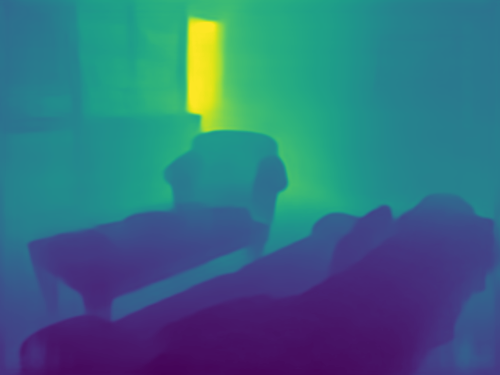}
        \\
    \end{tabular}
    \caption{\textbf{Qualitative Results on the NYU-Depth V2 Dataset~\cite{silberman2012indoor}} We show the performance of NeRFmentation (Ours) compared to vanilla-trained AdaBins~\cite{bhat_adabins_2021} and DepthFormer~\cite{li2023depthformer}. Both the NeRFmented models and vanilla-trained models have been trained on the NYU-Depth V2 train split and evaluated on the test split~\cite{eigen2014depth}. There is little discernible visual difference between the predictions of the vanilla-trained models and NeRFmented models. This is in line with the scores presented in \cref{tab:nyu_dataset}. Color scale: 1 (purple) to 6.5 meters (yellow).}
    \label{fig:qualitative_nyu}
\end{figure}

The aforementioned MDE networks are trained on the original NYU-Depth V2 train split and NeRFmented versions of it using the approach described in Section~4.5. The relevant results are provided in \cref{tab:nyu_dataset}. NeRFmentation appears to hinder performance for both models and all novel view synthesis strategies. Unlike in the case of KITTI, for NYU-Depth V2, the introduction of synthetic images may not improve generalization to a more diverse set of poses since the original data is not particularly lacking in pose diversity. Instead, the introduction of noisy images that reduce the data quality becomes the main factor determining the net outcome of the augmentation. Additionally, our experiments on the KITTI dataset suggest that dense supervision coming from NeRFmentation is a large factor in why NeRFmentation improves on vanilla training. With NYU-Depth V2, the vanilla dataset already provides dense supervision. Therefore, this advantage of NeRFmentation is lost when using the NYU-Depth V2 dataset. Please refer to \cref{fig:nerf_reconstruction_nyu} and \cref{fig:qualitative_nyu} for qualitative results.

\subsection{Zero-shot data transfer to Replica Dataset} 

To evaluate performance on a different indoor dataset, the Replica Dataset~\cite{straub_2019_CVPR} is used. Replica consists of high-quality 3D indoor scene reconstructions, of which we choose the two validation scenes, and the one test scene from Bauer  \etal \cite{bauer_nvs-monodepth_2021} as our validation split. We modify the source code of the provided ReplicaRenderer and ReplicaViewer \cite{straub_2019_CVPR} to be able to render customized sequences of synthetic RGB-D images following a natural trajectory through each scene. NeRFmentation deteriorates generalization to the unseen data distribution of synthetic images from the Replica dataset, as seen in \cref{tab:replica_dataset} and \cref{fig:qualitative_nyu_replica}. These results together with the NYU test set results show that the training data becomes too noisy for potential NeRFmentation benefits in generalization to unseen data distributions to be possible. 

\subsection{Ablation on NeRF-generated NYU-Depth V2 dataset}

Similar to the procedure applied in the case of KITTI-based NeRFmented MDE architectures, we aim to demonstrate generalization to perturbed poses for the NYU-Depth V2 dataset. The same pose disturbance scheme as in KITTI is used, and the scheme is applied to NeRFs trained on the test scenes remaining after the same filtration process as in KITTI experiments to eliminate subpar scenes. To create a test set of approximately the same size as the original NYU-Depth V2 test set, up to 10 frames per scene are selected at equal intervals along the scene sequence, and new images are rendered using corresponding perturbed poses. With this approach, the perturbed test set consists of 700 images covering 73 scenes, capturing a diverse set of indoor environments. As demonstrated in \cref{tab:nyu_augmented_dataset}, the NeRFmented MDE models surpass the baseline AdaBins~\cite{bhat_adabins_2021} models in the case of NeRFmentation using reconstructed views. In contrast, the DepthFormer~\cite{li2023depthformer} models do not exhibit a conclusive advantage for either the NeRFmented or the vanilla-trained models.

\begin{table}[htb!]
\normalsize
\setlength{\tabcolsep}{2.8pt}
    \caption{\textbf{Comparison of performances on the NYU-Depth-v2~\cite{silberman2012indoor} dataset.} \small This table shows the performance of AdaBins~\cite{bhat_adabins_2021} and DepthFormer~\cite{li2023depthformer} models trained on the NYU-Depth V2 train split~\cite{eigen2014depth, silberman2012indoor} and compare it to the NeRFmented versions of the same models, using different novel view generation strategies. Both vanilla-trained and NeRFmented models were evaluated against the NYU-Depth V2 test split~\cite{eigen2014depth}. Our method yields reduced performance in most metrics for both models. The best results are in \textbf{bold} dark green. The second best results are \underline{underlined} light green.}
    \begin{tabular}{clcccccc}
        \toprule
        \small
        & Augmentation & $\delta_{1}\uparrow$ & $\delta_{2}\uparrow$    & $\delta_{3}\uparrow$ & REL $\downarrow$ & RMS $\downarrow$  & RMS$_{LOG}$ $\downarrow$ \\
        \midrule
        \multirow{4}{*}{\rotatebox[origin=c]{90}{\textbf{AdaBins}}} & \textcolor{orange}{\spacedbullet}Classic & \fs{0.903} & \fs{0.984} & \fs{0.997} & \fs{0.103} & \fs{0.364}& \fs{0.044} \\
        &\textcolor{cyan}{\spacedbullet}Reconstructed & 0.877 & 0.977 & 0.994 & 0.118 & 0.414 & 0.049 \\
        &\textcolor{cyan}{\spacedbullet}Interpolated                    & \nd{0.898} & \nd{0.983} & \nd{0.996} & \nd{0.106} & \nd{0.375} & \nd{0.045} \\
        &\textcolor{cyan}{\spacedbullet}Angled $\pm 3^{\circ}$                    & 0.884 & 0.981 & 0.996 & 0.112 & 0.387 & 0.047 \\
        \midrule
        \multirow{4}{*}{\rotatebox[origin=c]{90}{\shortstack{\textbf{Depth} \\ \textbf{Former}}}}   & \textcolor{orange}{\spacedbullet}Classic    & \fs{0.921} & \fs{0.989} & \fs{0.998} & \fs{0.096} & \fs{0.339} & \fs{0.041} \\
        & \textcolor{cyan}{\spacedbullet}Reconstructed                    & \nd{0.909} & \nd{0.987} & \nd{0.997} & \nd{0.101} & \nd{0.350} & \nd{0.043} \\
        & \textcolor{cyan}{\spacedbullet}Interpolated                    & \nd{0.909} & \nd{0.987} & \nd{0.997} & \nd{0.101} & 0.353 & \nd{0.043} \\
        & \textcolor{cyan}{\spacedbullet}Angled $\pm 3^{\circ}$                    & 0.908 & \nd{0.987} & \fs{0.998} & 0.102 & \nd{0.350} & \nd{0.043} \\
        
        \bottomrule
    \end{tabular}
    \centering
  \label{tab:nyu_dataset}
\end{table}

\begin{table}[htb!]
\normalsize
\setlength{\tabcolsep}{2.8pt}
    \caption{\textbf{Comparison of performances on the Replica Dataset~\cite{straub_2019_CVPR}.} \small Models trained on the NYU-Test V2 train split~\cite{eigen2014depth, silberman2012indoor} are compared to the NeRFmented (Ours) versions of the same models evaluated against a subset of the Replica Dataset~\cite{straub_2019_CVPR}, a 3D scene reconstruction dataset. Our method impairs performance considerably on unseen synthetic indoor sequences, invalidating the possibility of improved generalization. The best results are in \textbf{bold} dark green. The second best results are \underline{underlined} light green.}
    \begin{tabular}{clcccccc}
        \toprule
        \small
        & Augmentation & $\delta_{1}\uparrow$ & $\delta_{2}\uparrow$    & $\delta_{3}\uparrow$ & REL $\downarrow$ & RMS $\downarrow$  & RMS$_{LOG}$ $\downarrow$ \\
        \midrule
        \multirow{4}{*}{\rotatebox[origin=c]{90}{\textbf{AdaBins}}} & \textcolor{orange}{\spacedbullet}Classic & \fs{0.512} & \fs{0.921} & \fs{0.982} & \fs{0.271} & \fs{0.821}& \fs{0.103} \\
        &\textcolor{cyan}{\spacedbullet}Reconstructed                    & 0.214 & 0.576 & 0.895 & 0.526 & 1.572 & 0.179 \\
        &\textcolor{cyan}{\spacedbullet}Interpolated                    & \nd{0.336} & \nd{0.772} & \nd{0.952} & \nd{0.398} & \nd{1.284} & \nd{0.140} \\
        &\textcolor{cyan}{\spacedbullet}Angled $\pm 3^{\circ}$                    & 0.300 & 0.713 & 0.916 & 0.418 & 1.297 & 0.157 \\
        \midrule
        \multirow{4}{*}{\rotatebox[origin=c]{90}{\shortstack{\textbf{Depth} \\ \textbf{Former}}}}   & \textcolor{orange}{\spacedbullet}Classic & \fs{0.410} & \fs{0.904} & \fs{0.978} & \fs{0.325} & \fs{0.937} & \fs{0.116} \\
        &\textcolor{cyan}{\spacedbullet}Reconstructed                    & \nd{0.150} & \nd{0.606} & \nd{0.943} & \nd{0.526} & \nd{1.540} & \nd{0.175} \\
        &\textcolor{cyan}{\spacedbullet}Interpolated                    & {0.115} & {0.542} & {0.908} & {0.582} & 1.634 & {0.190} \\
        &\textcolor{cyan}{\spacedbullet}Angled $\pm 3^{\circ}$                    & 0.109 & {0.486} & {0.908} & 0.604 & {1.752} & {0.196} \\
        
        \bottomrule
    \end{tabular}
    \centering
  \label{tab:replica_dataset}
\end{table}

\begin{table}[htb!]
\normalsize
\setlength{\tabcolsep}{2.8pt}
    \caption{\textbf{Comparison of performances on the perturbed NYU-Depth v2~\cite{silberman2012indoor} dataset.} \small This table shows the performance of baseline and NeRFmented models evaluated on a dataset of images generated using NeRFs trained on scenes not part of the NYU-Depth v2 train split~\cite{silberman2012indoor, eigen2014depth}. Images are generated using random disturbations to original image poses. Our method does not yield an architecture-independent performance increase, rendering the robustness ablation for indoor MDEs inconclusive. The best results are in \textbf{bold} dark green. The second best results are \underline{underlined} light green.}
    \begin{tabular}{clcccccc}
        \toprule
        \small
        & Augmentation & $\delta_{1}\uparrow$ & $\delta_{2}\uparrow$    & $\delta_{3}\uparrow$ & REL $\downarrow$ & RMS $\downarrow$  & RMS$_{LOG}$ $\downarrow$ \\
        \midrule
        \multirow{4}{*}{\rotatebox[origin=c]{90}{\textbf{AdaBins}}} & \textcolor{orange}{\spacedbullet}Classic & \nd{0.706} & \nd{0.907} & \nd{0.960} & \nd{0.794} & \nd{0.750}& \nd{0.094} \\
        &\textcolor{cyan}{\spacedbullet}Reconstructed                    & \fs{0.732} & \fs{0.912} & \fs{0.964} & \fs{0.622} & \fs{0.691} & \fs{0.083} \\
        &\textcolor{cyan}{\spacedbullet}Interpolated                    & {0.315} & {0.804} & {0.939} & {1.235} & {1.159} & {0.147} \\
        &\textcolor{cyan}{\spacedbullet}Angled $\pm 3^{\circ}$                    & 0.311 & 0.804 & 0.936 & 1.347 & 1.129 & 0.148 \\
        \midrule
        \multirow{4}{*}{\rotatebox[origin=c]{90}{\shortstack{\textbf{Depth} \\ \textbf{Former}}}} & \textcolor{orange}{\spacedbullet}Classic & 0.773 & 0.933 & 0.974 & \fs{0.882} & \fs{0.650} & 0.077 \\
        &\textcolor{cyan}{\spacedbullet}Reconstructed                    & 0.791 & \fs{0.935} & \fs{0.979} & \nd{0.916} & 0.742 & \nd{0.074} \\
        &\textcolor{cyan}{\spacedbullet}Interpolated                    & \fs{0.797} & \nd{0.934} &  \nd{0.978} & 0.948 & \nd{0.731} & \fs{0.072} \\
        &\textcolor{cyan}{\spacedbullet}Angled $\pm 3^{\circ}$                   & \nd{0.796} & \fs{0.935} & \fs{0.979} &  0.953 &  0.732 & \fs{0.072} \\
        \bottomrule
    \end{tabular}
    \centering
  \label{tab:nyu_augmented_dataset}
\end{table}

\newpage
\begin{figure}[htb!]
    \centering
    \scriptsize
    \setlength{\tabcolsep}{1pt}
    \renewcommand{\arraystretch}{0.8}
    \newcommand{\sz}{0.23}
    \newcommand{\sh}{1.9cm}
    \begin{tabular}{ccccc}
        \rotatebox{90}{\hspace{22pt}RGB} &
        \includegraphics[width=\sz\linewidth]{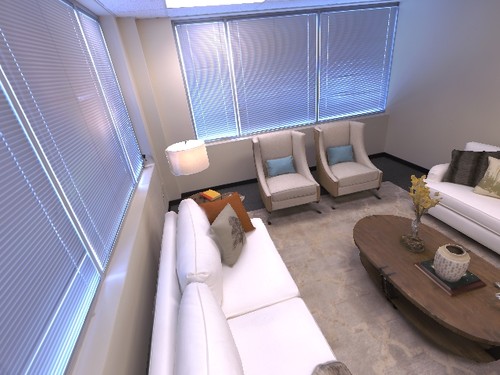} &
        \includegraphics[width=\sz\linewidth]{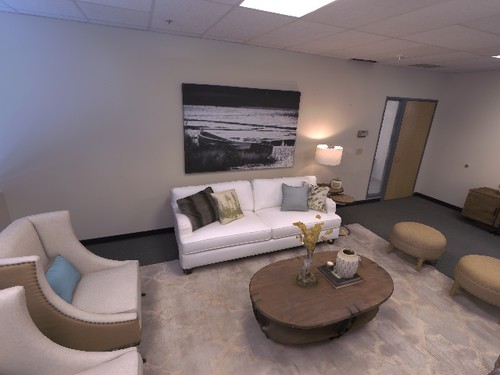} & 
        \includegraphics[width=\sz\linewidth]{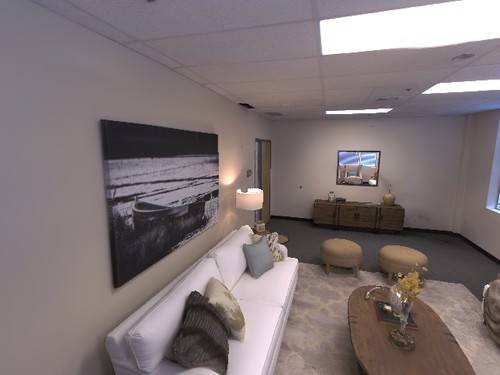} & 
        \includegraphics[width=\sz\linewidth]{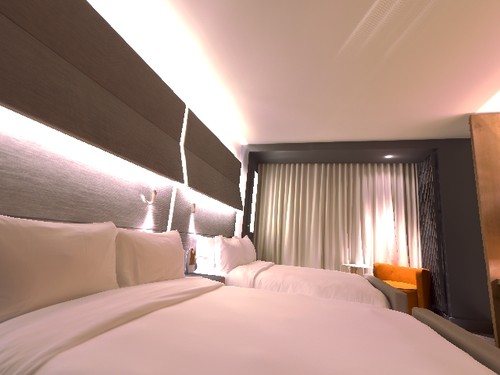} 
        \\
        \rotatebox{90}{\hspace{14pt}GT Depth} &
        \includegraphics[width=\sz\linewidth]{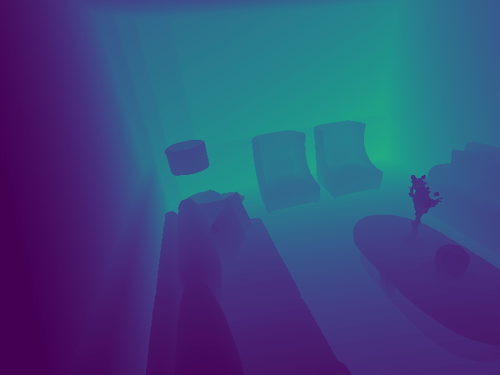} &
        \includegraphics[width=\sz\linewidth]{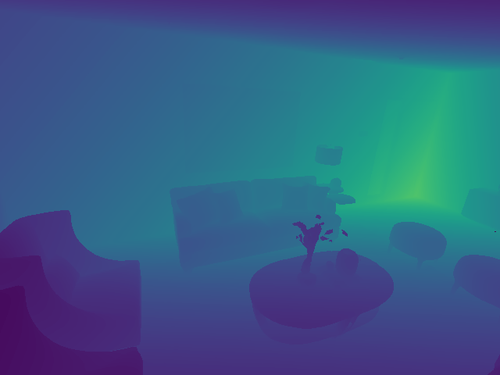} &
        \includegraphics[width=\sz\linewidth]{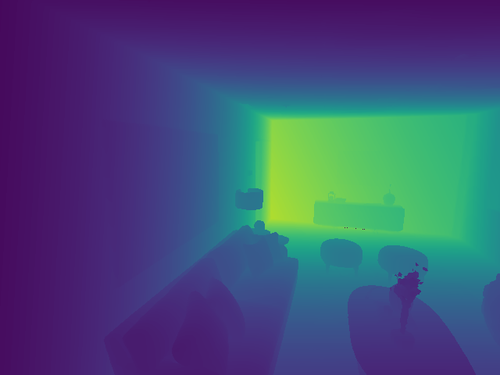} &
        \includegraphics[width=\sz\linewidth]{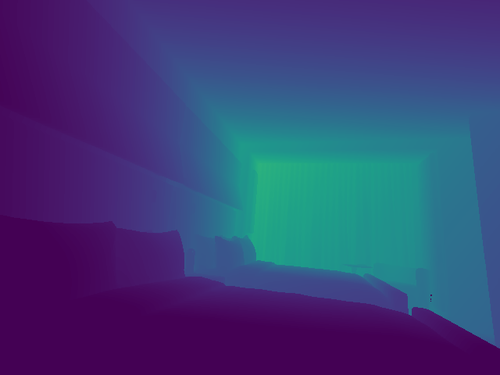}
        \\
        \rotatebox{90}{\hspace{12pt}AdaBins~\cite{bhat_adabins_2021}} &
        \includegraphics[width=\sz\linewidth]{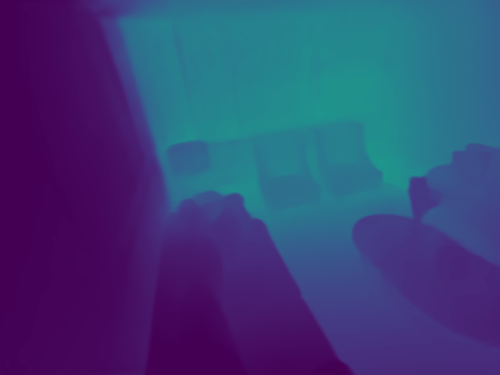} &
        \includegraphics[width=\sz\linewidth]{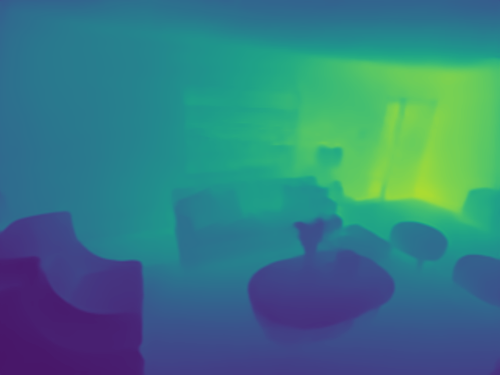} &
        \includegraphics[width=\sz\linewidth]{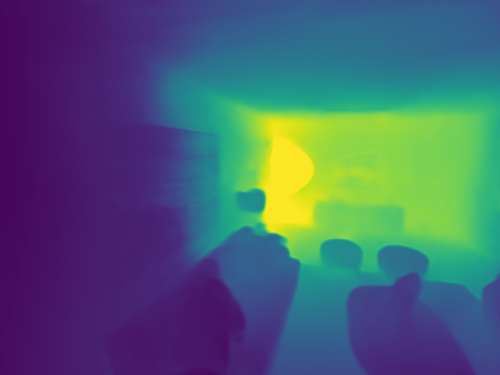} &
        \includegraphics[width=\sz\linewidth]{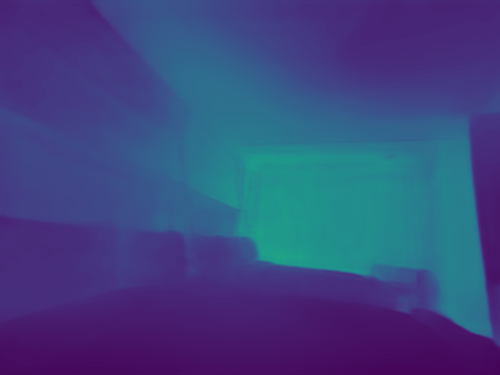}
        \\
        \rotatebox{90}{\hspace{4pt}\shortstack{\textbf{NeRFmented}\\AdaBins~\cite{bhat_adabins_2021}}} &
        \includegraphics[width=\sz\linewidth]{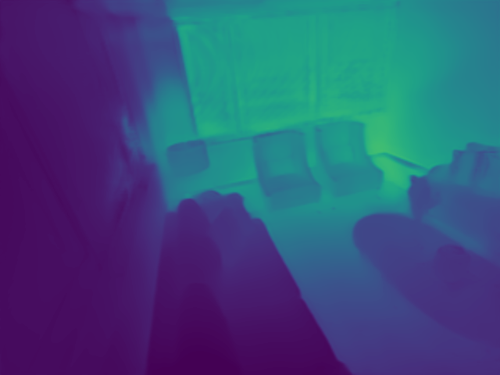} &
        \includegraphics[width=\sz\linewidth]{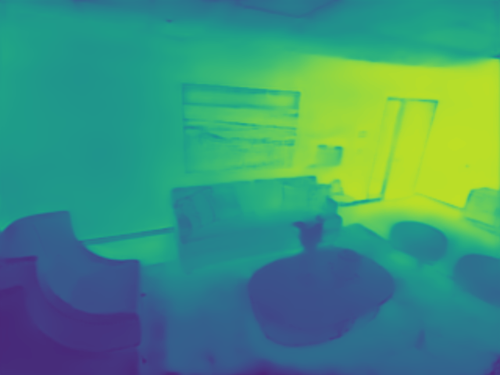} &
        \includegraphics[width=\sz\linewidth]{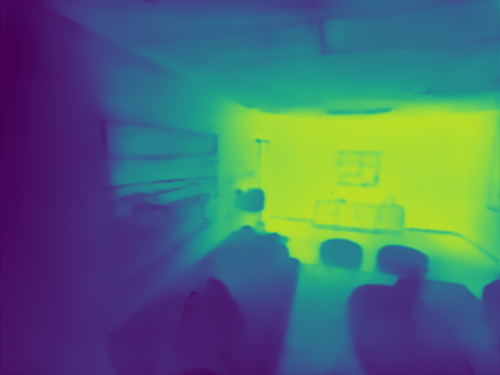} &
        \includegraphics[width=\sz\linewidth]{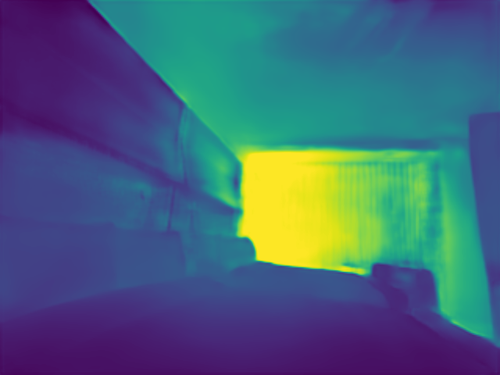}
        \\
        \rotatebox{90}{\hspace{2pt}DepthFormer~\cite{li2023depthformer}} &
        \includegraphics[width=\sz\linewidth]{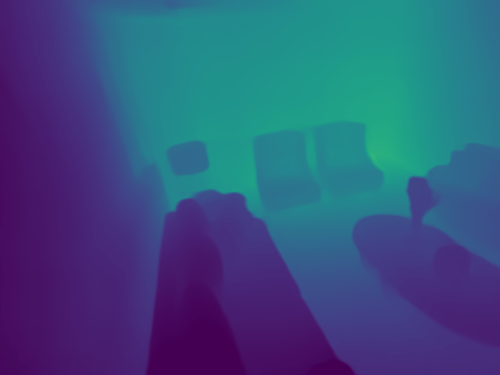} &
        \includegraphics[width=\sz\linewidth]{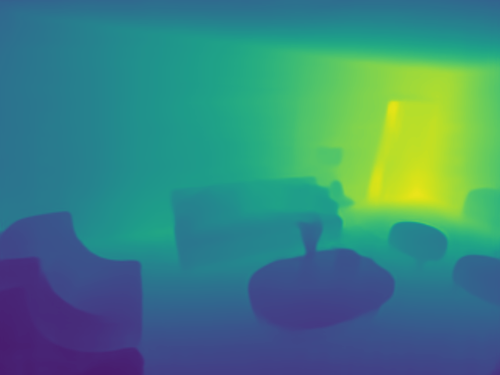} &
        \includegraphics[width=\sz\linewidth]{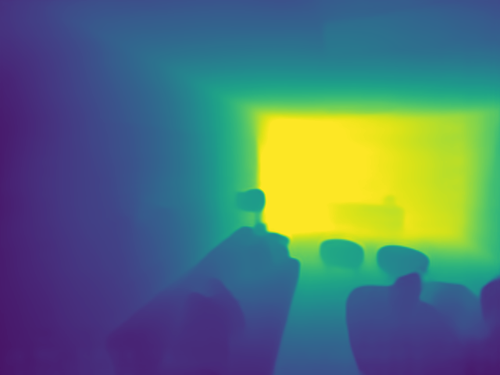} &
        \includegraphics[width=\sz\linewidth]{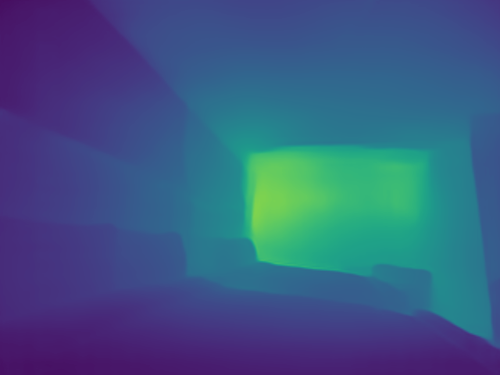}
        \\
        \rotatebox{90}{\hspace{0pt}\shortstack{\textbf{NeRFmented}\\DepthFormer~\cite{li2023depthformer}}} &
        \includegraphics[width=\sz\linewidth]{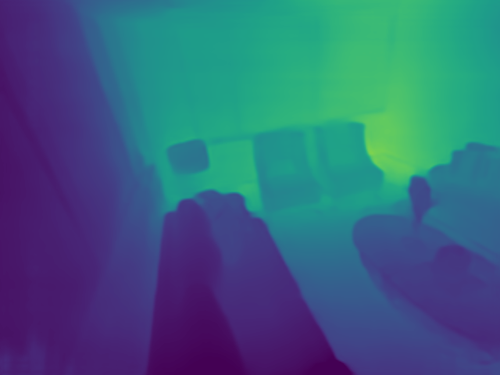} &
        \includegraphics[width=\sz\linewidth]{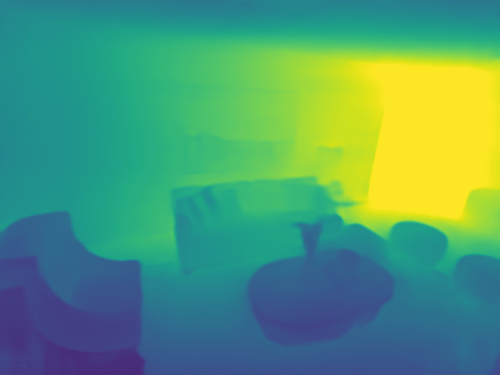} &
        \includegraphics[width=\sz\linewidth]{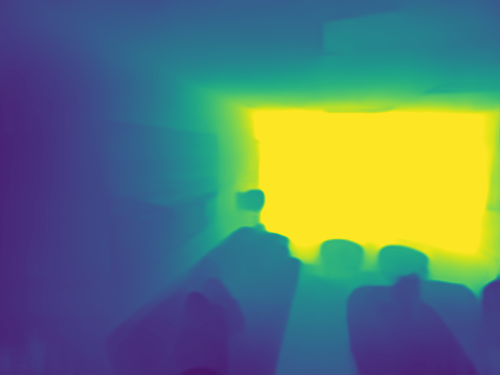} &
        \includegraphics[width=\sz\linewidth]{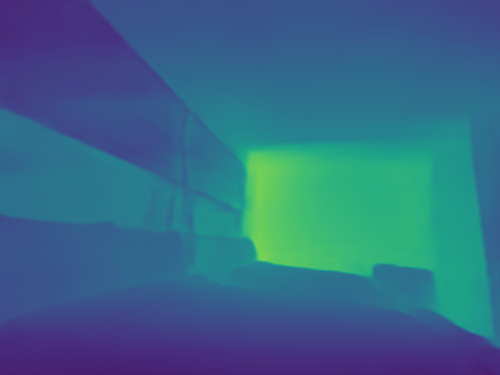}
        \\
    \end{tabular}
    \caption{\textbf{Qualitative Results on the Replica Dataset~\cite{straub_2019_CVPR}.} We show the performance of NeRFmentation (Ours) compared to vanilla-trained AdaBins~\cite{bhat_adabins_2021} and DepthFormer~\cite{li2023depthformer}. Both the NeRFmented models and vanilla-trained models were trained on the NYU-Depth V2 train split~\cite{eigen2014depth, silberman2012indoor} and evaluated on a subset of the unseen Replica Dataset~\cite{straub_2019_CVPR}. The NeRFmented models produce higher noise around edges of objects than the vanilla-trained models. This could be attributed to the noise introduced by NeRF-generated training data, as presented in \cref{fig:nerf_reconstruction_nyu}. Color scale: 1 (purple) to 8 meters (yellow).}
    \label{fig:qualitative_nyu_replica}
\end{figure}

\newpage

\FloatBarrier
\section{Future Works: Masking}
\label{sec:supp_masking}

Due to the sparsity of depth maps of datasets that were recorded using a LiDAR sensor such as KITTI, NeRFs that are trained on these datasets have to inter- and extrapolate data that they have never seen before. Interpolating data for a NeRF is mostly unproblematic. However, NeRFs struggle at the scene borders of the training data as they have to extrapolate in these regions. Since the level of supervision approaches zero in these regions, the NeRF learns to predict noise in these regions without any guarantee that the data there is useful.
In the best case, the data corresponds to the actual real-world geometry, but in the worst case, it will predict values that could hurt the overall training procedure and thus need to be taken care of.
This issue limits us from rendering more diverse poses for our NeRFmentation method, as any strong rotations or extreme translations will lead to noisy, but dense renders that we cannot mask out trivially without any extra steps.

In the following, we describe our proposed method to generate non-trivial binary occupancy masks that can be used to mask out unseen regions of NeRF-rendered depth maps during the training of an MDE network. These masks are designed to prevent the network from getting penalized for predicting incorrect values in pixels corresponding to these unseen regions. 

\paragraph{Point cloud generation.}
First, we need to obtain the 3D point cloud of each sub-scene $s$ in Cartesian coordinates. For that we must first obtain the homogeneous point cloud $\text{PC}_{s,h}$ using the following Equation~\eqref{eq:trafo}:

\begin{equation}
\begin{aligned}
    & \forall d_{k} \in \text{DM}_{\text{train},s} \subseteq \text{DM}_{\text{train}} \subseteq \mathcal{D}_{\text{train}}, \quad d_{k} \in \mathbb{R}^{U \times V}, \\
    & \text{PC}_{s,h} = \left\{ P_{W,s,h} \in \mathbb{R}^4 \mid d_k \ni d_{k,uv} > 0 \right\}, \\
    & \text{where} \\
    & P_{W,s,h} = T_{C_k}^{W} \cdot \begin{bmatrix} (K_C^{-1} \cdot p_{k,uv})^{T} \cdot d_{k,uv} \\ d_{k,uv} \end{bmatrix}, \\
    & K_C^{-1} \in \mathbb{R}^{3 \times 3}, \quad T_{C_k}^{W} \in \mathbb{R}^{4 \times 4}, \quad \text{PC}_{s,h} \in \mathbb{R}^{N \times 4}, \\
    & p_{c,uk} = p_c(u, v) \in \mathbb{R}, \quad N \leq U \cdot V \cdot \lvert \text{DM}_{\text{train},s} \rvert.
\end{aligned}
\label{eq:trafo}
\end{equation}
where $d_k$ is the k-th depth map of the training split $\text{DM}_{\text{train},s}$ of sub-scene $s$ which is a subset of the complete training dataset $\mathcal{D}_{\text{train}}$, and $P_{W,s,h}$ is the 4-dimensional homogeneous point in the world frame.
For each sub-scene $s$, we first unproject each valid 2D point of each sparse depth map $d_k$ that was part of the NeRF training split $\text{DM}_{\text{train},s}$ into the camera reference frame $C$ using homogeneous coordinates, the depth value $d_{k,uv}$ at every pixel index $u \in [0, U-1], v \in [0, V-1]$ and the inverse of the given intrinsic camera matrix $K_C \in \mathbb{R}^{3x3}$. Then we transform each homogeneous point into the world coordinate system $W$ using the given extrinsic matrices $T_{C_k}^{W} \in \mathbb{R}^{4x4}$. 
After this step, the 3D point cloud $\text{PC}_s$ is derived by dividing each point in $\text{PC}_{s,h}$ by its fourth homogeneous coordinate 
\begin{equation}
\text{PC}_s = \frac{1}{\text{PC}_{s,h}[3]} \cdot \begin{bmatrix} \text{PC}_{s,h}[0] \\ \text{PC}_{s,h}[1] \\ \text{PC}_{s,h}[2] \end{bmatrix},
\end{equation}
where $\text{PC}_{s,h}[i]$ denotes the $i$-th element of the homogeneous point cloud $\text{PC}_{s,h}$.

\paragraph{Mask generation.} 
When we generate an RGB-D image from a novel view, we store the pose used for each RGB-D image and split it into a rotation matrix $R_W^{NV_k}$ and a translation vector $t_W^{NV_k}$ to project the point cloud into the reference frame of the novel view $NV$ and using the intrinsic camera matrix $K_C$, we can then project the point-cloud into the image as a 2D projection and set the value for each pixel to 1 if it is occupied by data from at least one 3D point. This operation is the inverse of the point cloud generation process. It is described in the following \cref{eq:inv_trafo}:

\begin{equation}
\begin{aligned}
    &\forall NV_{k} \in \mathcal{D}_{\text{nerfmentation},s}  \subseteq \mathcal{D}_{\text{nerfmentation}}, \\
    & \text{mask}_{NV_k}(u,v) = \mathds{1} \{d_{NV_k, uv} > 0\}, \\
    & \text{where} \\
    & d_{NV_k, h} = K_C \cdot (R_W^{NV_k} \cdot \text{PC}_{s} + t_W^{NV_k}), \\
    & K_C \in \mathbb{R}^{3 \times 3}, \quad R_W^{NV_k} \in \mathbb{R}^{3 \times 3}, \quad t_W^{NV_k} \in \mathbb{R}^{3}, \\
    & \text{PC}_{s} \in \mathbb{R}^{N \times 3}.
\end{aligned}
\label{eq:inv_trafo}
\end{equation}

where $NV_k$ is the novel view which is defined by the rotation matrix $R_W^{NV_k} \mathbb{R}^{3 \times 3}$ and the translation vector $t_W^{NV_k} \in \mathbb{R}^{3}$. The depth map of novel view $NV_k$ retrieved from the ground-truth point cloud $PC_s$ might have multiple depth values per pixel $u,v$ which is irrelevant in our case since we only need to consider occupancy which is unaffected by multi-occupancy. 

\begin{figure}[htb!]
    \centering
    \scriptsize
    \setlength{\tabcolsep}{1pt}
    \renewcommand{\arraystretch}{0.8}
    \newcommand{\sz}{0.31}
    \newcommand{\sh}{1.65cm}
    \begin{tabular}{cccc}
        \rotatebox{90}{\hspace{6pt}Depth} &
        \includegraphics[width=\sz\linewidth]{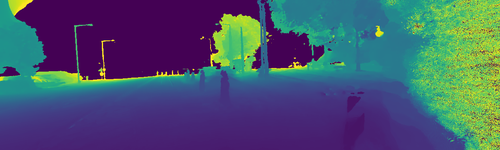} &
        \includegraphics[width=\sz\linewidth]{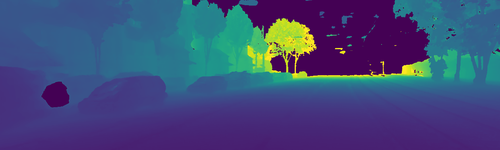} &
        \includegraphics[width=\sz\linewidth]{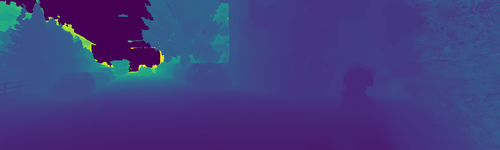} \\
        \rotatebox{90}{\hspace{8pt}RGB} &
        \includegraphics[width=\sz\linewidth]{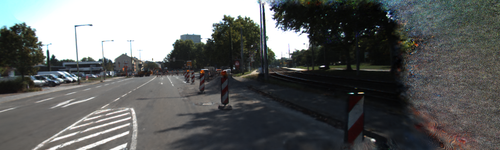} &
        \includegraphics[width=\sz\linewidth]{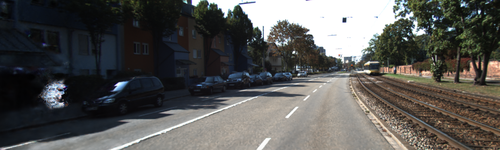} &
        \includegraphics[width=\sz\linewidth]{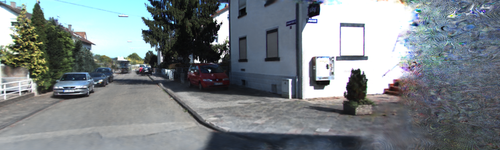} \\
        \rotatebox{90}{\hspace{6pt}Mask} &
        \includegraphics[width=\sz\linewidth]{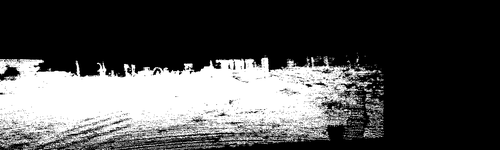} &
        \includegraphics[width=\sz\linewidth]{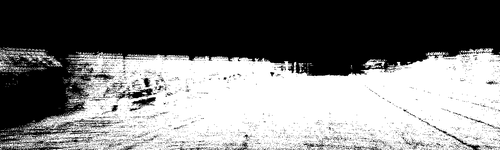} &
        \includegraphics[width=\sz\linewidth]{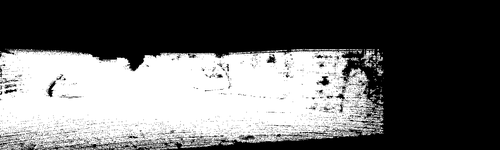} \\
        \rotatebox{90}{\hspace{2pt}\shortstack{Masked\\Depth}} &
        \includegraphics[width=\sz\linewidth]{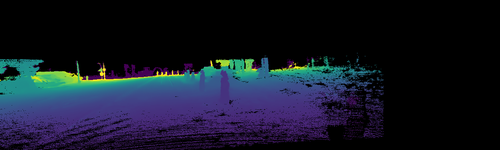} &
        \includegraphics[width=\sz\linewidth]{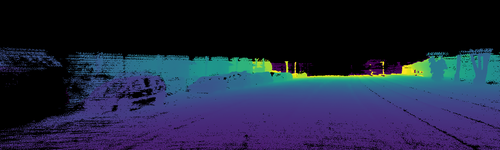} &
        \includegraphics[width=\sz\linewidth]{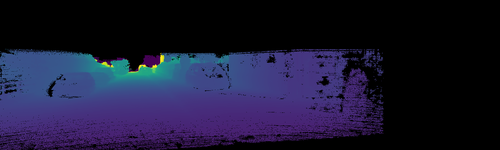} \\
        \rotatebox{90}{\hspace{2pt}\shortstack{Masked\\RGB}} &
        \includegraphics[width=\sz\linewidth]{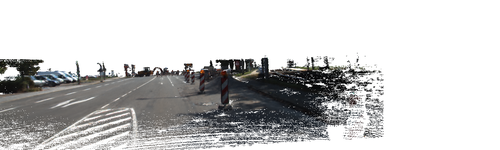} &
        \includegraphics[width=\sz\linewidth]{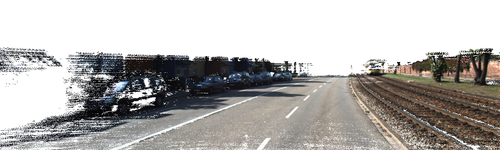} &
        \includegraphics[width=\sz\linewidth]{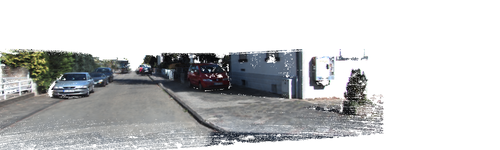} \\
    \end{tabular}
    \caption{\textbf{NeRF reconstruction results with 15 degrees rotation and occupancy masks on KITTI~\cite{geiger_kitti_2013}}. We show novel views from KITTI that were generated by rotating 15 degrees to the left or right, starting from a training pose. We show the rendered, dense depth map and the rendered RGB image which clearly show artifacts in the border regions. Using the occupancy masks, it is possible to filter out these artifacts, but at the same time sparsify the depth maps.}
    \label{fig:sup:masks}
\end{figure}

For future work, we plan to investigate the impact of this proposed method. Our proposed method sparsifies each RGB-D image again and excludes the sky, a step that is already done during the evaluation on KITTI. As a result, we expect this method to not yield a significant improvement over our proposed vanilla NeRFmentation. However, we do expect substantial benefits for renders captured from extreme poses. 
Since these renders would deviate strongly from the data distribution of KITTI and would be too sparse for Waymo, we expect a notable performance loss on KITTI and no gain on Waymo. As a result, our extended method might not be beneficial for current datasets. \cref{fig:sup:masks} shows rendered RGB-D images from challenging poses and how the occupancy masks can effectively remove these artifacts.

\newpage

\end{document}